\documentclass{article} 
\usepackage{iclr2026_conference,times}

\usepackage{amsmath,amsfonts,bm}

\def\eqref#1{equation~\ref{#1}}

\def\1{\bm{1}}

\def\rr{{\textnormal{r}}}

\DeclareMathAlphabet{\mathsfit}{\encodingdefault}{\sfdefault}{m}{sl}
\SetMathAlphabet{\mathsfit}{bold}{\encodingdefault}{\sfdefault}{bx}{n}

\usepackage{iac_pkg}
\usepackage{amsthm}
\usepackage{url}
\usepackage{lipsum}
\usepackage{colortbl}
\usepackage{xcolor}
\usepackage[most]{tcolorbox}
\definecolor{cvprblue}{rgb}{0.21,0.49,0.74}
\usepackage[pagebackref,breaklinks,colorlinks,allcolors=cvprblue]{hyperref}
\usepackage{tikz}
\usetikzlibrary{tikzmark}
\usepackage[capitalize]{cleveref}
\usepackage{enumitem}
\usepackage{cancel}
\usepackage{graphicx}
\usepackage{url} 
\usepackage{amssymb}
\theoremstyle{definition}
\newtheorem{definition}{Definition}[section]
\usepackage{booktabs}       
\usepackage{amsfonts}       
\usepackage{nicefrac}       
\usepackage{microtype}      
\usepackage{xcolor}         
\usepackage{wrapfig}        
\usepackage{subcaption}     
\usepackage{tikz}
\usepackage{pifont}
\newcommand{\cmark}{\ding{51}}%
\newcommand{\xmark}{\ding{55}}%
\usetikzlibrary{positioning}
\usetikzlibrary{calc}
\usepackage{multirow}
\definecolor{headerbg}{gray}{0.95}

\usepackage{comment}
 
\definecolor{green}{RGB}{102, 194, 165}
\definecolor{orange}{RGB}{252, 141, 98}
\definecolor{purple}{RGB}{141, 160, 203}
\definecolor{pink}{RGB}{231, 138, 195}
\definecolor{light-green}{RGB}{166, 216, 84}
\definecolor{green-correct}{RGB}{148, 212, 191}
\definecolor{orange-incorrect}{RGB}{253, 174, 145}
\definecolor{violet-marginal}{RGB}{141, 160, 203}

\definecolor{redgradient}{rgb}{0.95,0,0}
\definecolor{greengradient}{rgb}{0,0.95,0}
\definecolor{greendarkgradient}{rgb}{0.04, .6, .12}

\newcommand{\methodimagetag}[1]{%
  \tikz[baseline=(char.base)]{%
    \node[
      anchor=base, 
      fill=white,
      rounded corners=1pt,
      text=white,
      font=\bfseries,
      minimum width=1em,
      minimum height=1em,
      align=center,
      inner sep=0pt
    ](char){\includegraphics[height=1.1em]{#1}};%
  }%
}

\newtcolorbox{emphabox}[1][breakable,enhanced]{drop shadow={black!20!white}, 
	coltitle=black,
	top=1ex,
	attach boxed title to top left=
	{xshift=0em,yshift=-\tcboxedtitleheight/2},
	boxed title style={size=small,colback=black!50!white},#1}

\title{Spilled Energy in Large Language Models}

\author{Adrian R. Minut~$^{1,2}$ \And Hazem Dewidar~$^{1,2}$ \And Iacopo Masi~$^{1}$ 
\AND \vspace{-0.8cm} \\\methodimagetag{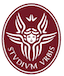} Sapienza University of Rome, Italy \quad\quad $^1$\methodimagetag{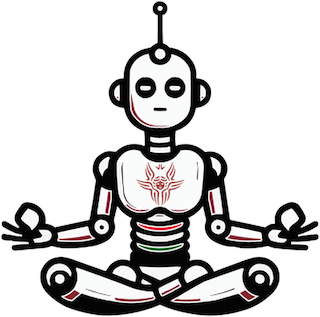}~\href{https://omnai.di.uniroma1.it}{{OmnAI Lab}} \, $^2$\methodimagetag{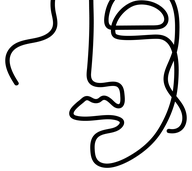}~\href{https://gladia.di.uniroma1.it}{{GLADIA}}
}

\newcommand{\eat}[1]{}

\definecolor{transpgray}{gray}{0.9}
\newcolumntype{b}{>{\columncolor{transpgray}}r}
\usepackage{booktabs}
\usepackage{cleveref}
\usepackage{cancel}
\iclrfinalcopy 

\begin{document}

\maketitle

\begin{abstract}
We reinterpret the final Large Language Model (LLM) softmax classifier as an Energy-Based Model (EBM), decomposing the sequence-to-sequence probability chain into multiple interacting EBMs at inference. This principled approach allows us to track ``energy spills" during decoding, which we empirically show correlate with factual errors, biases, and failures.  Similar to~\cite{orgad2024llms}, our method localizes the exact answer token and subsequently tests for hallucinations. Crucially, however, we achieve this without requiring trained probe classifiers or activation ablations. Instead, we introduce two completely training-free metrics derived directly from output logits: \textbf{spilled energy}, which captures the discrepancy between energy values across consecutive generation steps that should theoretically match, and \textbf{marginalized energy}, which is measurable at a single step. Evaluated on nine benchmarks across state-of-the-art LLMs (including LLaMA, Mistral, and Gemma) and on synthetic algebraic operations (Qwen3), our approach demonstrates robust, competitive hallucination detection and cross-task generalization. Notably, these results hold for both pretrained and instruction-tuned variants without introducing any training overhead. Code available at \href{http://github.com/OmnAI-Lab/spilled-energy/}{github.com/OmnAI-Lab/spilled-energy/}
\end{abstract}
\begin{figure}[htb]
\begin{center}
{\footnotesize\texttt{Q/A: ``What is the capital of Italy?  Answer:''}}\vspace{2pt}
\end{center}
\begin{center}
{\footnotesize
\noindent\begin{minipage}[t]{.45\textwidth}
\begin{center}
    {Logit}
\end{center}
\begin{emphabox}
\stkLOGITS{The}{0.56} \stkLOGITS{capital}{0.92} \stkLOGITS{of}{0.89} \stkLOGITS{Italy}{0.84} \stkLOGITS{is}{0.76} \stkLOGITS{Rome}{1.00}
\qquad~~\correct
\vspace{4pt}
\stkLOGITS{The}{0.61} \stkLOGITS{capital}{1.00} \stkLOGITS{of}{0.96} \stkLOGITS{Italy}{0.91} \stkLOGITS{is}{0.83} \stkLOGITS{Sydney}{0.76} \qquad~\incorrect
\end{emphabox}
\end{minipage}
~~
\begin{minipage}[t]{.45\textwidth}
\begin{center}
    \textbf{Spilled (Ours)}
\end{center}
\begin{emphabox}
\stk{The}{0.41} \stk{capital}{0.47} \stk{of}{0.18} \stk{Italy}{0.18} \stk{is}{0.09} \stk{Rome}{0.18} \qquad~ \correct
\vspace{4pt}
\stk{The}{0.40} \stk{capital}{0.46} \stk{of}{0.18} \stk{Italy}{0.18} \stk{is}{0.09} \stk{Sydney}{1.00}\qquad~ \incorrect
\end{emphabox}
\end{minipage}
}    
\end{center}
\vspace{3pt}
\begin{center}
{\footnotesize\texttt{Reasoning: ``A farmer has 12 chickens. Each chicken lays 2 eggs per day. How many eggs will the farmer collect in 5 days?''}}\vspace{2pt}
\end{center}
\begin{center}
{\footnotesize
\noindent\begin{minipage}[t]{.45\textwidth}
\begin{center}
    {Logit}
\end{center}
\begin{emphabox}
\stkLOGITS{12}{0.26} \stkLOGITS{chickens}{0.55} \stkLOGITS{lay}{0.61} \stkLOGITS{2}{0.39} \stkLOGITS{eggs}{0.56} \stkLOGITS{per}{0.66} \stkLOGITS{day}{0.47} \stkLOGITS{.}{0.28} \stkLOGITS{In}{0.51}  \stkLOGITS{5}{0.32} \stkLOGITS{days}{0.50} \stkLOGITS{,}{0.37} \stkLOGITS{the}{0.61} \stkLOGITS{farmer}{0.59} \stkLOGITS{will}{0.64} \stkLOGITS{collect}{0.29} \stkLOGITS{12}{0.35} \stkLOGITS{x}{0.74}  \stkLOGITS{2}{0.23} \stkLOGITS{x}{0.74}  \stkLOGITS{5}{0.34}  \stkLOGITS{=}{0.73}  \stkLOGITS{120}{0.38} \stkLOGITS{eggs}{0.32} \stkLOGITS{in}{0.66}  \stkLOGITS{5}{0.41} \stkLOGITS{days}{0.40}
\qquad~\correct
\vspace{5pt}
\stkLOGITS{12}{0.27} \stkLOGITS{chickens}{0.57} \stkLOGITS{lay}{0.63} \stkLOGITS{2}{0.40} \stkLOGITS{eggs}{0.58} \stkLOGITS{per}{0.69} \stkLOGITS{day}{0.49} \stkLOGITS{.}{0.29} \stkLOGITS{In}{0.53}  \stkLOGITS{5}{0.33} \stkLOGITS{days}{0.52} \stkLOGITS{,}{0.39} \stkLOGITS{the}{0.63} \stkLOGITS{farmer}{0.61} \stkLOGITS{will}{0.66} \stkLOGITS{collect}{0.30}\stkLOGITS{12}{0.36} \stkLOGITS{x}{0.76}  \stkLOGITS{2}{0.24} \stkLOGITS{x}{0.77}  \stkLOGITS{5}{0.35} \stkLOGITS{=}{0.76}  \stkLOGITS{470}{0.40} \stkLOGITS{eggs}{0.35} \stkLOGITS{in}{0.68} \stkLOGITS{5}{0.43} \stkLOGITS{days}{0.41} \qquad~~ \incorrect
\end{emphabox}
\end{minipage}
~~
\begin{minipage}[t]{.45\textwidth}
\begin{center}
    \textbf{Spilled (Ours)}
\end{center}
\begin{emphabox}
\stk{12}{0.18} \stk{chickens}{0.35} \stk{lay}{0.48}  \stk{2}{0.16} \stk{eggs}{0.26} \stk{per}{0.49} \stk{day}{0.05} \stk{.}{0.33} \stk{In}{0.61}  \stk{5}{0.10}  \stk{days}{0.56} \stk{,}{0.34} \stk{the}{0.47} \stk{farmer}{0.35} \stk{will}{0.40} \stk{collect}{0.06}  \stk{12}{0.31} \stk{x}{0.55}  \stk{2}{0.19} \stk{x}{0.33}  \stk{5}{0.09}  \stk{=}{0.42}   \stk{120}{0.15} \stk{eggs}{0.16} \stk{in}{0.80}  \stk{5}{0.25} \stk{days}{0.00}
\qquad~ \correct
\vspace{5pt}
\stk{12}{0.17} \stk{chickens}{0.33} \stk{lay}{0.47}  \stk{2}{0.15} \stk{eggs}{0.24} \stk{per}{0.48} \stk{day}{0.03} \stk{.}{0.31} \stk{In}{0.60}  \stk{5}{0.08}  \stk{days}{0.55} \stk{,}{0.33} \stk{the}{0.46} \stk{farmer}{0.34} \stk{will}{0.38} \stk{collect}{0.04}  \stk{12}{0.30} \stk{x}{0.55}  \stk{2}{0.17} \stk{x}{0.31}  \stk{5}{0.07}  \stk{=}{0.41}   \stk{470}{0.83} \stk{eggs}{0.23} \stk{in}{0.79}  \stk{5}{0.26} \stk{days}{0.00}
\qquad~~ \incorrect
\end{emphabox}
\end{minipage}
}    
\end{center}

\caption{Color-coded comparison of hallucination detection with LLaMa-3 8B using logit confidence and \tbf{spilled energy}. 
Our method generalizes well across topics (e.g., Q\&A, reasoning) and diverse LLMs. 
\correct~indicates a correct answer and \incorrect~an incorrect one. 
While \tbf{our approach focuses on the exact answer tokens} (e.g. Rome/Sydney and 120/470, see \cref{sec:detect-hallucinations}), here we apply min–max normalization to the full answer for visualization, as truthful \protect\tikz[baseline=-0ex]\protect\shade[left color=green!80,middle color=green!50,right color=red!35] (0,0) rectangle (3,0.25); hallucination.}

\label{fig:teaser}
\end{figure}

\section{Introduction}
The widespread adoption of Large Language Models (LLMs) across various domains has brought increasing attention 
to their critical limitation: their tendency to generate 
incorrect or misleading information---commonly referred 
to as ``hallucinations.'' This issue supports the idea 
that LLMs are just stochastic parrots~\citep{bender2021dangers} answering in a way that is statistically plausible with respect to the input prompt despite not having a real understanding of it. 
On the other side, recent reasoning capabilities proper to ChatGPT 4o~\citep{openai2023gpt4} or Deepseek~\citep{liu2024deepseek} offer counter evidence to actually support this.

Ongoing research seeks to characterize and categorize hallucinations, setting them apart from other error types \citep{liu-etal-2022-token, ji2023survey, huang2023survey, rawte2023troubling}. At the same time, recent discussions have introduced terms such as confabulations \citep{millidge2023llms} and fabrications \citep{mcgowan2023chatgpt}, sometimes attributing a form of ``intention'' to LLMs—though the very idea of LLM ``intentionality'' and other human-like qualities remains contested \citep{salles2020anthropomorphism, serapio2023personality, harnad2024language}.
Research on LLM hallucinations can be categorized into two main branches: the first one is the extrinsic branch, where the hallucinations are measured with respect to the interpretation that humans give to those errors \citep{bang2023multitask, ji2023survey, huang2023survey, rawte2023troubling}.
The second branch was started by \cite{kadavath2022language}, proposing to study the hallucinations \emph{within} the model itself. Following~\cite{kadavath2022language}, the work in \cite{li2024inference} proposes Inference-Time Intervention (ITI) as a way to improve the ``truthfulness'' of LLMs at inference time.
ITI functions by altering model activations at inference time, steering them along specific directions within a restricted set of attention heads.
Our work is also different from~\cite{yin2023large}, since we care about detecting errors in LLMs, whereas they introduce an automated methodology to detect when LLMs are aware that they do not know how to answer. 

In this work, we follow the definition of hallucinations given by~\cite{orgad2024llms} as any form of error produced by an LLM—including factual mistakes, biased outputs, breakdowns in common-sense reasoning, and related issues. 
Like them, we also confirm that the truthfulness signal is concentrated in the ``exact answer tokens.'' Nevertheless, unlike them, we abandon the idea of using a probe classifier~\citep {belinkov2021probingclassifierspromisesshortcomings} trained for each task and dataset. Given that LLMs are foundational models, user interactions typically occur \emph{in the wild}, making it difficult to predict which probe classifier is best suited for detecting hallucinations in real-world scenarios. Furthermore, in this setting, classifier weights should not only be updated dynamically for each task, but the optimal token–layer combination is also dataset-dependent, which conflicts with the broad LLM applicability.
Indeed, in the work by~\cite{orgad2024llms}, the authors report:
\begin{quote}
\begin{center}
``We find that probing classifiers do not generalize across different tasks.''     
\end{center}
\end{quote}
In our paper, we propose to solve this problem with a training-free method that generalizes better across different tasks and is mathematically principled using the framework of Energy-based Models (EBMs). \cref{fig:teaser} reports a qualitative comparison across tasks, comparing to the logit confidence. Additional samples are shown in \cref{sec:supp-qual}.

We reinterpret the final softmax classifier over the vocabulary of LLM as an EBM, taking inspiration from what~\cite{grathwohl2020your} did for classifiers.
This perspective enables us to decompose the sequence-to-sequence probability chain into multiple interacting EBMs that operate jointly during inference. Through this decomposition, we introduce the notion of ``spilled energy'' in LLM decoding and show empirically that such spill strongly correlates with errors. Given that our method is solely based on the mathematics of EBMs and the chain rule of probability, we do not have to train or tune our detector, striking a good generalization across tasks and LLMs.
Building on this foundation, our contributions are as follows:
\begin{itemize}[itemsep=2pt, leftmargin=*]
\item Training-free, LLM hallucination detection generalizing across tasks using the EBM framework. We introduce a method for detecting hallucinations that requires no additional training, in contrast to prior work that relies on trained classifiers and ablations of model activations. Our approach directly reads values inside the LLM, enabling natural generalization across tasks and performing better than logit-based detection.
\item Two energy-based metrics. We define two complementary measures of energy spills: (i) delta energy $\Delta\Eseq{i}$, which captures discrepancies between energy values across two time steps that should be mathematically equivalent, and (ii) marginal energy $\Eseqq{i}{m}$, which can be evaluated at a single time step.
\item Scalable and generalizable analysis. Our framework is mathematically principled, training-free, and exhibits strong cross-dataset generalization. We scale our analysis to state-of-the-art LLMs, including Llama 3-8B-Instruct and Mistral-7B-Instruct, and demonstrate competitive performance across nine benchmarks, showing robustness across datasets and architectures.
\end{itemize}

\begin{figure}
    \centering
    \includegraphics[width=0.9\linewidth]{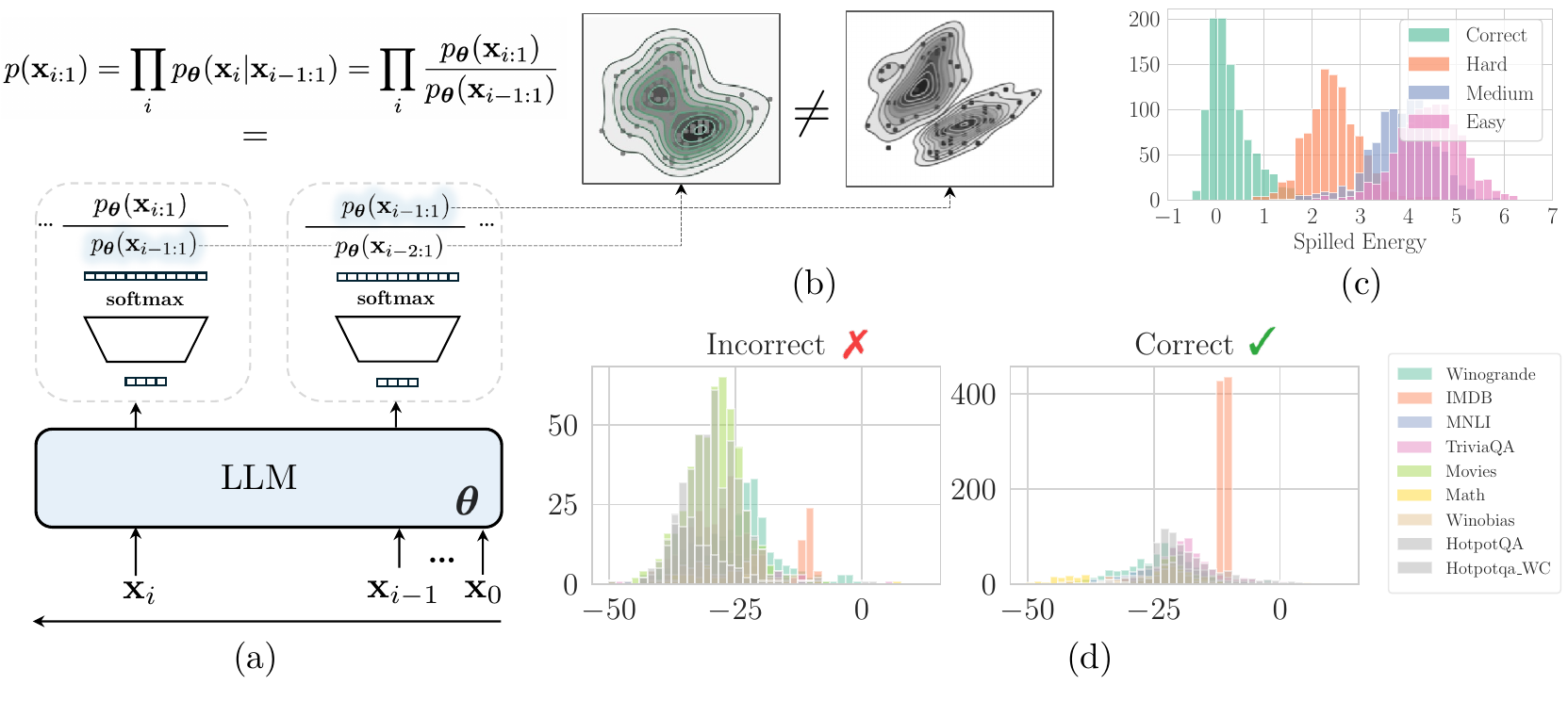}
    \caption{{\bf How energy spills in LLMs}. (a) Language Modeling $p(\compactseq[]{i})$ is attained as a decomposition problem following the chain rule of probability, implemented as autoregressive: we recursively apply a discriminative classifier over the vocabulary $\mathcal{V}$ to attain generative modeling with larger context size i.e. $p(\bx_{i}|\compactseq[]{i-1})$. (b) We reinterpret each discriminative classifier as a generative EBM, finding a connection between two quantities that should be the same across time steps yet are different. We call this difference ``the spilled energy'' $\Delta\Eseq{i}$ in \cref{def:spilled-energy}. (c) Given that we simply read values inside the LLM, our approach is training-free and correlates well with hallucinations on a synthetic math dataset with increasing difficulty; (d) histograms of spilled energy values, for incorrect and correct answers on all nine datasets using $\min$ pooling for Llama-3-Instruct. The two distributions are easily separable by using a simple threshold, resulting in a generalization across real-world tasks.}
    \label{fig:spilled-idea}
\end{figure}

\cref{fig:spilled-idea}(a) illustrates the core idea of our method: rather than using a na\"{i}ve approach, such as simply recording the logit or training a probe classifier at the activations of the answer token, we first reinterpret the LLM as an autoregressive EBM via the chain rule of probabilities. We then further decompose each conditional probability, incorporating insights from~\cite{grathwohl2020your}. At the time step of the exact token $i-1$, we extract the energy, which corresponds to the logit, and compare it with the marginal energy at the next time step $i$, corresponding to the denominator of the softmax. According to the chain rule, these two quantities should be identical; however, they differ in the LLM implementation---\cref{fig:spilled-idea}(b).
We find that the discrepancy, which we term spilled energy $\Delta\Eseq{i}$, correlates strongly with instances where the LLM produces an incorrect output—see \cref{fig:spilled-idea}(c). Moreover, its detection signal separates well correct and incorrect classes across datasets, reflecting the model’s confidence, as shown in \cref{fig:spilled-idea}(d).
\section{Related Work}

\minisection{EBM applications to Trustworthy AI} EBMs have been applied to improve the reliability and interpretability of Deep Nets. For example, Energy-Based Out-of-Distribution Detection (OOD)~\citep{energyOOD} uses the energy score as a more robust alternative to softmax confidence. At the same time, \cite{grathwohl2020your} presents how to reinterpret a discriminative classifier as EBM to train models that are both discriminative and generative. Following this work,~\cite{zhu2021towardsunderstandingthegenerativecapabilities} provides new insights into the role of energy when training EBMs and robust classifiers using adversarial training. Instead, \cite{mirza2024shedding, mirza2025understandingadversarialtrainingenergybased} explain adversarial attacks by reinterpreting the softmax classifier as an EBM, showing that these perturbations correspond to shifts in the underlying energy landscape.

\minisection{Foundations of Hallucination in LLMs} LLMs are prone to diverse errors---including bias, reasoning failures, and generation of factually incorrect information unsupported by reliable sources. \cite{karpowicz2025fundamentalimpossibilityhallucinationcontrol} frames hallucination and imagination as mathematically identical phenomena, both emerging from a necessary violation of information conservation. Also \cite{xu2025hallucinationinevitableinnatelimitation} provides a formal learning-theoretic proof that hallucinations are unavoidable. They define a \textit{formal world} in which both the LLM and the ground-truth are computable functions, showing through classic results in computability theory, that no LLM can learn all such functions. As a consequence, hallucination is not just a practical artifact but a fundamental limitation of LLMs, valid even under idealized conditions. 
Recently, \cite{kalai2025hallucinate} showed that hallucinations come from the statistical problem of the pretraining methodology: minimizing the cross entropy naturally causes errors because it does not train the model to express uncertainty and say ``I do not know.'' \cite{kalai2025hallucinate} proposes changing the evaluation practices to not reward models for guessing, but rather to mimic the human exams that penalize only wrong answers. 

\minisection{Detecting and Mitigating LLM Hallucinations} \cite{orgad2024llms} train classifiers on the internal representations of the LLMs to predict, based on the features, the correctness of the answer. Given an LLM in a white-box setting, an input prompt, and the generated response $\hat{y}$, the classifier's task is to predict whether $\hat{y}$ is a hallucination. 
\citeauthor{orgad2024llms} suggested that LLMs may encode more factual knowledge in their latent subspaces than is revealed in their outputs. \cite{gekhman2025insideouthiddenfactualknowledge} proposed a framework for studying hidden knowledge. Finally, \cite{santilli2025revisiting} point out that uncertainty quantification in language models is often evaluated using metrics like AuROC. This shares biases between detection methods and correctness functions (e.g., length effects) that systematically distort results.
One way to mitigate hallucinations is to act at the decoding stage, where the output generation can be steered \cite{subramani2022extractinglatentsteeringvectors}. Steering vectors provide a straightforward way to control a model by adding a fixed vector to its activations ~\citep{dunefsky2025oneshotoptimizedsteeringvectors}. 
\cite{fu2025deepthinkconfidence} introduced DeepConf, a test-time method that leverages model-internal confidence signals to filter out low-quality reasoning traces during or after generation. 
\cite{kuhn2023semanticuncertaintylinguisticinvariances,fadeeva2024factcheckingoutputlargelanguage,farquhar2024semanticentropy}, and its follow-up by \cite{kossen2025semantic} in which they approximate the semantic entropy in a more efficient way. 
Constrained decoding approaches \cite{li-etal-2023-contrastive, peng2023checkfactstryagain} modify token selection policies. 
Similarly, reinforcement learning with fact-based rewards \cite{ouyang2022traininglanguagemodelsfollow} has been used to bias decoding trajectories toward verifiable outcomes. 
Incorrect answers may also be given due to an ambiguous prompt: \cite{kuhn2023clamselectiveclarificationambiguous}'s CLAM framework uses few-shot prompts to classify a question’s ambiguity and then asks the user to clarify.

\section{Background and Foundations}

\subsection{Energy-Based Models}
We give an overview of Energy-based Models (EBMs) and their use in discriminative classifiers.

\minisection{EBMs}  Energy-Based Models are a class of probabilistic models in which the probability distribution over data points $\bx$ is defined in terms of an energy function $\Ex$. The energy function, parameterized by a neural network $\net$ \citep{LeCun06EBM}, assigns a scalar energy to each configuration of $\bx$, where lower energy values correspond to higher likelihood. The resulting probability distribution is given by $p_{\net}(\bx) = \frac{\exp(-\Ex)}{Z_{\net}}$ where $Z_{\net}$ denotes the partition function (normalizing constant), defined as  $Z_{\net} = \sum_{\bx} \exp(-\Ex)$ for discrete $\bx$, or equivalently  $Z_{\net} = \int \exp(-\Ex)\, d\bx$ for continuous $\bx$. Standard neural networks are often deterministic function approximators, mapping $\bx \mapsto y$, EBMs instead define a full probability distribution over data or latent variables.

One of the strengths of EBMs is their flexibility in modeling arbitrary distributions without being tied to a specific parametric form. This flexibility comes from the fact that the energy function $E(\bx)$ can be defined in various ways. Training involves learning the parameters of the energy function such that the probability distribution $p_{\net}(\bx)$ matches the empirical distribution of the data. This is typically achieved using techniques like contrastive divergence, score matching, or maximum likelihood.

\minisection{Notation} 
Let $\vocab$ denote the vocabulary of an LLM, i.e., the set of all tokens that can be processed as input and generated at each decoding step, with size $\vert \vocab \vert = V$. 
We shorten the sequence of tokens $\seq{N}$ as $\xcal = \compactseq{N}$, and $\bx_i \in \vocab$ denotes the token in the $i$-th position along the sequence. 
We model the LLM as a function 
$\net : \mathbb{R}^{N \times V} \rightarrow \mathbb{R}^{V}$,
implemented by a transformer, or any other sequence-to-sequence mechanism. 
For a sequence $\compactseq{i}$ as input, we write $\net\big(\compactseq[]{i}\big)[k]$ to denote the predicted logit assigned to the $k$-th token class in $\vocab$ for the $i+1$ token in the sequence, as is standard in autoregressive LLM training \citep{ouyang2022traininglanguagemodelsfollow}.

\subsection{Autoregressive Large Language Models}
Generative modeling has been pursued through a variety of approaches beyond autoregression (AR). Variational Autoencoders (VAEs) \citep{kingma2013auto} learn a probabilistic latent variable model by encoding inputs into a latent space and decoding samples back to the data domain. Generative Adversarial Networks (GANs) \citep{goodfellow2014generative} frame generation as a min-max game between a generator and a discriminator. The diffusion process has been incorporated into neural nets \citep{sohl2015deep} and, more recently, Diffusion Models \citep{ho2020denoising} have emerged as a powerful class of generative models. 
While these paradigms differ in how they approximate the data distribution, AR models are special in their kind and take a more direct route by factorizing the joint probability of sequences into conditionals, making them especially suitable for language modeling. We now focus on the AR formulation that underlies most LLMs.
Textual data is segmented into a sequence of tokens $\xcal = \seq{i}$, and a language modeling objective is employed to maximize the likelihood of such data \citep{improvinglu}.
In other words, we model the joint probability of tokens in the sequence $\mathcal{X}$, through a conditional probability parameterized by $\net$:
\begin{equation}
p(\compactseq[]{i}) =  p(\bx_{i}\ |\ \compactseq[]{i-1}) \ldots p(\bx_2\ | \ \bx_1) \ p(\bx_1) = \prod\limits_{i} \underbrace{p_{\net}(\bx_{i} \ | \ \compactseq[]{i-1})}_{\text{discriminative model}}\ p_{\net}(\bx_1).
\label{likelihood-prod}
\end{equation}

What we find interesting about this factorization is that, although it seeks to attain \emph{generative modeling}, i.e., $p(\compactseq[]{i})$, it actually uses recursively \emph{discriminative classifiers}, parameterized by a transformer network $\net$, that predicts a discrete distribution of the next token $\bx_{i}$ over the vocabulary $\vocab$, given previous tokens $\compactseq[]{i-1}$. This is used to model each conditional probability.

\section{How Energy Spills in LLMs}

When predicting the token at position $i$, the conditional probability modeled by $\net$ can be decomposed using the probabilities of the sequences. As a result, the marginal term from step $i$ cancels out with the sequence probability from the decomposition at the previous step $i-1$, which means we have: 
\begin{equation}
p(\compactseq[]{i}) = \prod\limits_{i} p_{\net}(\bx_i | \compactseq[]{i-1}) 
= \prod\limits_{i} \frac{ p_{\net}( \compactseq[]{i} ) }{ p_{\net} ( \compactseq[]{i-1} ) } \implies     \dots
    \frac{ p_{\net}( \compactseq[]{i} ) }{ \underbrace{\cancel{p_{\net} ( \compactseq[]{i-1} )}}_{\text{step $i$}} } \ \frac{ \overbrace{\cancel{p_{\net}( \compactseq[]{i-1} )}}^{\text{step $i-1$}} }{ p_{\net} ( \compactseq[]{i-2} ) }
    \dots = p(\compactseq[]{i}).
\label{eq:cond-prob-exp}
\end{equation}

This indeed confirms that \cref{likelihood-prod} results in the correct formulation for language modeling, which is $p(\compactseq[]{i})$. Following the mathematics, these quantities should cancel out along the sequence, but we will now show that, in practice, \emph{this constraint is not explicitly optimized for, and we can exploit it for hallucination detection.}

\subsection{Interpreting LLMs as Energy-based models (EBMs)}
Let us continue the expansion from \cref{eq:cond-prob-exp}. Writing the conditional as the ratio between the joint distribution in the numerator and the marginal distribution in the denominator, we note that this ratio is actually implemented in LLMs as a softmax classifier that digests the embedding of the prior sentence $\compactseq[]{i-1}$ and predicts the next token $\bx_i$; thus, this chain of equality holds true. We can then apply the ``trick'' from~\cite{grathwohl2020your} as:
\begin{align}
p_{\net}(\bx_i | \compactseq[]{i-1}) =
\frac{ p_{\net}( \compactseq[]{i} ) }{p_{\net} ( \compactseq[]{i-1} ) } = 
\frac{ \exp{ \net(\compactseq[]{i-1}) \left[\texttt{id}(\bx_{i})\right] } }{ \sum\limits_{k=1}^{V} \exp{ \net(\compactseq[]{i-1}) [k] }} \ \text{where} \  \texttt{id}: \ \{0,1\}^{V} \mapsto [1,\ldots,V]. 
\label{eq:prob-next-token-practice}
\end{align}
 \texttt{id} is the map that takes as input a one-hot encoding vector $\bx_i$ for a word token at position $i$ in the text and outputs its index in the vocabulary.
A typical cross-entropy loss only optimizes with the supervision provided by the ground-truth token, through the vocabulary index $\texttt{id}(\bx_{i})$. This loss ignores all other quantities or constraints related to the complete sequence $\mathcal{X}$, i.e., it ignores all the time steps higher than $i+1$.

We can write the conditional probability of \cref{eq:prob-next-token-practice} as a ratio of two EBMs as:
\begin{equation}
\log p_{\net}(\bx_{i} | \compactseq[]{i-1}) = \log \frac{\exp(-\Eseqq{i}{\ell})}{\exp(-\Eseqq{i-1}{m})} \frac{\widetilde{Z}(\net)}{Z(\net)} = -\Eseqq{i}{\ell} + \Eseqq{i-1}{m}.
\label{eq:cond-prob-to-energy}
\end{equation}
Following~\cite{zhu2021towardsunderstandingthegenerativecapabilities}, the partition functions simplify since $\log\widetilde{Z}(\net)=\log Z(\net)$\footnote{For a formal proof, please see \cref{appendix:partitions}.}.

$E^{\ell}_{\net},~E^{m}_{\net}$ are computed from the output of the model, but with two big differences: $E^{\ell}_{\net}$ as a single \textit{logit} extracted using the \texttt{id} of the sampled token, $E^{m}_{\net}$ by \textit{marginalizing} over all \texttt{ids} in the vocabulary.

The two energies can be derived from the softmax of the logits, by connecting ~\cref{eq:cond-prob-to-energy} and \cref{eq:prob-next-token-practice}:
\begin{align}
-\log p_{\net}(\bx_{i}\ |\ \compactseq[]{i-1}) &= -\log  \Bigg( \frac{\exp(\net(\compactseq[]{i-1})[\texttt{id}(\bx_{i})])}{\sum\limits_{k} \exp(\net(\compactseq[]{i-1})[k])} \Bigg ) = \\
& = \underbrace{-\net(\compactseq[]{i-1}) \left[\texttt{id}(\bx_{i})\right]}_{\Eseqq{i}{\ell}} + \underbrace{\log \sum\limits_{k=1}^{V} \exp{\net(\compactseq[]{i-1}) [k]}}_{-\Eseqq{i-1}{m} }
\label{eq:p-next-token}
\end{align}
where $\net(\compactseq[]{i-1})$ produces the logits over the entire vocabulary $\mathcal{V}$, and $\texttt{id}(\bx_i)$ allows us to extract the logit of the sampled token at decoding step $i$.

We can think of $\Eseqq{i}{\ell}$ as the energy of the sampled tokens $\compactseq{i}$, and $\Eseqq{i-1}{m}$ as the energy $\Eseq{i}$, marginalized over all possible $\bx_{i}$. Considering the decoding at step $i$ in \cref{eq:cond-prob-to-energy}, we get:
\begin{equation}
\Eseqq{i}{\ell} = -\net(\compactseq[]{i-1})[\texttt{id}(\bx_{i})], \quad \Eseqq{i-1}{m} = - \log \sum\limits_{k=1}^{V} \exp{\net(\compactseq[]{i-1}) [k]}.
\end{equation}
Using the chain rule and \cref{eq:p-next-token}, we can write the negative log-likelihood in terms of energies as:
\[
-\log p(\compactseq[]{N}) = -\log \prod\limits_{i} p_{\net}(\bx_i | \compactseq[]{i-1}) = \sum_i \Eseqq{i}{\ell} - \Eseqq{i-1}{m} \]

without considering the base case $p_{\net}(\bx_1)$. Now, if we develop the above equation as done for \cref{eq:cond-prob-exp}, we write the total energy of a sequence of length $N$ as $\Eseq{N}$. Observe that the two energies, not interacting at the same step but at steps $i$ and $i-1$, \tbf{should be equal, but they are measured in the LLM at different generation steps and from different components.}
\resizebox{\textwidth}{!}{$
\Eseq{N}=\sum_{i=1}^{N-1} \Eseqq{i+1}{\ell} - \Eseqq{i}{m} \;=\;
\ldots\;
\overbrace{ \Eseqq{i+1}{\ell} \ \lefteqn{\underbrace{\phantom{-\Eseq{i}+\Eseq{i}}}_{\Delta\Eseq{i}}} -\Eseqq{i}{m}}^{\text{timestep}~i+1}
+ \overbrace{\Eseqq{i}{\ell} - \Eseqq{i-1}{m}}^{\text{timestep}~i}\; \ldots
$
}
At timestep $i+1$, first  $-\Eseqq{i}{m}$ is measured, taking the denominator in the softmax as in the right part of \cref{eq:p-next-token}, whereas at timestep $i$, the second $\Eseqq{i}{\ell}$ is taken, reading the logit in the softmax, left part of \cref{eq:p-next-token}. We thus define the discrepancy between the two quantities as \tbf{spilled energy}:
\begin{emphabox}
\begin{definition}[Spilled Energy $\Delta\Eseq{i}$] The spilled energy in an LLM is the difference between two energies that, in principle, should be equal, but given that they are measured i) at different time steps ii) in different components, could be different.
\begin{equation}
\Delta\Eseq{i} \triangleq  -\Eseqq{i}{m} + \Eseqq{i}{\ell} =  \underbrace{-\log \sum\nolimits_{k} \exp(\net(\compactseq[]{i})[k])}_{\text{timestep}~ i+1} + \underbrace{\net(\compactseq[]{i-1})[\texttt{id}(\bx_{i})]}_{\text{timestep}~i}
\label{def:spilled-energy}
\end{equation}
\end{definition}
\end{emphabox}
Since both terms on the right side should be equal to $\Eseq{i}$, delta values should always be zero when we are correctly modeling the energy at timestep $i$. A shorter explanation for why spilled energy needs to be zero is given in \cref{sec:small-prof}.

\subsection{Detecting hallucinations with spilled energy}
\label{sec:detect-hallucinations}
EBMs have previously been used to assess neural network credibility~\citep{energyOOD}, and calibration for LLMs has been explored by the Anthropic team~\citep{kadavath2022language}. 
However, dominant training-free baselines such as logits or ``$p(\text{true})$'' remain weak. 
We likewise adopt a training-free approach, but rely on \cref{def:spilled-energy} and its variants as discriminants.

We feed the prompt $\seq{i-1}$ to the LLM $\net$ and obtain the completion $\seqq{N}{i}$. 
Following \citet{orgad2024llms}, we focus on the ``exact answer'' tokens—those in $[i+1, N]$ that contain the precise answer (e.g., \texttt{Rome} in \cref{fig:teaser}), denoted $[u,w] \subseteq [i+1, N]$. For instance, it would be the tokens associated with \texttt{Rome} in the question in \cref{fig:teaser}. 
We identify this span by prompting the LLM for a brief answer. 
When the answer spans multiple tokens, we apply a pooling strategy, which we ablate in \cref{sec:expt}.
We propose measuring two values that correlate well with hallucinations:
\begin{enumerate}
    \item Marginal energy $\Eseqq{i}{m}$;
    \item Spilled energy $\Delta \Eseqq{i}{}$ by definition of \cref{def:spilled-energy}.
\end{enumerate}
We also attempt to combine the two metrics into scaled spilled energy $\Delta E_{s}$, where the spilled energy is multiplied by the absolute value of the marginal energy as
        $\Delta E_{s}(\compactseq[]{i}) = \lvert\Eseqq{i}{m}\rvert\Delta \Eseqq{i}{}$.
The metrics proposed here are independent, new for LLMs, and can all be tested efficiently. These measures can be computed over the full sequence, but for error detection, as discussed in \cref{sec:limitations}, we must extract the values in the localized exact interval $[u,w]$ to avoid false positives. Note that $\Eseqq{i}{\ell}$ is the classic baseline which in literature is referred to as ``logits'' or ``logits confidence''.

\section{Experiments}\label{sec:expt}

To evaluate spilled energy, we consider two complementary settings. 
First, a controlled synthetic environment, where we generate both correct and incorrect multi-digit arithmetic solutions. Second, established real-world benchmarks, where errors arise naturally across diverse reasoning and comprehension tasks. Together, these experiments test whether insights from the clean synthetic setup transfer to the complexity of open-domain language understanding.

\subsection{Spilled Energy under Synthetic Arithmetic}
\label{sec:synth-arithmetic}
\minisection{Experimental Setting}
We first evaluate spilled energy in a controlled setting: multi-digit arithmetic problems with more than 14 digits. For each instance, we generate both correct and incorrect solutions. We tested three different LLMs: Llama-3 8B (\citeauthor{dubey2024llama}), Qwen-3 8B (\citeauthor{qwen2024qwen3}), and Mistral-7B-Instruct v0.3 (\citeauthor{jiang2023mistral7b}).
Incorrect solutions are obtained by introducing random numerical errors of varying magnitude. 
Specifically, we define three error ranges that differ in their difficulty of detection:
\begin{itemize}[itemsep=1pt, leftmargin=*]
    \item \textbf{Easy}: random offset in the range $[1000, 10000]$, which are typically easier to identify.
    \item \textbf{Medium}: random offset in the range $[100, 1000]$, where detection requires closer inspection.
    \item \textbf{Hard}: random offset in $[1, 10]$, much harder to detect since they appear plausible at first glance.  
\end{itemize}

This design allows us to systematically probe whether spilled energy can distinguish between correct and incorrect generations across different levels of error subtlety.

\begin{figure}[htb]
\centering
\vspace{0.8em}
{LLama-3-8B-Instruct}\\[0.8em]
\begin{subfigure}{0.24\textwidth}
  \tikz[remember picture,baseline] \node (tripletNW) {};
  \includegraphics[width=\linewidth]{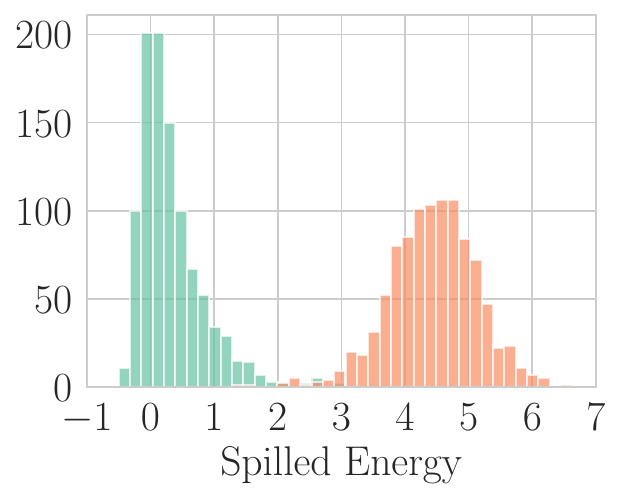}
  \caption{Easy}
  \label{fig:synth-math-llama-inst-large-range}
\end{subfigure}\hfill
\begin{subfigure}{0.24\textwidth}
  \tikz[remember picture,baseline] \node (tripletN) {};
  \includegraphics[width=\linewidth]{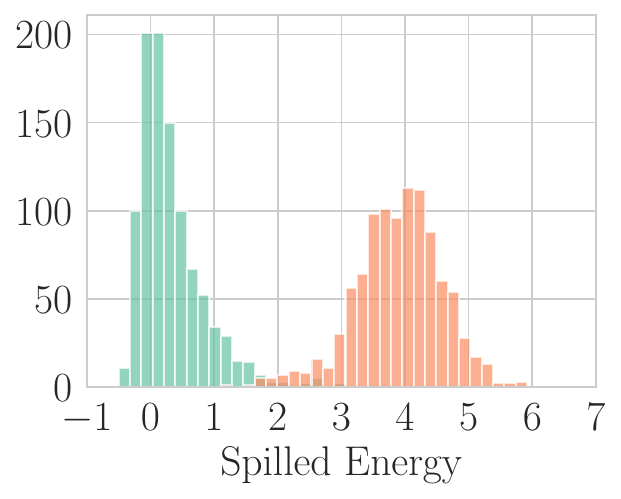}
  \caption{Medium}
  \label{fig:synth-math-llama-inst-mid-range}
\end{subfigure}\hfill
\begin{subfigure}{0.24\textwidth}
  \tikz[remember picture,baseline] \node (tripletNE) {};
  \includegraphics[width=\linewidth]{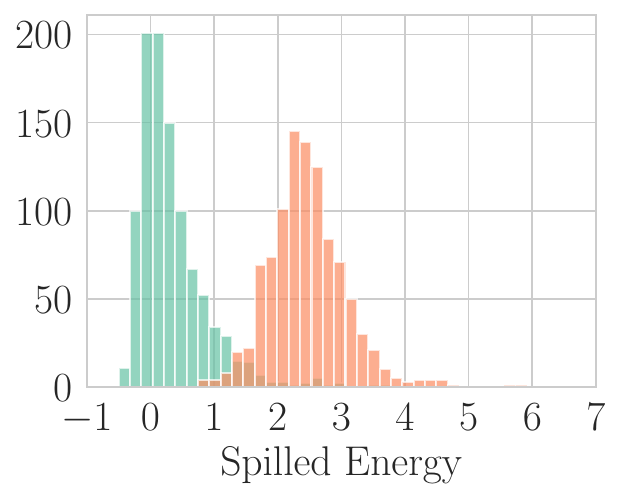}
  \caption{Hard}
  \label{fig:synth-math-llama-inst-small-range}
\end{subfigure}
\begin{subfigure}{0.24\textwidth}
  \tikz[remember picture,baseline] \node (rocAnchor) {};
  \includegraphics[width=\linewidth]{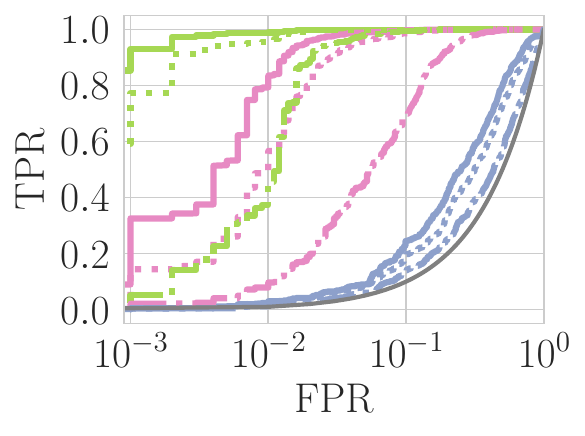}
  \caption{ROC}
  \label{fig:synth-math-llama-inst-roc}
\end{subfigure}

\begin{tikzpicture}[remember picture,overlay]
  \path let
    \p1 = (tripletNW),  
    \p2 = (tripletNE)   
  in node[anchor=center, yshift=16pt, xshift=-2pt,
          fill=white, draw=black, rounded corners=1pt, inner sep=3pt]
     at ($(\x1,\y1)!0.5!(\x2,\y2)$) 
     {\footnotesize
       \legendcolor{green}{Correct}\quad
       \legendcolor{orange}{Incorrect}\quad
     };
\end{tikzpicture}

\begin{tikzpicture}[remember picture,overlay]
  \node[anchor=north, xshift=13pt, yshift=37pt,
        fill=white, draw=black, rounded corners=1pt, inner sep=4pt]
       at (rocAnchor.north) 
       {\footnotesize
         \begin{tabular}{@{}l l@{}}
           \legendcolor{light-green}{Spilled Energy} &
           \legendline{black}{solid}{Easy} \\
           \legendcolor{pink}{Logit Energy} &
           \legendline{black}{dash pattern=on 3pt off 2pt}{Medium} \\
           \legendcolor{purple}{Marginal Energy} &
           \legendline{black}{densely dotted}{Hard} \\
         \end{tabular}
       };
\end{tikzpicture}

\vspace{-1.5em}

{Qwen-3 8B}\\[0.3em]
\begin{subfigure}{0.24\textwidth}
  \includegraphics[width=\linewidth]{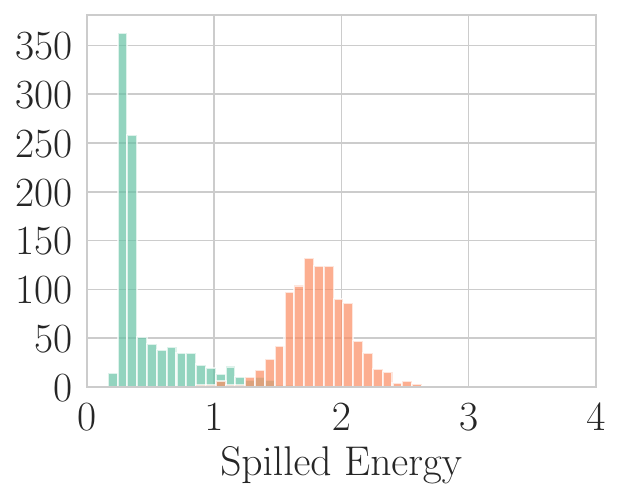}
  \caption{Easy} 
  \label{fig:synth-math-qwen-large-range}
\end{subfigure}
\begin{subfigure}{0.24\textwidth}
  \includegraphics[width=\linewidth]{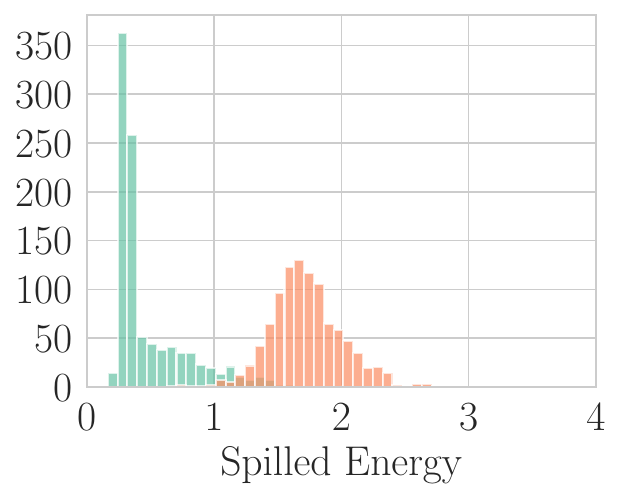}
  \caption{Medium} 
  \label{fig:synth-math-qwen-mid-range}
\end{subfigure}
\begin{subfigure}{0.24\textwidth}
  \includegraphics[width=\linewidth]{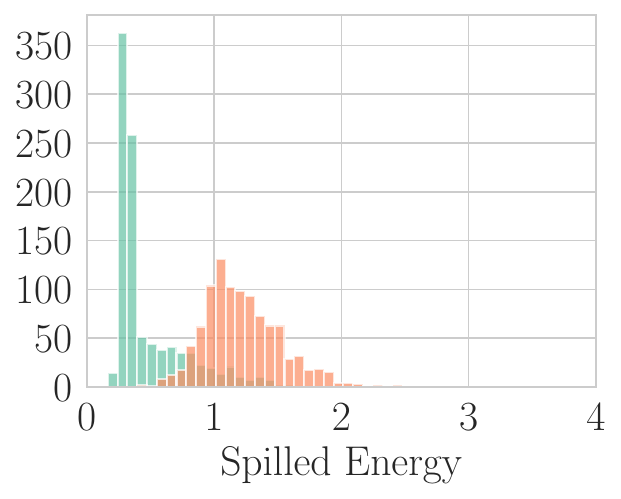}
  \caption{Hard} 
  \label{fig:synth-math-qwen-small-range}
\end{subfigure}
\begin{subfigure}{0.25\textwidth}
  \includegraphics[width=\linewidth]{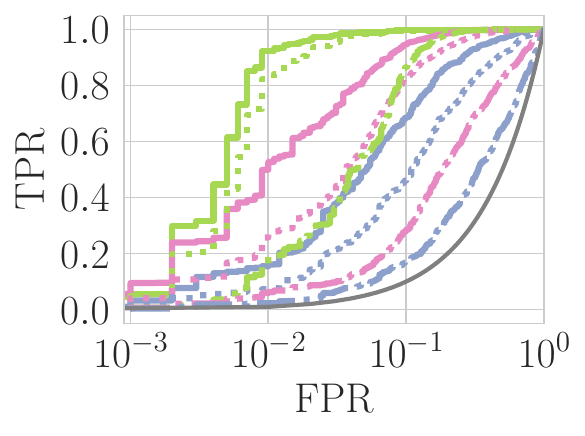}
  \caption{ROC}
  \label{fig:synth-math-qwen-roc}
\end{subfigure}

\caption{Histograms of Spilled Energy values across models (rows) on Math Sums with different error ranges in the answer (columns, decreasing range left to right, making it harder to detect errors). All sums are performed on 13-digit integers. In the fourth column, we show ROC curves for Hallucination Detection across the error ranges (colors) and methods (line styles).}
\label{fig:spilled-synth-arithmetic}
\end{figure}
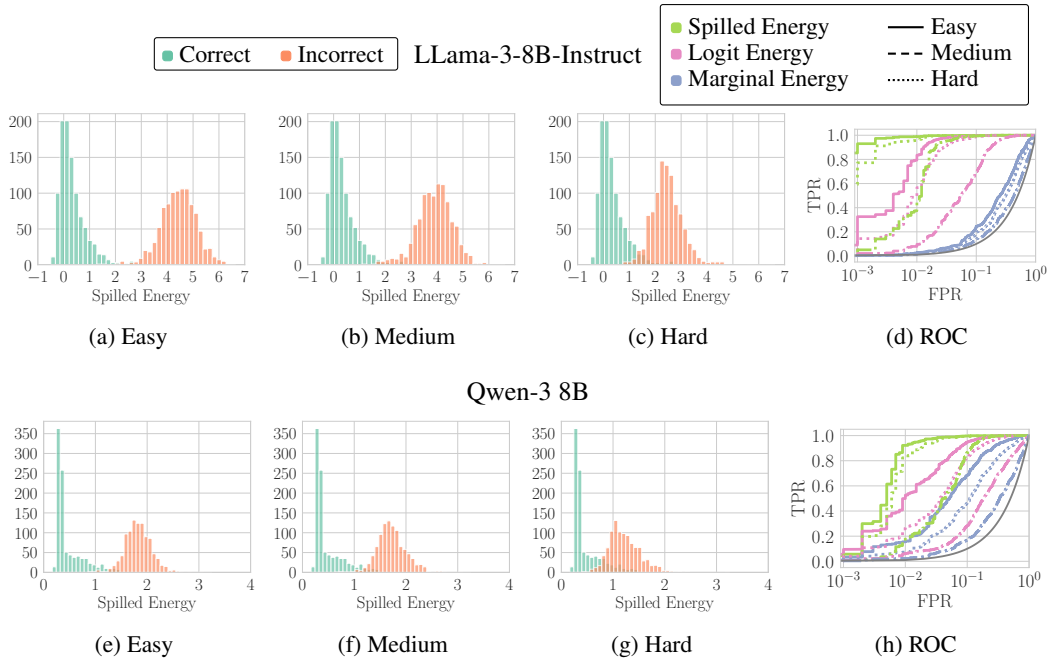

\minisection{Results}
We observe that spilled energy values separate correct from incorrect solutions with high reliability across all error ranges and across all LLMs.
In particular, spilled energy consistently assigns lower values to correct generations and higher values to incorrect ones, producing a clear margin of separation. 
Compared to standard baselines such as \emph{logits}, spilled energy achieves superior discriminative power, especially for errors in the more challenging range $[1, 10]$, see \cref{fig:spilled-synth-arithmetic}. We offer more results in \cref{fig:spilled-synth-arithmetic-add}. Larger, better-detailed ROC and histograms are in \cref{fig:models-vs-datasets,fig:models-vs-datasets-hist} respectively.

\begin{figure}[tbh]
    \centering
    
    \begin{subfigure}{0.48\textwidth}
        \centering
        \includegraphics[width=\linewidth]{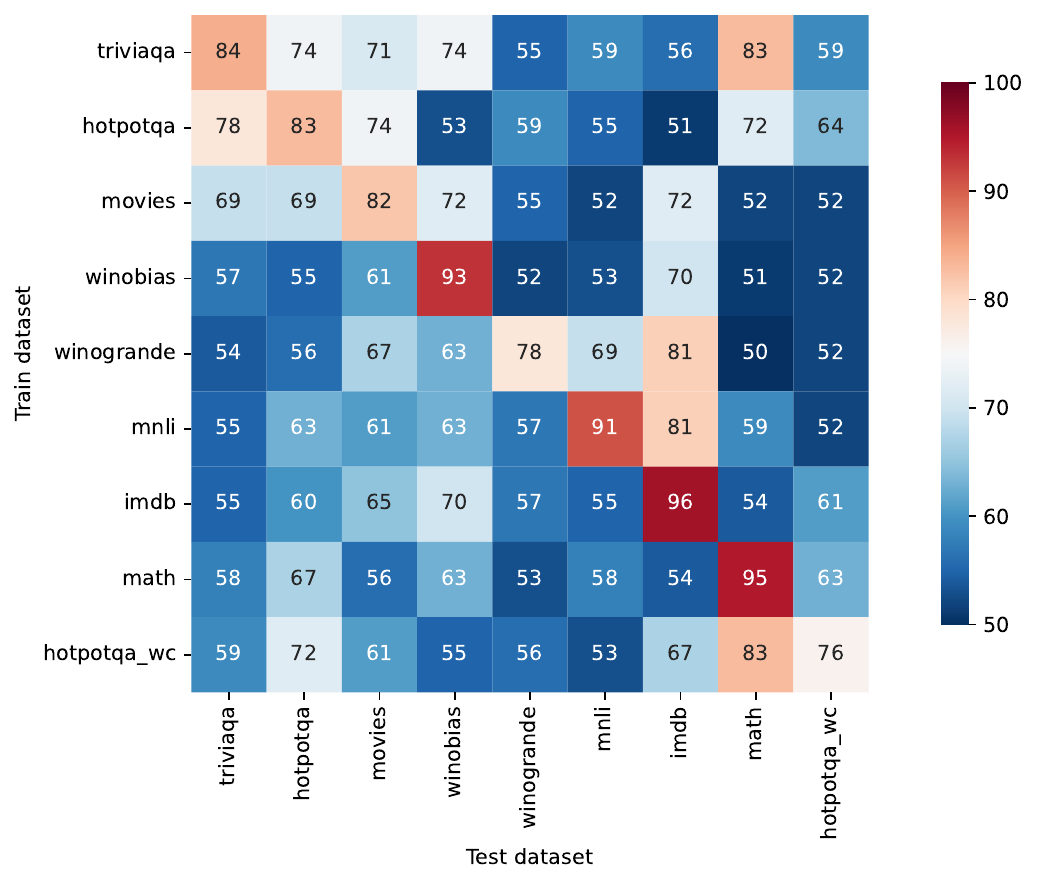}
        \caption{Results by \citeauthor{orgad2024llms}}
        \label{fig:orgadHeatMapsLlamaInstr}
    \end{subfigure}
    \hfill
    \begin{subfigure}{0.48\textwidth}
        \centering
        \includegraphics[width=\linewidth]{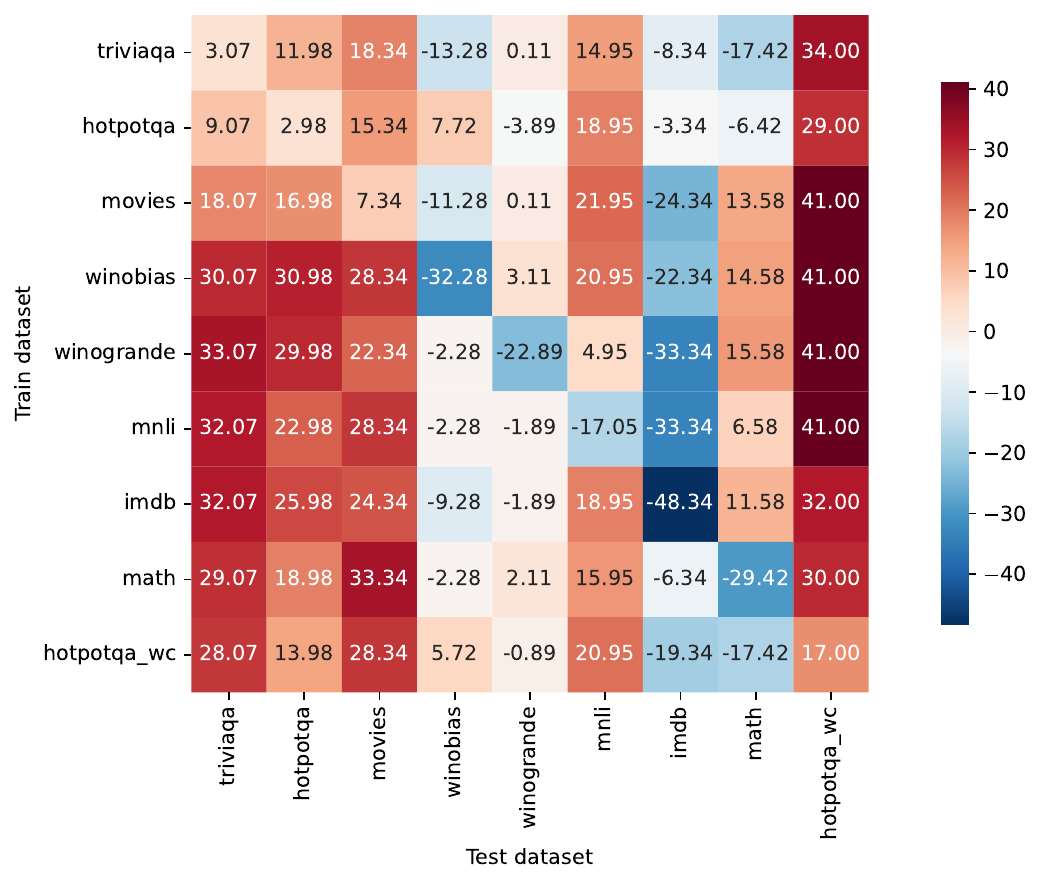}
        \caption{Spilled Energy Improvement over \citeauthor{orgad2024llms}}
        \label{fig:diffOrgadLlamaInstr}
    \end{subfigure}

    \caption{(a) AuROC performance as percentages of probing classifiers on exact answer tokens by \citeauthor{orgad2024llms} for LlaMA-3-Instruct. (b) depicts the performance difference between our Spilled $\Delta E$  with Min pooling and theirs. Positive values indicate cases where Spilled $\Delta E$ outperforms \citeauthor{orgad2024llms}. This comparison highlights the generalization capabilities of our method, compared to probing classifiers. Legend: low performance \protect\tikz[baseline=-0ex]\protect\shade[left color=blue!100,middle color=white!100,right color=red!100] (0,0) rectangle (2,0.25); high performance.} 
    \label{fig:confusion_heatmap}
\end{figure}

\subsection{Cross-dataset Results in Real-World Benchmarks}\label{sec:spilled-real-world}
\minisection{Experimental Setting}
We evaluate our methods on a diverse set of established NLP benchmarks, including Math (\citeauthor{hendrycksmath2021}), TriviaQA (\citeauthor{joshi2017triviaqa}), HotpotQA (\citeauthor{yang2018hotpotqa}), Winogrande (\citeauthor{sakaguchi2021winogrande}), Winobias (\citeauthor{winobias}), Movies (\citeauthor{orgad2024llms}), MNLI (\citeauthor{mnli}) and IMDB (\citeauthor{imdb_dataset}). 
These datasets span a wide range of reasoning and error-detection tasks, allowing us to test whether the patterns observed in the synthetic arithmetic setting extend to real-world, open-domain scenarios. Here too, we evaluate multiple LLMs that are either instruction-aligned or not aligned, such as LLaMA-3 (\citeauthor{dubey2024llama}), and Mistral (\citeauthor{jiang2023mistral7b}).
As emphasized by \citeauthor{orgad2024llms}, it is essential to first localize the tokens most relevant to the final answer before applying error detection. Since exact answer tokens may consist of multiple tokens, we further adopt a pooling strategy across the localized span to obtain a final score per sentence.
We compare spilled and marginal energy against baselines such as the probing classifiers of \citeauthor{orgad2024llms}, logit confidence of \citeauthor{varshney2023stitch} and $p(\text{true})$ of \citeauthor{languagemodelsmostlyknow}. 

\begin{table}[tb]
\caption{Hallucination detection performance, in terms of AuROC, across nine benchmarks and four different LLMs. We measure the generalization across all tasks by computing the average.\vspace{-10pt}}
\label{tab:detection_NO_FLIP_STD_Corretta}
\centering
\setlength{\tabcolsep}{2pt}
\resizebox{\textwidth}{!}{
\begin{tabular}{lccccccccccb}
\toprule
 & Pool & HotpotQA & HotpotQA-WC & IMDB & Math & MNLI & Movies & TriviaQA & Winobias & Winogrande & Average \\
\midrule 
& \multicolumn {11}{c}{LLaMA-Instruct \cite{dubey2024llama}} \\\midrule 

$p(\text{true})$ & --- & 58.31\stdev{0.32} & 51.66\stdev{1.05} & 50.72\stdev{1.20} & 49.53\stdev{2.16} & 52.33\stdev{0.98} & 59.30\stdev{0.85} & 45.99\stdev{0.51} & 45.47\stdev{1.58} & 48.33\stdev{0.68} & 51.29\stdev{04.86} \\
{\footnotesize \citeauthor{orgad2024llms}} & Mean & 66.56\stdev{9.10} & 59.00\stdev{8.14} & \textit{69.78}\stdev{14.76} & \textit{66.56}\stdev{17.04} & 60.56\stdev{12.53} & 66.44\stdev{8.06} & 63.22\stdev{11.11} & \textbf{67.33}\stdev{11.97} & \textbf{58.00}\stdev{7.79} & 64.16\stdev{03.90} \\
Logit $E^\ell$ & Max & 72.85\stdev{2.12} & 91.11\stdev{1.52} & 42.08\stdev{5.07} & 57.81\stdev{3.82} & 25.52\stdev{3.00} & 43.97\stdev{1.38} & 68.89\stdev{1.96} & 39.95\stdev{2.41} & 49.40\stdev{2.16} & 54.62\stdev{18.97} \\
\midrule 
Marginal $E^m$ & Max & \textit{76.72}\stdev{1.38} & 30.74\stdev{3.45} & \textbf{85.63}\stdev{2.39} & 27.08\stdev{5.06} & \textbf{89.90}\stdev{1.25} & \textbf{96.17}\stdev{0.63} & \textit{80.13}\stdev{1.87} & 57.67\stdev{2.94} & 47.47\stdev{1.83} & \textit{65.72}\stdev{24.39} \\
Marginal $E^m$ & Min & 75.91\stdev{1.62} & \textbf{97.57}\stdev{0.75} & 14.37\stdev{2.39} & \textbf{70.55}\stdev{2.43} & 61.21\stdev{3.24} & 72.21\stdev{1.60} & 73.38\stdev{1.86} & 47.19\stdev{2.71} & 53.98\stdev{2.30} & 62.93\stdev{21.89} \\
Spilled $\Delta E_{s}$ & Max & 53.65\stdev{1.40} & 36.28\stdev{2.99} & 55.80\stdev{4.32} & 35.44\stdev{3.41} & 58.81\stdev{2.58} & 70.30\stdev{1.49} & 48.70\stdev{2.44} & 36.53\stdev{2.98} & 44.32\stdev{1.70} & 48.87\stdev{11.26} \\
Spilled $\Delta E$ & Min & \textbf{85.98}\stdev{1.09} & \textit{93.00}\stdev{1.61} & 47.66\stdev{4.06} & 65.58\stdev{3.02} & \textit{73.95}\stdev{1.97} & \textit{89.34}\stdev{1.04} & \textbf{87.07}\stdev{1.33} & \textit{60.72}\stdev{2.74} & \textit{55.11}\stdev{2.05} & \textbf{73.16}\stdev{15.64} \\

\midrule 
& \multicolumn {11}{c}{LLaMA \cite{dubey2024llama}} \\\midrule 

$p(\text{true})$ & --- & 52.83\stdev{0.71} & 49.33\stdev{0.86} & 52.30\stdev{0.58} & 58.63\stdev{1.26} & 53.78\stdev{0.70} & 60.76\stdev{0.69} & 62.94\stdev{0.51} & 50.02\stdev{1.24} & 53.47\stdev{0.54} & 54.90\stdev{04.77} \\
{\footnotesize \citeauthor{orgad2024llms}} & Mean & 61.22\stdev{9.95} & 56.78\stdev{8.70} & \textbf{72.67}\stdev{13.91} & \textit{69.67}\stdev{15.07} & 60.33\stdev{13.77} & \textit{64.00}\stdev{8.40} & 66.44\stdev{8.20} & \textbf{60.89}\stdev{12.60} & \textbf{53.56}\stdev{4.36} & \textit{62.84}\stdev{05.71} \\ 
Logit $E^\ell$ & Max & 53.47\stdev{2.13} & 49.02\stdev{1.79} & 48.27\stdev{1.32} & 57.38\stdev{6.09} & 91.76\stdev{0.91} & 57.42\stdev{1.43} & 52.77\stdev{2.58} & 50.74\stdev{1.51} & 51.17\stdev{1.83} & 56.89\stdev{12.70} \\
\midrule 
Marginal $E^m$ & Max & \textit{78.00}\stdev{1.30} & \textit{76.90}\stdev{1.09} & 48.29\stdev{1.16} & 68.77\stdev{8.33} & 10.93\stdev{1.42} & \textbf{80.70}\stdev{1.98} & 67.49\stdev{1.69} & 51.91\stdev{2.32} & \textit{51.28}\stdev{2.47} & 59.36\stdev{20.69} \\
Marginal $E^m$ & Min & 58.39\stdev{2.79} & 59.20\stdev{1.95} & \textit{51.71}\stdev{1.16} & 34.13\stdev{8.78} & \textit{97.42}\stdev{0.51} & 50.37\stdev{2.43} & \textit{69.88}\stdev{1.40} & 49.05\stdev{2.20} & 49.00\stdev{2.30} & 57.68\stdev{16.75} \\
Spilled $\Delta E_{s}$ & Min & 77.75\stdev{1.52} & 79.44\stdev{2.05} & 43.39\stdev{1.82} & \textit{72.87}\stdev{6.10} & \textbf{99.97}\stdev{0.08} & 61.56\stdev{2.95} & 77.55\stdev{1.62} & \textit{52.34}\stdev{2.57} & 48.17\stdev{1.62} & 68.12\stdev{17.15} \\
Spilled $\Delta E$ & Min & \textbf{79.04}\stdev{1.78} & \textbf{80.83}\stdev{1.87} & 43.22\stdev{1.67} & \textbf{74.36}\stdev{5.54} & \textbf{99.97}\stdev{0.08} & 61.97\stdev{2.81} & \textbf{78.54}\stdev{1.57} & \textit{52.11}\stdev{2.58} & 48.21\stdev{1.62} & \textbf{68.69}\stdev{17.48} \\

\midrule 
& \multicolumn {11}{c}{Mistral-Instruct \cite{jiang2023mistral7b}} \\\midrule 

$p(\text{true})$ & --- & 56.67\stdev{0.80} & 53.41\stdev{0.68} & 48.84\stdev{0.78} & 51.63\stdev{1.29} & 54.93\stdev{0.53} & 60.64\stdev{0.47} & 63.59\stdev{0.57} & 56.34\stdev{0.92} & 56.92\stdev{0.57} & 55.88\stdev{04.45} \\
{\footnotesize \citeauthor{orgad2024llms}} & Mean & 64.78\stdev{10.56} & 56.78\stdev{7.95} & \textbf{82.67}\stdev{11.63} & \textbf{68.78}\stdev{11.43} & 64.22\stdev{12.12} & 64.89\stdev{11.55} & 65.44\stdev{12.10} & \textbf{61.00}\stdev{12.23} & \textbf{61.44}\stdev{11.31} & 65.56\stdev{06.84} \\ 
Logit $E^\ell$ & Max & 77.24\stdev{1.66} & 83.84\stdev{1.66} & 22.28\stdev{2.54} & 57.67\stdev{3.29} & 78.98\stdev{1.58} & 76.89\stdev{1.49} & 80.35\stdev{1.88} & 45.53\stdev{2.60} & 48.17\stdev{1.97} & 63.44\stdev{19.99} \\
\midrule 
Marginal $E^m$ & Max & 64.63\stdev{1.97} & 33.42\stdev{1.90} & \textit{81.33}\stdev{2.32} & 26.52\stdev{2.28} & 17.62\stdev{1.20} & \textit{86.60}\stdev{1.20} & 65.46\stdev{2.25} & \textit{56.41}\stdev{4.44} & \textit{51.14}\stdev{1.71} & 53.68\stdev{22.53} \\
Marginal $E^m$ & Min & \textit{87.58}\stdev{1.35} & \textbf{97.94}\stdev{0.62} & 18.67\stdev{2.27} & \textit{67.58}\stdev{3.37} & \textbf{97.96}\stdev{0.55} & 84.90\stdev{1.37} & \textit{87.75}\stdev{1.73} & 49.19\stdev{3.97} & 48.49\stdev{1.86} & \textit{71.12}\stdev{25.68} \\
Spilled $\Delta E_{s}$ & Max & 49.13\stdev{2.50} & 36.37\stdev{2.40} & 46.45\stdev{2.56} & 29.05\stdev{2.57} & 53.79\stdev{1.55} & 55.24\stdev{2.17} & 46.73\stdev{1.98} & 53.30\stdev{3.66} & 51.20\stdev{1.84} & 46.81\stdev{08.24} \\
Spilled $\Delta E$ & Min & \textbf{91.12}\stdev{1.10} & \textit{97.47}\stdev{0.78} & 59.77\stdev{2.57} & 66.63\stdev{3.46} & \textit{95.95}\stdev{0.83} & \textbf{94.99}\stdev{0.93} & \textbf{91.75}\stdev{1.01} & 50.74\stdev{3.15} & 49.00\stdev{1.92} & \textbf{77.49}\stdev{19.42} \\

\midrule 
& \multicolumn {11}{c}{Mistral \cite{jiang2023mistral7b}} \\\midrule 

$p(\text{true})$ & --- & 54.21\stdev{0.76} & 51.68\stdev{0.76} & 50.40\stdev{0.50} & 45.86\stdev{2.05} & 51.94\stdev{0.50} & 49.12\stdev{0.63} & 58.00\stdev{0.67} & 53.76\stdev{1.17} & 47.29\stdev{0.55} & 51.36\stdev{03.73} \\
{\footnotesize \citeauthor{orgad2024llms}} & Mean & 61.78\stdev{9.27} & 57.44\stdev{6.95} & \textbf{76.22}\stdev{12.82} & 65.78\stdev{15.27} & 56.67\stdev{11.83} & 64.22\stdev{8.91} & 64.33\stdev{10.40} & \textbf{58.00}\stdev{12.29} & \textbf{54.56}\stdev{4.36} & 62.11\stdev{06.21} \\ 
Logit $E^\ell$ & Max & 49.54\stdev{1.42} & 52.47\stdev{1.61} & 32.72\stdev{2.89} & 57.21\stdev{3.89} & 92.49\stdev{1.15} & 30.52\stdev{2.00} & 39.73\stdev{2.03} & 46.53\stdev{3.80} & 44.41\stdev{2.42} & 49.51\stdev{17.28} \\
\midrule 
Marginal $E^m$ & Max & 83.57\stdev{1.13} & \textit{86.83}\stdev{1.70} & 45.31\stdev{2.49} & 62.26\stdev{4.29} & 96.03\stdev{0.83} & \textbf{99.27}\stdev{0.24} & \textbf{92.26}\stdev{1.31} & 51.31\stdev{3.35} & \textit{54.49}\stdev{2.48} & \textit{74.59}\stdev{19.91} \\
Marginal $E^m$ & Min & \textbf{87.52}\stdev{1.31} & \textbf{90.91}\stdev{1.58} & 54.69\stdev{2.49} & \textbf{86.21}\stdev{1.96} & \textbf{98.80}\stdev{0.35} & \textit{94.41}\stdev{0.62} & \textit{83.66}\stdev{2.16} & \textit{52.15}\stdev{1.74} & 46.37\stdev{2.02} & \textbf{77.19}\stdev{19.05} \\
Spilled $\Delta E_{s}$ & Max & 60.54\stdev{1.81} & 60.18\stdev{1.84} & 43.47\stdev{2.76} & 71.93\stdev{3.62} & 45.94\stdev{2.40} & 78.84\stdev{1.53} & 67.92\stdev{1.32} & 57.24\stdev{3.72} & 51.88\stdev{1.90} & 59.77\stdev{11.08} \\
Spilled $\Delta E$ & Min & \textit{84.24}\stdev{1.18} & 83.74\stdev{1.41} & \textit{57.43}\stdev{2.99} & \textit{78.26}\stdev{2.93} & \textit{96.69}\stdev{0.62} & 84.47\stdev{1.17} & 81.27\stdev{1.83} & 50.62\stdev{1.72} & 48.72\stdev{1.75} & 73.94\stdev{16.18} \\
\bottomrule \vspace{-35pt}
\end{tabular}
}
\end{table}

\minisection{Ablation of the exact answer token} We provide an ablation experiment on the impact of selecting the exact answer tokens. \cref{tab:ablation-exact} reports average AuROC over 9 benchmarks and 4 LLMs with the exact answer, along with another column that offers the improvement provided by using the exact answer. Like prior work, we confirm that searching for the exact answer provides a notable boost: the improvement is very pronounced ($\sim 24\%$) for spilled and marginal energy, while the logit baseline receives a modest increase of $9\%$.

\minisection{Cross-dataset results} We next evaluate in the more general setting of cross-dataset transfer, which better reflects real-world usage. 
For methods requiring training, we report the average performance on each dataset when trained separately on each of the other datasets (e.g., performance on IMDB is the average accuracy of classifiers trained on each of the other nine datasets).
\cref{fig:confusion_heatmap} shows a confusion matrix of cross-dataset performance, where the rows represent the training dataset and the columns represent the testing dataset, and where red indicates good performance and blue indicates low accuracy. The model tested is LlaMA-3-Instruct.
\cref{fig:orgadHeatMapsLlamaInstr} shows that probing classifiers, as soon as they go out-of-distribution from the training dataset, perform only marginally better than random guessing. The sharp drop observed in the off-diagonal elements supports our premise that this standard, in-distribution setup significantly overestimates the utility of trained probes for broad LLM deployment. Meanwhile, \cref{fig:diffOrgadLlamaInstr} displays the improvement of  Spilled $\Delta E$ over the probing classifier, where a positive red result means improvement of our method. Ours exhibits greater performance across most datasets without requiring training. The generalization is proved with a strong increment over the off-diagonal. Moreover, in some cases, such as TriviaQA, HotpotQA, and Movies, we have improvements \emph{even on the diagonal}. Additional confusion matrices are available in \cref{sec:appendix-cross}.

\cref{tab:detection_NO_FLIP_STD_Corretta} summarizes results across nine benchmarks. The result reported in each cell is the average of the accuracies of \cref{fig:orgadHeatMapsLlamaInstr} within a column.
Spilled energy consistently outperforms \emph{logit} confidence, and substantially surpasses the probing classifiers of \citet{orgad2024llms}. 
While this latter performs well when trained and tested on the same dataset, their performance drops sharply under cross-dataset evaluation, as reflected in their higher standard deviations. 
By contrast, ours requires no training and generalizes robustly across diverse benchmarks.
We observe that instruction-tuned models tend to amplify the margin by which spilled energy outperforms other methods, whereas on non-aligned Mistral, spilled energy may rank slightly behind marginal energy. 
We also compare pooling strategies and find that $\min$ pooling yields the best overall performance across methods.
\cref{tab:detection-gemma} shows our method generalizes to Gemma over different LLM size, 1B and 4B.

 \begin{wraptable}{r}{0.375\textwidth}   
\vspace{-10pt}
\centering
\resizebox{\linewidth}{!}{
\setlength{\tabcolsep}{3pt}
    \begin{tabular}{lccc}
    \toprule
     & Pool & Average~\%& Exact  \\
     &  &  w/ exact & answer \\
     &  &  answer & increase \\
    \midrule

Logit $E^\ell$ & Max & 56.12 & +9.23 \\ \citeauthor{orgad2024llms} & Mean & 63.67 & -- \\
\midrule
Marginal $E^m$ & Min & 67.23 & +20,02\\
Marginal $E^m$ & Max & 63.34 & +3,62 \\
Spilled $\Delta E$ & Min & \textbf{73.32} & \textbf{+24.06}\\
\bottomrule

    \end{tabular}
}

\caption{Improvements in AuROC with the exact answer. Average across 4 LLMs and 9 benchmarks.}
\label{tab:ablation-exact}
\vspace{-10pt}
\end{wraptable}

\minisection{Impact of Instruction Tuning} 
\begin{table}[tb]
\caption{Hallucination detection performance on the Gemma Model Instruct for different parameters of the model, 1B and 4B.\vspace{-10pt}}
   \label{tab:detection-gemma}
\centering
\setlength{\tabcolsep}{2pt}
\resizebox{\textwidth}{!}{
\begin{tabular}{lccccccccccb}
\toprule
& Pool & IMBD & Movies & TriviaQA & Winogrande & Winobias & MNLI & Math & HotpotQA & HotpotQA-WC & Average \\
\midrule 
& \multicolumn {11}{c}{Gemma-Instruct 4B \cite{gemma3}} \\\midrule 
Logit $E^\ell$ & Max & 50.09\stdev{0.45} & 60.88\stdev{3.96} & 53.95\stdev{2.10} & 49.77\stdev{0.15} & \textbf{54.43\stdev{2.80}} & 27.00\stdev{2.16} & \textit{78.64}\stdev{3.47} & 62.84\stdev{1.97} & 64.49\stdev{2.02} & 55.79\stdev{13.24} \\\midrule

Marginal $E^m$ & Max & 49.14\stdev{2.70} & \textit{83.02}\stdev{1.56} & \textit{84.14}\stdev{1.39} & \textbf{51.49\stdev{1.97}} & 47.97\stdev{1.80} & \textbf{100.00\stdev{0.00}} & 74.57\stdev{3.60} & \textit{83.70}\stdev{0.77} & \textbf{85.95}\stdev{2.03} & \textit{73.33}\stdev{17.94} \\
Marginal $E^m$ & Min & 50.86\stdev{2.70} & 51.29\stdev{3.30} & 55.33\stdev{1.80} & 48.12\stdev{1.89} & 51.91\stdev{2.10} & 99.01\stdev{0.50} & 76.03\stdev{3.27} & 62.59\stdev{1.49} & 71.84\stdev{2.72} & 63.00\stdev{15.75} \\

Spilled $\Delta E_{s}$ & Max & \textbf{50.89\stdev{1.65}} & 50.77\stdev{5.72} & 56.08\stdev{2.48} & \textit{50.59}\stdev{1.72} & \textit{53.53}\stdev{2.81} & 95.61\stdev{0.56} & 43.94\stdev{3.21} & 50.87\stdev{1.87} & 51.21\stdev{1.68} & 55.94\stdev{14.35} \\

Spilled $\Delta E$ & Min & \textbf{50.89\stdev{1.65}} & \textbf{86.13}\stdev{4.28} & \textbf{89.01\stdev{1.06}} & 50.18\stdev{1.97} & 53.10\stdev{3.05} & \textit{99.66}\stdev{0.21} & \textbf{82.29}\stdev{2.46} & \textbf{89.10\stdev{1.75}} & 8\textit{2.70}\stdev{1.35} & \textbf{75.89}\stdev{17.98} \\

\midrule
& \multicolumn {11}{c}{Gemma-Instruct 1B \cite{gemma3}} \\\midrule 

Logit $E^\ell$ & Max & \textit{46.33}\stdev{0.82} & 48.12\stdev{11.45} & 58.89\stdev{1.61} & 50.50\stdev{2.45} & \textbf{53.49\stdev{3.71}} & 49.28\stdev{2.12} & \textbf{65.12\stdev{6.62}} & 62.24\stdev{3.62} & \textit{75.67}\stdev{1.96} & 56.63\stdev{9.13} \\
\midrule

Marginal $E^m$ & Max & 45.42\stdev{1.78} & \textbf{94.15\stdev{8.44}} & \textbf{83.66\stdev{1.82}} & 50.23\stdev{3.83} & 49.93\stdev{1.56} & \textbf{98.17\stdev{0.39}} & \textit{64.21}\stdev{6.67} & \textbf{86.87\stdev{1.39}} & \textbf{82.33\stdev{1.27}} & \textbf{72.77}\stdev{19.33} \\
Marginal $E^m$ & Min & \textbf{54.58\stdev{1.78}} & 28.93\stdev{14.50} & 39.80\stdev{2.54} & 49.84\stdev{4.38} & \textit{50.39}\stdev{1.80} & 56.33\stdev{1.60} & 63.20\stdev{4.27} & 41.58\stdev{2.85} & 61.56\stdev{1.61} & 49.58\stdev{10.47} \\

Spilled $\Delta E_{s}$ & Max & 45.17\stdev{2.37} & 33.27\stdev{11.49} & 49.01\stdev{1.67} & \textit{52.27}\stdev{3.56} & 49.91\stdev{2.59} & 77.48\stdev{1.92} & 40.49\stdev{4.17} & 49.18\stdev{3.93} & 35.77\stdev{2.13} & 48.06\stdev{12.13} \\

Spilled $\Delta E$ & Min & 45.02\stdev{2.45} & \textit{82.82}\stdev{12.91} & \textit{80.73}\stdev{2.16} & \textbf{52.48\stdev{3.75}} & 49.77\stdev{2.82} & \textit{92.93}\stdev{1.79} & 56.82\stdev{6.90} & \textit{85.64}\stdev{2.23} & 71.86\stdev{1.77} & \textit{68.67}\stdev{16.84} \\
\bottomrule \vspace{-40pt}
\end{tabular}
}
\end{table}
We observe a difference in the behavior in the base models and their instruction-tuned ones. While instruction-tuning generally improves generation quality, it can degrade the calibration of classical confidence metrics, as described in \cite{survey-hallucinations-llms, review-hallucinations}. For instance, examining the average performance in \cref{tab:detection_NO_FLIP_STD_Corretta}, the logit baseline $E_{\theta}^{\ell}$ decreases from 56.89\% to 54.62\% for LLaMA-3, indicating that fine-tuning may lead to overconfidence.
In contrast, Spilled Energy ($\Delta E_{\theta}$) consistently benefits from instruction tuning, showing improved detection rates across both LLaMA-3 (68.69\% to 73.16\%) and Mistral (73.94\% to 77.49\%).

\minisection{Variance and Generalization} 
A notable observation in \cref{tab:detection_NO_FLIP_STD_Corretta} is the higher standard deviation associated with marginal and spilled energy compared to the probing classifiers in the average column. This variance is not a weakness but a reflection of the method's training-free nature. Since $\Delta E_{\theta}$ relies on the intrinsic energy landscape of the LLM, its magnitude and sensitivity are naturally dependent on the specific domain (e.g., the sharp energy peaks in \textit{Math} and \textit{HotpotQA} versus the flatter distributions in \textit{Winobias} and \textit{IMDB}). Probing classifiers, by contrast, have high-variance when cross-testing yet the average of cross-testing results is mostly constant just above random chance ($\approx 62-64\%$).

\minisection{Limitations} \label{sec:limitations}
A current limitation of spilled energy is that it sometimes produces false positives on tokens that are not semantically informative, as shown in \cref{sec:supp-qual}. 
We observe this effect most prominently on punctuation tokens (e.g., commas, periods) and on words at the beginning of sentences. 
In these cases, the probability mass over the next token is naturally spread across many plausible options, leading to inflated spilled energy values even in otherwise correct generations. 
This highlights the importance of accurately identifying the \emph{exact answer tokens}, as detection is most reliable when restricted to the parts of the output that carry the semantic content of the answer.

\vspace{-5pt}
\section{Conclusion}
We reinterpreted the softmax layer of LLMs as an EBM, which lets us define
\emph{spilled energy}: the discrepancy between energy values that should be equal across
consecutive time steps. We show theoretically and empirically that this discrepancy provides a
strong, training-free signal for detecting hallucinations and errors in LLM outputs.
Through synthetic arithmetic experiments, we demonstrate that spilled energy reliably separates
correct from incorrect generations, outperforming baselines such as logits and marginal energy.
Across diverse real-world NLP benchmarks, spilled energy generalizes robustly without requiring
additional classifiers or task-specific training, unlike probing methods that struggle with transfer.
Overall, spilled energy offers a principled and practical framework for error detection in LLMs and
a new perspective on the internal energy dynamics of autoregressive models.

\section*{Ethics Statement}

This work adheres to the ICLR Code of Ethics. 
Our study focuses on methodological contributions to error and hallucination detection in Large Language Models. 
We do not train new models or collect additional data; instead, we rely exclusively on publicly available datasets and widely used benchmark models for evaluation. 

We note that part of our evaluation includes the Math dataset, which was publicly accessible at the time of experimentation but has since been taken down following a copyright claim. 
We emphasize that this dataset was used solely for evaluation purposes of our method, and only prior to the date of the takedown. 
No redistribution of the dataset was made, and our reported results are limited to demonstrating methodological effectiveness. 

Our work does not involve personally identifiable information, sensitive content, or human subjects, and does not raise foreseeable risks of harm. 
We believe the proposed approach contributes positively to research on trustworthy AI by providing a training-free and generalizable framework for error detection in language models.

\section*{Reproducibility Statement}

We are committed to ensuring the reproducibility of our results. 
All experimental details, including model configurations, evaluation protocols, and datasets used, are described in the main text and \cref{appendix:reproducibility}. 
Upon acceptance of this work, we will publicly release the code implementing our method, along with instructions to reproduce all reported experiments. 
This will allow the community to verify our findings and build upon our work.

\section*{Acknowledgment} This work was supported by projects PNRR MUR PE0000013-FAIR under the MUR National Recovery and Resilience Plan funded by the European Union - NextGenerationEU, PRIN 2022 project 20227YET9B ``AdVVent'' CUP code B53D23012830006. It was also partially supported by Sapienza research projects D2QNeT and BEAT (Better dEep leArning securiTy) --- bando per la ricerca di Ateneo 2024, and via the Seed of ERC grant ``MINT.AI" (cup B83C25001040001). This work is additionally supported by the MUR FIS2 grant n. FIS-2023-00942 ``NEXUS" (cup B53C25001030001). The work of Hazem Dewidar was carried out while he was enrolled in the Italian National Doctorate on Artificial Intelligence run by Sapienza University of Rome. Computing was supported by CINECA through the Italian SuperComputing Resource Allocation (ISCRA) projects Ge-Di HP10CRPUVC and SLEY HP10CX9CMC.

\bibliography{spilled,LLMknowmore_iclr2025,llms,ebms,datasets,hallucinations}

@String{ECCV = "ECCV"}

@String{ICCV = "ICCV"}

@String{ICML = "ICML"}

@String{NIPS = "NeurIPS"}

@String{ACL = "ACL"}

@String{ICLR = "ICLR"}

@article{huang2023survey,
  title={A survey on hallucination in large language models: Principles, taxonomy, challenges, and open questions},
  author={Huang, Lei and Yu, Weijiang and Ma, Weitao and Zhong, Weihong and Feng, Zhangyin and Wang, Haotian and Chen, Qianglong and Peng, Weihua and Feng, Xiaocheng and Qin, Bing and others},
  journal={arXiv preprint arXiv:2311.05232},
  year={2023}
}

@article{kadavath2022language,
  title={Language models (mostly) know what they know},
  author={Kadavath, Saurav and Conerly, Tom and Askell, Amanda and Henighan, Tom and Drain, Dawn and Perez, Ethan and Schiefer, Nicholas and Hatfield-Dodds, Zac and DasSarma, Nova and Tran-Johnson, Eli and others},
  journal={arXiv preprint arXiv:2207.05221},
  year={2022}
}

@article{li2024inference,
  title={Inference-time intervention: Eliciting truthful answers from a language model},
  author={Li, Kenneth and Patel, Oam and Vi{\'e}gas, Fernanda and Pfister, Hanspeter and Wattenberg, Martin},
  journal={Advances in Neural Information Processing Systems},
  volume={36},
  year={2024}
}

@misc{varshney2023stitch,
      title={A Stitch in Time Saves Nine: Detecting and Mitigating Hallucinations of LLMs by Validating Low-Confidence Generation}, 
      author={Neeraj Varshney and Wenlin Yao and Hongming Zhang and Jianshu Chen and Dong Yu},
      year={2023},
      eprint={2307.03987},
      archivePrefix={arXiv},
      primaryClass={cs.CL}
}

@inproceedings{rawte2023troubling,
    title = "The Troubling Emergence of Hallucination in Large Language Models - An Extensive Definition, Quantification, and Prescriptive Remediations",
    author = "Rawte, Vipula  and
      Chakraborty, Swagata  and
      Pathak, Agnibh  and
      Sarkar, Anubhav  and
      Tonmoy, S.M Towhidul Islam  and
      Chadha, Aman  and
      Sheth, Amit  and
      Das, Amitava",
    editor = "Bouamor, Houda  and
      Pino, Juan  and
      Bali, Kalika",
    booktitle = "Proceedings of the 2023 Conference on Empirical Methods in Natural Language Processing",
    month = dec,
    year = "2023",
    address = "Singapore",
    publisher = "Association for Computational Linguistics",
    url = "https://aclanthology.org/2023.emnlp-main.155/",
    doi = "10.18653/v1/2023.emnlp-main.155",
    pages = "2541--2573"
}

@article{ji2023survey,
  title={Survey of hallucination in natural language generation},
  author={Ji, Ziwei and Lee, Nayeon and Frieske, Rita and Yu, Tiezheng and Su, Dan and Xu, Yan and Ishii, Etsuko and Bang, Ye Jin and Madotto, Andrea and Fung, Pascale},
  journal={ACM Computing Surveys},
  volume={55},
  number={12},
  pages={1--38},
  year={2023},
  publisher={ACM New York, NY}
}

@article{bang2023multitask,
  title={A multitask, multilingual, multimodal evaluation of chatgpt on reasoning, hallucination, and interactivity},
  author={Bang, Yejin and Cahyawijaya, Samuel and Lee, Nayeon and Dai, Wenliang and Su, Dan and Wilie, Bryan and Lovenia, Holy and Ji, Ziwei and Yu, Tiezheng and Chung, Willy and others},
  journal={arXiv preprint arXiv:2302.04023},
  year={2023}
}

@article{millidge2023llms,
  title={{LLMs confabulate not hallucinate}},
  author={Millidge, Beren},
  journal={Beren's Blog},
  year={2023},
  month={March},
  url={https://www.beren.io/2023-03-19-LLMs-confabulate-not-hallucinate/}
}

@article{mcgowan2023chatgpt,
  title={ChatGPT and Bard exhibit spontaneous citation fabrication during psychiatry literature search},
  author={McGowan, Alessia and Gui, Yunlai and Dobbs, Matthew and Shuster, Sophia and Cotter, Matthew and Selloni, Alexandria and Goodman, Marianne and Srivastava, Agrima and Cecchi, Guillermo A and Corcoran, Cheryl M},
  journal={Psychiatry Research},
  volume={326},
  pages={115334},
  year={2023},
  publisher={Elsevier}
}

@inproceedings{liu-etal-2022-token,
    title = "A Token-level Reference-free Hallucination Detection Benchmark for Free-form Text Generation",
    author = "Liu, Tianyu  and
      Zhang, Yizhe  and
      Brockett, Chris  and
      Mao, Yi  and
      Sui, Zhifang  and
      Chen, Weizhu  and
      Dolan, Bill",
    editor = "Muresan, Smaranda  and
      Nakov, Preslav  and
      Villavicencio, Aline",
    booktitle = "Proceedings of the 60th Annual Meeting of the Association for Computational Linguistics (Volume 1: Long Papers)",
    month = may,
    year = "2022",
    address = "Dublin, Ireland",
    publisher = "Association for Computational Linguistics",
    url = "https://aclanthology.org/2022.acl-long.464",
    doi = "10.18653/v1/2022.acl-long.464",
    pages = "6723--6737",
}

@misc{jiang2023mistral7b,
      title={Mistral 7B}, 
      author={Albert Q. Jiang and Alexandre Sablayrolles and Arthur Mensch and Chris Bamford and Devendra Singh Chaplot and Diego de las Casas and Florian Bressand and Gianna Lengyel and Guillaume Lample and Lucile Saulnier and Lélio Renard Lavaud and Marie-Anne Lachaux and Pierre Stock and Teven Le Scao and Thibaut Lavril and Thomas Wang and Timothée Lacroix and William El Sayed},
      year={2023},
      eprint={2310.06825},
      archivePrefix={arXiv},
      primaryClass={cs.CL},
      url={https://arxiv.org/abs/2310.06825}, 
}

@article{salles2020anthropomorphism,
  title={Anthropomorphism in AI},
  author={Salles, Arleen and Evers, Kathinka and Farisco, Michele},
  journal={AJOB neuroscience},
  volume={11},
  number={2},
  pages={88--95},
  year={2020},
  publisher={Taylor \& Francis}
}

@article{harnad2024language,
  title={Language writ large: Llms, chatgpt, grounding, meaning and understanding},
  author={Harnad, Stevan},
  journal={arXiv preprint arXiv:2402.02243},
  year={2024}
}

@article{serapio2023personality,
  title={Personality traits in large language models},
  author={Serapio-Garc{\'\i}a, Greg and Safdari, Mustafa and Crepy, Cl{\'e}ment and Sun, Luning and Fitz, Stephen and Romero, Peter and Abdulhai, Marwa and Faust, Aleksandra and Matari{\'c}, Maja},
  journal={arXiv preprint arXiv:2307.00184},
  year={2023}
}

@article{belinkov2021probingclassifierspromisesshortcomings,
    title = "Probing Classifiers: Promises, Shortcomings, and Advances",
    author = "Belinkov, Yonatan",
    journal = "Computational Linguistics",
    volume = "48",
    number = "1",
    month = mar,
    year = "2022",
    address = "Cambridge, MA",
    publisher = "MIT Press",
    url = "https://aclanthology.org/2022.cl-1.7/",
    doi = "10.1162/coli_a_00422",
    pages = "207--219",
}

@article{hendrycksmath2021,
  title={Measuring Mathematical Problem Solving With the MATH Dataset},
  author={Dan Hendrycks and Collin Burns and Saurav Kadavath and Akul Arora and Steven Basart and Eric Tang and Dawn Song and Jacob Steinhardt},
  journal={NeurIPS},
  year={2021}
}

@inproceedings{winobias,
    title = "Gender Bias in Coreference Resolution: Evaluation and Debiasing Methods",
    author = "Zhao, Jieyu  and
      Wang, Tianlu  and
      Yatskar, Mark  and
      Ordonez, Vicente  and
      Chang, Kai-Wei",
    editor = "Walker, Marilyn  and
      Ji, Heng  and
      Stent, Amanda",
    booktitle = "Proceedings of the 2018 Conference of the North {A}merican Chapter of the Association for Computational Linguistics: Human Language Technologies, Volume 2 (Short Papers)",
    month = jun,
    year = "2018",
    address = "New Orleans, Louisiana",
    publisher = "Association for Computational Linguistics",
    url = "https://aclanthology.org/N18-2003/",
    doi = "10.18653/v1/N18-2003",
    pages = "15--20",
}

@InProceedings{imdb_dataset,
  author    = {Maas, Andrew L.  and  Daly, Raymond E.  and  Pham, Peter T.  and  Huang, Dan  and  Ng, Andrew Y.  and  Potts, Christopher},
  title     = {Learning Word Vectors for Sentiment Analysis},
  booktitle = {Proceedings of the 49th Annual Meeting of the Association for Computational Linguistics: Human Language Technologies},
  month     = {June},
  year      = {2011},
  address   = {Portland, Oregon, USA},
  publisher = {Association for Computational Linguistics},
  pages     = {142--150},
  url       = {http://www.aclweb.org/anthology/P11-1015}
}

@inproceedings{joshi2017triviaqa,
  title={TriviaQA: A Large Scale Distantly Supervised Challenge Dataset for Reading Comprehension},
  author={Joshi, Mandar and Choi, Eunsol and Weld, Daniel S and Zettlemoyer, Luke},
  booktitle={Proceedings of the 55th Annual Meeting of the Association for Computational Linguistics (Volume 1: Long Papers)},
  pages={1601--1611},
  year={2017}
}

@inproceedings{yang2018hotpotqa,
  title={HotpotQA: A Dataset for Diverse, Explainable Multi-hop Question Answering},
  author={Yang, Zhilin and Qi, Peng and Zhang, Saizheng and Bengio, Yoshua and Cohen, William and Salakhutdinov, Ruslan and Manning, Christopher D},
  booktitle={Proceedings of the 2018 Conference on Empirical Methods in Natural Language Processing},
  pages={2369--2380},
  year={2018}
}

@article{sakaguchi2021winogrande,
  title={Winogrande: An adversarial winograd schema challenge at scale},
  author={Sakaguchi, Keisuke and Bras, Ronan Le and Bhagavatula, Chandra and Choi, Yejin},
  journal={Communications of the ACM},
  volume={64},
  number={9},
  pages={99--106},
  year={2021},
  publisher={ACM New York, NY, USA}
}

@InProceedings{mnli,
  author = "Williams, Adina
            and Nangia, Nikita
            and Bowman, Samuel",
  title = "A Broad-Coverage Challenge Corpus for
           Sentence Understanding through Inference",
  booktitle = "Proceedings of the 2018 Conference of
               the North American Chapter of the
               Association for Computational Linguistics:
               Human Language Technologies, Volume 1 (Long
               Papers)",
  year = "2018",
  publisher = "Association for Computational Linguistics",
  pages = "1112--1122",
  location = "New Orleans, Louisiana",
  url = "http://aclweb.org/anthology/N18-1101"
}

@inbook{LeCun06EBM,
title = "A tutorial on energy-based learning",
author = "Yann Lecun and Sumit Chopra and Raia Hadsell and Ranzato, {Marc Aurelio} and Huang, {Fu Jie}",
year = {2006},
language = "English (US)",
booktitle = "Predicting structured data",
publisher = "MIT Press",
}

@inproceedings{energyOOD,
author = {Liu, Weitang and Wang, Xiaoyun and Owens, John D. and Li, Yixuan},
title = {Energy-based out-of-distribution detection},
year = {2020},
booktitle = NIPS,
}

@inproceedings{mirza2024shedding, 
  author = {Mirza, Mujtaba Hussain and Maria Rosaria Briglia and Senad Beadini and Iacopo Masi}, 
  title = {Shedding More Light on Robust Classifiers under the lens of Energy-based Models}, 
  booktitle = ECCV,
  year = {2024}
}

@inproceedings{grathwohl2020your,
  title={Your classifier is secretly an energy based model and you should treat it like one},
  author={Grathwohl, Will and Wang, Kuan-Chieh and Jacobsen, Joern-Henrik and Duvenaud, David and Norouzi, Mohammad and Swersky, Kevin},
  booktitle=ICLR,
  year = {2020}
}

@misc{mirza2025understandingadversarialtrainingenergybased,
  title={Understanding Adversarial Training with Energy-based Models}, 
  author={Mujtaba Hussain Mirza and Maria Rosaria Briglia and Filippo Bartolucci and Senad Beadini and Giuseppe Lisanti and Iacopo Masi},
  year={2025},
  eprint={2505.22486},
  archivePrefix={arXiv},
  primaryClass={cs.LG},
  url={https://arxiv.org/abs/2505.22486}, 
}

@misc{xu2025hallucinationinevitableinnatelimitation,
    title={Hallucination is Inevitable: An Innate Limitation of Large Language Models}, 
    author={Ziwei Xu and Sanjay Jain and Mohan Kankanhalli},
    year={2025},
    eprint={2401.11817},
    archivePrefix={arXiv},
    primaryClass={cs.CL},
    url={https://arxiv.org/abs/2401.11817}, 
}

@techreport{kalai2025hallucinate,
    title        = {Why Language Models Hallucinate},
    author       = {Kalai, Adam Tauman and Nachum, Ofir and Vempala, Santosh S. and Zhang, Edwin},
    institution  = {OpenAI and Georgia Tech},
    year         = {2025},
    month        = sep,
    note         = {Technical Report},
}

@inproceedings{santilli2025revisiting,
    title={Revisiting uncertainty quantification evaluation in language models: Spurious interactions with response length bias results},
    author={Santilli, Andrea and Golinski, Adam and Kirchhof, Michael and Danieli, Federico and Blaas, Arno and Xiong, Miao and Zappella, Luca and Williamson, Sinead},
    booktitle=ACL,
    year={2025}
}

@misc{peng2023checkfactstryagain,
    title={Check Your Facts and Try Again: Improving Large Language Models with External Knowledge and Automated Feedback}, 
    author={Baolin Peng and Michel Galley and Pengcheng He and Hao Cheng and Yujia Xie and Yu Hu and Qiuyuan Huang and Lars Liden and Zhou Yu and Weizhu Chen and Jianfeng Gao},
    year={2023},
    eprint={2302.12813},
    archivePrefix={arXiv},
    primaryClass={cs.CL},
    url={https://arxiv.org/abs/2302.12813}, 
}

@misc{fu2025deepthinkconfidence,
    title={Deep Think with Confidence}, 
    author={Yichao Fu and Xuewei Wang and Yuandong Tian and Jiawei Zhao},
    year={2025},
    eprint={2508.15260},
    archivePrefix={arXiv},
    primaryClass={cs.LG},
    url={https://arxiv.org/abs/2508.15260}, 
}

@article{farquhar2024semanticentropy,
    author    = {Farquhar, Sebastian and Kossen, Jannik and Kuhn, Lorenz and Gal, Yarin},
    title     = {Detecting hallucinations in large language models using semantic entropy},
    journal   = {Nature},
    volume    = {630},
    number    = {8017},
    pages     = {625--630},
    year      = {2024},
    month     = jun,
    doi       = {10.1038/s41586-024-07421-0},
    url       = {https://doi.org/10.1038/s41586-024-07421-0},
    pmid      = {38898292},
    pmcid     = {PMC11186750}
}

@inproceedings{kuhn2023semanticuncertaintylinguisticinvariances,
    title={Semantic Uncertainty: Linguistic Invariances for Uncertainty Estimation in Natural Language Generation},
    author={Lorenz Kuhn and Yarin Gal and Sebastian Farquhar},
    booktitle={The Eleventh International Conference on Learning Representations },
    year={2023},
    url={https://openreview.net/forum?id=VD-AYtP0dve}
}

@inproceedings{fadeeva2024factcheckingoutputlargelanguage,
    title = "Fact-Checking the Output of Large Language Models via Token-Level Uncertainty Quantification",
    author = "Fadeeva, Ekaterina  and
      Rubashevskii, Aleksandr  and
      Shelmanov, Artem  and
      Petrakov, Sergey  and
      Li, Haonan  and
      Mubarak, Hamdy  and
      Tsymbalov, Evgenii  and
      Kuzmin, Gleb  and
      Panchenko, Alexander  and
      Baldwin, Timothy  and
      Nakov, Preslav  and
      Panov, Maxim",
    editor = "Ku, Lun-Wei  and
      Martins, Andre  and
      Srikumar, Vivek",
    booktitle = "Findings of the Association for Computational Linguistics: ACL 2024",
    month = aug,
    year = "2024",
    address = "Bangkok, Thailand",
    publisher = "Association for Computational Linguistics",
    url = "https://aclanthology.org/2024.findings-acl.558/",
    doi = "10.18653/v1/2024.findings-acl.558",
    pages = "9367--9385",
}

@misc{karpowicz2025fundamentalimpossibilityhallucinationcontrol,
    title={On the Fundamental Impossibility of Hallucination Control in Large Language Models}, 
    author={Michał P. Karpowicz},
    year={2025},
    eprint={2506.06382},
    archivePrefix={arXiv},
    primaryClass={stat.ML},
    url={https://arxiv.org/abs/2506.06382}, 
}

@misc{kuhn2023clamselectiveclarificationambiguous,
    title={CLAM: Selective Clarification for Ambiguous Questions with Generative Language Models}, 
    author={Lorenz Kuhn and Yarin Gal and Sebastian Farquhar},
    year={2023},
    eprint={2212.07769},
    archivePrefix={arXiv},
    primaryClass={cs.CL},
    url={https://arxiv.org/abs/2212.07769}, 
}

@misc{kossen2025semantic,
    title={Semantic Entropy Probes: Robust and Cheap Hallucination Detection in {LLM}s},
    author={Jannik Kossen and Jiatong Han and Muhammed Razzak and Lisa Schut and Shreshth A Malik and Yarin Gal},
    year={2025},
    url={https://openreview.net/forum?id=YQvvJjLWX0}
}

@inproceedings{orgad2024llms,
    title={LLMs Know More Than They Show: On the Intrinsic Representation of LLM Hallucinations},
    author={Orgad, Hadas and Toker, Michael and Gekhman, Zorik and Reichart, Roi and Szpektor, Idan and Kotek, Hadas and Belinkov, Yonatan},
    booktitle=ICLR,
    year={2025}
}

@inproceedings{yin2023large,
    title={Do Large Language Models Know What They Don’t Know?},
    author={Yin, Zhangyue and Sun, Qiushi and Guo, Qipeng and Wu, Jiawen and Qiu, Xipeng and Huang, Xuan-Jing},
    booktitle={ACL},
    year={2023}
}

@misc{gemma3,
      title={Gemma 3 Technical Report}, 
      author={Aishwarya Kamath and Johan Ferret and Shreya Pathak and Nino Vieillard and Ramona Merhej and Sarah Perrin and Tatiana Matejovicova and Alexandre Ramé and Morgane Rivière and Louis Rouillard and Thomas Mesnard and Geoffrey Cideron and Jean-bastien Grill and Sabela Ramos and Edouard Yvinec and Michelle Casbon and Etienne Pot and Ivo Penchev and Gaël Liu and Francesco Visin and Kathleen Kenealy and Lucas Beyer and Xiaohai Zhai and Anton Tsitsulin and Robert Busa-Fekete and Alex Feng and Noveen Sachdeva and Benjamin Coleman and Yi Gao and Basil Mustafa and Iain Barr and Emilio Parisotto and David Tian and Matan Eyal and Colin Cherry and Jan-Thorsten Peter and Danila Sinopalnikov and Surya Bhupatiraju and Rishabh Agarwal and Mehran Kazemi and Dan Malkin and Ravin Kumar and David Vilar and Idan Brusilovsky and Jiaming Luo and Andreas Steiner and Abe Friesen and Abhanshu Sharma and Abheesht Sharma and Adi Mayrav Gilady and Adrian Goedeckemeyer and Alaa Saade and Alex Feng and Alexander Kolesnikov and Alexei Bendebury and Alvin Abdagic and Amit Vadi and András György and André Susano Pinto and Anil Das and Ankur Bapna and Antoine Miech and Antoine Yang and Antonia Paterson and Ashish Shenoy and Ayan Chakrabarti and Bilal Piot and Bo Wu and Bobak Shahriari and Bryce Petrini and Charlie Chen and Charline Le Lan and Christopher A. Choquette-Choo and CJ Carey and Cormac Brick and Daniel Deutsch and Danielle Eisenbud and Dee Cattle and Derek Cheng and Dimitris Paparas and Divyashree Shivakumar Sreepathihalli and Doug Reid and Dustin Tran and Dustin Zelle and Eric Noland and Erwin Huizenga and Eugene Kharitonov and Frederick Liu and Gagik Amirkhanyan and Glenn Cameron and Hadi Hashemi and Hanna Klimczak-Plucińska and Harman Singh and Harsh Mehta and Harshal Tushar Lehri and Hussein Hazimeh and Ian Ballantyne and Idan Szpektor and Ivan Nardini and Jean Pouget-Abadie and Jetha Chan and Joe Stanton and John Wieting and Jonathan Lai and Jordi Orbay and Joseph Fernandez and Josh Newlan and Ju-yeong Ji and Jyotinder Singh and Kat Black and Kathy Yu and Kevin Hui and Kiran Vodrahalli and Klaus Greff and Linhai Qiu and Marcella Valentine and Marina Coelho and Marvin Ritter and Matt Hoffman and Matthew Watson and Mayank Chaturvedi and Michael Moynihan and Min Ma and Nabila Babar and Natasha Noy and Nathan Byrd and Nick Roy and Nikola Momchev and Nilay Chauhan and Noveen Sachdeva and Oskar Bunyan and Pankil Botarda and Paul Caron and Paul Kishan Rubenstein and Phil Culliton and Philipp Schmid and Pier Giuseppe Sessa and Pingmei Xu and Piotr Stanczyk and Pouya Tafti and Rakesh Shivanna and Renjie Wu and Renke Pan and Reza Rokni and Rob Willoughby and Rohith Vallu and Ryan Mullins and Sammy Jerome and Sara Smoot and Sertan Girgin and Shariq Iqbal and Shashir Reddy and Shruti Sheth and Siim Põder and Sijal Bhatnagar and Sindhu Raghuram Panyam and Sivan Eiger and Susan Zhang and Tianqi Liu and Trevor Yacovone and Tyler Liechty and Uday Kalra and Utku Evci and Vedant Misra and Vincent Roseberry and Vlad Feinberg and Vlad Kolesnikov and Woohyun Han and Woosuk Kwon and Xi Chen and Yinlam Chow and Yuvein Zhu and Zichuan Wei and Zoltan Egyed and Victor Cotruta and Minh Giang and Phoebe Kirk and Anand Rao and Kat Black and Nabila Babar and Jessica Lo and Erica Moreira and Luiz Gustavo Martins and Omar Sanseviero and Lucas Gonzalez and Zach Gleicher and Tris Warkentin and Vahab Mirrokni and Evan Senter and Eli Collins and Joelle Barral and Zoubin Ghahramani and Raia Hadsell and Yossi Matias and D. Sculley and Slav Petrov and Noah Fiedel and Noam Shazeer and Oriol Vinyals and Jeff Dean and Demis Hassabis and Koray Kavukcuoglu and Clement Farabet and Elena Buchatskaya and Jean-Baptiste Alayrac and Rohan Anil and Dmitry and Lepikhin and Sebastian Borgeaud and Olivier Bachem and Armand Joulin and Alek Andreev and Cassidy Hardin and Robert Dadashi and Léonard Hussenot},
      year={2025},
      eprint={2503.19786},
      archivePrefix={arXiv},
      primaryClass={cs.CL},
      url={https://arxiv.org/abs/2503.19786}, 
}

@misc{openai2023gpt4,
 archiveprefix = {arXiv},
 author = {OpenAI-Team},
 eprint = {2303.08774},
 primaryclass = {cs.CL},
 title = {GPT-4 Technical Report},
 year = {2023},
 url = {https://cdn.openai.com/papers/gpt-4.pdf}
}

@article{liu2024deepseek,
  title={Deepseek-v3 technical report},
  author={Liu, Aixin and Feng, Bei and Xue, Bing and Wang, Bingxuan and Wu, Bochao and Lu, Chengda and Zhao, Chenggang and Deng, Chengqi and Zhang, Chenyu and Ruan, Chong and others},
  journal={arXiv preprint arXiv:2412.19437},
  year={2024}
}

@article{dubey2024llama,
  title={The {L}{L}a{M}a 3 herd of models},
  author={Dubey, Abhimanyu and Jauhri, Abhinav and Pandey, Abhinav and Kadian, Abhishek and Al-Dahle, Ahmad and Letman, Aiesha and Mathur, Akhil and Schelten, Alan and Yang, Amy and Fan, Angela and others},
  journal={arXiv e-prints},
  pages={arXiv--2407},
  year={2024}
}

@misc{qwen2024qwen3,
  author       = {Qwen-Team},
  title        = {Qwen3: Think Deeper, Act Faster},
  year         = {2025},
  url = {https://qwen.ai/blog?id=1e3fa5c2d4662af2855586055ad037ed9e555125},
  note         = {Accessed: 2025-09-23}
}

@misc{ouyang2022traininglanguagemodelsfollow,
      title={Training language models to follow instructions with human feedback}, 
      author={Long Ouyang and Jeff Wu and Xu Jiang and Diogo Almeida and Carroll L. Wainwright and Pamela Mishkin and Chong Zhang and Sandhini Agarwal and Katarina Slama and Alex Ray and John Schulman and Jacob Hilton and Fraser Kelton and Luke Miller and Maddie Simens and Amanda Askell and Peter Welinder and Paul Christiano and Jan Leike and Ryan Lowe},
      year={2022},
      eprint={2203.02155},
      archivePrefix={arXiv},
      primaryClass={cs.CL},
      url={https://arxiv.org/abs/2203.02155}, 
}

@inproceedings{li-etal-2023-contrastive,
    title = "Contrastive Decoding: Open-ended Text Generation as Optimization",
    author = "Li, Xiang Lisa  and
      Holtzman, Ari  and
      Fried, Daniel  and
      Liang, Percy  and
      Eisner, Jason  and
      Hashimoto, Tatsunori  and
      Zettlemoyer, Luke  and
      Lewis, Mike",
    editor = "Rogers, Anna  and
      Boyd-Graber, Jordan  and
      Okazaki, Naoaki",
    booktitle = "Proceedings of the 61st Annual Meeting of the Association for Computational Linguistics (Volume 1: Long Papers)",
    month = jul,
    year = "2023",
    address = "Toronto, Canada",
    publisher = "Association for Computational Linguistics",
    url = "https://aclanthology.org/2023.acl-long.687/",
    doi = "10.18653/v1/2023.acl-long.687",
    pages = "12286--12312",
}

@inproceedings{improvinglu,
  title={Improving Language Understanding by Generative Pre-Training},
  author={Alec Radford and Karthik Narasimhan},
  year={2018},
  url={https://cdn.openai.com/research-covers/language-unsupervised/language_understanding_paper.pdf},
  booktitle={Open{AI} Technical Report}
}

@inproceedings{bender2021dangers,
  title={On the dangers of stochastic parrots: Can language models be too big?},
  author={Bender, Emily M and Gebru, Timnit and McMillan-Major, Angelina and Shmitchell, Shmargaret},
  booktitle={Proceedings of the 2021 ACM conference on fairness, accountability, and transparency},
  pages={610--623},
  year={2021}
}

@inproceedings{gekhman2025insideouthiddenfactualknowledge,
    title={Inside-Out: Hidden Factual Knowledge in {LLM}s},
    author={Zorik Gekhman and Eyal Ben-David and Hadas Orgad and Eran Ofek and Yonatan Belinkov and Idan Szpektor and Jonathan Herzig and Roi Reichart},
    booktitle={Second Conference on Language Modeling},
    year={2025},
    url={https://openreview.net/forum?id=f7GG1MbsSM}
}

@inproceedings{zhu2021towardsunderstandingthegenerativecapabilities,
    author={Zhu, Yao and Ma, Jiacheng and Sun, Jiacheng and Chen, Zewei and Jiang, Rongxin and Chen, Yaowu and Li, Zhenguo},
    booktitle=ICCV,
    title={Towards Understanding the Generative Capability of Adversarially Robust Classifiers}, 
    year={2021},
    pages={7708-7717},
}

@inproceedings{kingma2013auto,
  title={Auto-Encoding Variational Bayes},
  author={Kingma, Diederik P and Welling, Max},
  booktitle=ICLR,
  year={2014}
}

@inproceedings{goodfellow2014generative,
  title={Generative Adversarial Nets},
  author={Goodfellow, Ian and Pouget-Abadie, Jean and Mirza, Mehdi and Xu, Bing and Warde-Farley, David and Ozair, Sherjil and Courville, Aaron and Bengio, Yoshua},
  booktitle=NIPS,
  year={2014}
}

@inproceedings{sohl2015deep,
  title={Deep unsupervised learning using nonequilibrium thermodynamics},
  author={Sohl-Dickstein, Jascha and Weiss, Eric and Maheswaranathan, Niru and Ganguli, Surya},
  booktitle=ICML,
  year={2015},
}

@inproceedings{ho2020denoising,
  title={Denoising diffusion probabilistic models},
  author={Ho, Jonathan and Jain, Ajay and Abbeel, Pieter},
  booktitle=NIPS,
  year={2020}
}

@inproceedings{subramani2022extractinglatentsteeringvectors,
    title = "Extracting Latent Steering Vectors from Pretrained Language Models",
    author = "Subramani, Nishant  and
      Suresh, Nivedita  and
      Peters, Matthew",
    editor = "Muresan, Smaranda  and
      Nakov, Preslav  and
      Villavicencio, Aline",
    booktitle = "Findings of the Association for Computational Linguistics: ACL 2022",
    month = may,
    year = "2022",
    address = "Dublin, Ireland",
    publisher = "Association for Computational Linguistics",
    url = "https://aclanthology.org/2022.findings-acl.48/",
    doi = "10.18653/v1/2022.findings-acl.48",
    pages = "566--581",
}

@inproceedings{dunefsky2025oneshotoptimizedsteeringvectors,
    title={One-shot Optimized Steering Vectors Mediate Safety-relevant Behaviors in {LLM}s},
    author={Jacob Dunefsky and Arman Cohan},
    booktitle={Second Conference on Language Modeling},
    year={2025},
    url={https://openreview.net/forum?id=teW4nIZ1gy}
}

@article{survey-hallucinations-llms,
  author       = {Lei Huang and
                  Weijiang Yu and
                  Weitao Ma and
                  Weihong Zhong and
                  Zhangyin Feng and
                  Haotian Wang and
                  Qianglong Chen and
                  Weihua Peng and
                  Xiaocheng Feng and
                  Bing Qin and
                  Ting Liu},
  title        = {A Survey on Hallucination in Large Language Models: Principles, Taxonomy,
                  Challenges, and Open Questions},
  journal      = {CoRR},
  volume       = {abs/2311.05232},
  year         = {2023},
}

@misc{review-hallucinations,
      title={Review of Hallucination Understanding in Large Language and Vision Models}, 
      author={Zhengyi Ho and Siyuan Liang and Dacheng Tao},
      year={2025},
      eprint={2510.00034},
      archivePrefix={arXiv},
      primaryClass={cs.CV},
      url={https://arxiv.org/abs/2510.00034}, 
}

@misc{languagemodelsmostlyknow,
      title={Language Models (Mostly) Know What They Know}, 
      author={Saurav Kadavath and Tom Conerly and Amanda Askell and Tom Henighan and Dawn Drain and Ethan Perez and Nicholas Schiefer and Zac Hatfield-Dodds and Nova DasSarma and Eli Tran-Johnson and Scott Johnston and Sheer El-Showk and Andy Jones and Nelson Elhage and Tristan Hume and Anna Chen and Yuntao Bai and Sam Bowman and Stanislav Fort and Deep Ganguli and Danny Hernandez and Josh Jacobson and Jackson Kernion and Shauna Kravec and Liane Lovitt and Kamal Ndousse and Catherine Olsson and Sam Ringer and Dario Amodei and Tom Brown and Jack Clark and Nicholas Joseph and Ben Mann and Sam McCandlish and Chris Olah and Jared Kaplan},
      year={2022},
      eprint={2207.05221},
      archivePrefix={arXiv},
      primaryClass={cs.CL},
}
\bibliographystyle{iclr2026_conference}

\appendix
\section{Appendix}
\subsection{Partition Functions Proof used in \cref{eq:cond-prob-to-energy}} \label{appendix:partitions}
We extend the proof of \citeauthor{zhu2021towardsunderstandingthegenerativecapabilities} to the sequence-to-sequence setting by treating next-token prediction as a multi-class classification problem. 
At step $i$, the input is the prefix $\compactseq{i-1}$, and the model outputs logits over the vocabulary $\vocab$ of size $V$. 
For notational consistency, we define the following energy terms:
\begin{equation}
\left\{
\begin{array}{l}
\Eseqq{i}{\ell} = -\log\!\left(\exp\big(\net(\compactseq[]{i-1})[\texttt{id}(\bx_{i})]\big)\right), \\[0.5em]
\Eseqq{i-1}{m} = -\log\!\left(\sum_{k=1}^{V} \exp\big(\net(\compactseq[]{i-1})[k]\big)\right).
\end{array}
\right.
\label{eq:energies-proof-part-func}
\end{equation}

The probability of the sequence up to position $i$ can be expressed as
\begin{equation}
    p_{\net}(\compactseq[]{i}) = \frac{\exp(-\Eseqq{i}{\ell})}{Z_{\net}},
\end{equation}
where $Z_{\net}$ is the global partition function (normalizing constant), defined over all possible continuations of all prefixes:
\begin{equation}
    Z_{\net} 
    = \sum_{\compactseq[]{i-1}} \sum_{\bx_i} \exp\!\left(\net(\compactseq[]{i-1})[\texttt{id}(\bx_{i})]\right) 
    = \sum_{\compactseq[]{i-1}} \sum_{k=1}^{V} \exp\!\left(\net(\compactseq[]{i-1})[k]\right).
    \label{eq:Z-part-func}
\end{equation}

Similarly, the probability of the prefix $\compactseq[]{i-1}$ can be written using the marginal energy:
\begin{equation}
    p_{\net}(\compactseq[]{i-1}) = \frac{\exp(-\Eseqq{i-1}{m})}{\widetilde{Z}_{\net}},
\end{equation}
where $\widetilde{Z}_{\net}$ is the corresponding normalizing constant:
\begin{equation}
    \widetilde{Z}_{\net} 
    = \sum_{\compactseq[]{i-1}} \exp\!\left(-\Eseqq{i-1}{m}\right) 
    = \sum_{\compactseq[]{i-1}} \exp\!\left(\log \sum_{k=1}^{V} \exp\!\big(\net(\compactseq[]{i-1})[k]\big)\right).
    \label{eq:Z-tilde-part-func}
\end{equation}

By expanding the logarithm in \cref{eq:Z-tilde-part-func}, we obtain
\begin{equation}
    \widetilde{Z}_{\net} = \sum_{\compactseq[]{i-1}} \sum_{k=1}^{V} \exp\!\left(\net(\compactseq[]{i-1})[k]\right),
\end{equation}
which is identical to \cref{eq:Z-part-func}. 
Hence, the two partition functions coincide:
\begin{equation}
    Z_{\net} = \widetilde{Z}_{\net}.
\end{equation}

\subsection{The Role of Temperature in Spilled Energy}

We now analyze how the temperature parameter $\tau$ affects the definition of spilled energy. 
Starting from \cref{eq:prob-next-token-practice}, the probability of the next token under temperature scaling is
\begin{align}
\log p_{\theta}(\bx_{i}\ |\ \compactseq[]{i-1}) 
&= \log \frac{\exp\!\left(\tfrac{1}{\tau} \net(\compactseq[]{i-1})[\text{Id}(\bx_{i})]\right)}{\sum\limits_{k} \exp\!\left(\tfrac{1}{\tau} \net(\compactseq[]{i-1})[k]\right)} \\
&= \frac{1}{\tau}\, \net(\compactseq[]{i-1})[\text{Id}(\bx_{i})] 
 - \log \sum\limits_{k} \exp\!\left(\tfrac{1}{\tau} \net(\compactseq[]{i-1})[k]\right).
\end{align}

Accordingly, the spilled energy becomes
\begin{align}
\Delta E_{\theta}(\compactseq[]{i}) 
= \frac{1}{\tau}\, \net(\compactseq[]{i-1})[\text{Id}(\bx_{i})] 
 - \log \sum_{k=1}^{|V|} \exp\!\left(\tfrac{1}{\tau} \net(\bx_{i}, \dots, \bx_{1})[k]\right).
\end{align}

\paragraph{Limit case $\tau \to \infty$.}  
When the temperature tends to infinity, the logits are scaled down towards zero, making all tokens equally likely:
\begin{align}
\lim_{\tau \to +\infty} \Delta E_{\theta}(\compactseq[]{i}) 
&= \lim_{\tau \to \infty} \frac{1}{\tau} \net(\compactseq[]{i-1})[\text{Id}(\bx_{i})] 
 - \log \sum_{k=1}^{|V|} \exp\!\left(\tfrac{1}{\tau} \net(\compactseq[]{i-1})[k]\right) \\
&= 0 - \log \sum_{k=1}^{|V|} \exp(0) \\
&= -\log |V|.
\end{align}
Thus, for $\tau \to \infty$ the model degenerates into a uniform random classifier over the vocabulary.

\paragraph{Interpretation.}  
Varying $\tau$ perturbs the balance between the two energy terms, introducing a systematic error in $\Delta E_{\theta}$. 
From the perspective of the Boltzmann distribution, scaling by $\tfrac{1}{\tau}$ corresponds to injecting or removing energy from the system. 
At high temperatures ($\tau \to \infty$), the system approaches maximum entropy, where all tokens have equal probability. 
At low temperatures ($\tau \to 0^+$), the distribution collapses onto the maximum logit token, making the model highly deterministic.

\paragraph{Error accumulation.}  
As we generate tokens sequentially, we accumulate deviations in $\Delta E_{\theta}$:
\begin{align}
\log p_{\theta}(\compactseq[]{i-1}) 
= \frac{1}{\tau} \net(\compactseq[]{i-1})[\text{Id}(\bx_{i})] 
- \log \sum_{k} \exp\!\left(\tfrac{1}{\tau} \net(\compactseq[]{i-1})[k]\right) 
+ \sum_{j=1}^{i}\Delta E_{\theta}(\compactseq[]{j}).
\end{align}
Hence, temperature scaling not only modifies the probabilities but also reshapes the cumulative error landscape traced by spilled energy.

\subsection{Why Spilled Energy should be zero?}\label{sec:small-prof}

\tbf{TL;DR} Consider Eq. (2) in our paper and the simplification that occurs between the two probabilities between step $i$ and step $i-1$: that simplification occurs because the probability in the denominator at step $i$ is the same as the probability in the numerator at step $i-1$ in order to perform language modeling correctly. We measure those inside and LLMs in terms of energy, and the spilled energy is the amount by which they differ.

Please see the definition below. Let us assume a sequence of three tokens  $\mathbf{x}_2, \mathbf{x}_1, \mathbf{x}_0$. If we do language modeling with autoregression, minimizing the negative log-likelihood, we have:

$$
-\log p(\mathbf{x}_2, \mathbf{x}_1, \mathbf{x}_0) = -\log \underbrace{p(\mathbf{x}_2|\mathbf{x}_1,\mathbf{x}_0)}_{\text{step 2}}p(\mathbf{x}_1|\mathbf{x}_0)p(\mathbf{x}_0)
$$

Now, every conditional probability on the right side is implemented with a transformer ending in a softmax discriminative classifier. \cref{eq:prob-next-token-practice} and \cref{eq:p-next-token} allow us to re-interpret:

\begin{align}
\text{\textbf{step 2:}}\quad -\log p(\mathbf{x}_2|\mathbf{x}_1,\mathbf{x}_0) = -\log \frac{p(\mathbf{x}_2,\mathbf{x}_1,\mathbf{x}_0)}{p(\mathbf{x}_1,\mathbf{x}_0)} = -\log \Bigg[ \frac{ \exp\big(\theta(\mathbf{x}_1,\mathbf{x}_0)[id(\mathbf{x}_2)]\big)}{\sum_k^V\exp\big(\theta(\mathbf{x}_1,\mathbf{x}_0)[k]\big)} \Bigg] =  \\ =  E^{\ell}(\mathbf{x}_2,\mathbf{x}_1,\mathbf{x}_0)-E^m(\mathbf{x}_1,\mathbf{x}_0).
\end{align}
In other words, we reinterpret:

\begin{itemize}
    \item the numerator $p(\mathbf{x}_2,\mathbf{x}_1,\mathbf{x}_0)$ as the energy $E^{\ell}(\mathbf{x}_2,\mathbf{x}_1,\mathbf{x}_0)$, which is the \tbf{logit ($\ell$)} of the softmax at timestep 2;
    \item The denominator as the energy $E^m(\mathbf{x}_1,\mathbf{x}_0)$ obtained with the \tbf{marginalization (m)} across the vocabulary $V$. This value can be read  “read” simply by taking the denominator of the softmax at timestep 2. Please remember this term.
\end{itemize}

It is better to indicate them as energies (since they are not probabilities), and given their logarithmic properties, we obtain a difference. We use the notation $l$ for logits and $m$ for marginalization.

Now, \tbf{when we go across steps and we connect two-time steps:}

$\text{\tbf{step 1:}}\quad-\log p(\mathbf{x}_1|\mathbf{x}_0) = -\log \frac{p(\mathbf{x}_1,\mathbf{x}_0)}{p(\mathbf{x}_0)} = E^{\ell}(\mathbf{x}_1,\mathbf{x}_0)-E^m(\mathbf{x}_0).$

We see that at timestep 1, the value $E^{\ell}(\mathbf{x}_1,\mathbf{x}_0)$ \tbf{appears again, but measured at the logit level.}

In other words, across the time-steps 2 and 1, the quantity $E(\mathbf{x}_1,\mathbf{x}_0)$ is measured twice:

\begin{itemize}
    \item at timestep 2, as the marginalization;
    \item at timestep 1, as the logit.
\end{itemize}

In the architecture or in the loss, there is no mechanism that forces these quantities to be the same, but they should be equal, given the language modeling objective. This is the same as saying that in \cref{eq:cond-prob-exp}, the probabilities across time steps need to cancel out as shown.

In other words, the following:

$$p(\mathbf{x}_2, \mathbf{x}_1, \mathbf{x}_0) = p(\mathbf{x}_2|\mathbf{x}_1,\mathbf{x}_0)p(\mathbf{x}_1|\mathbf{x}_0)p(\mathbf{x}_0)$$

Implies:

$$E(\mathbf{x}_2,\mathbf{x}_1,\mathbf{x}_0) = E^{\ell}(\mathbf{x}_2,\mathbf{x}_1,\mathbf{x}_0)~\underbrace{- E^m(\mathbf{x}_1,\mathbf{x}_0) + E^{\ell}(\mathbf{x}_1,\mathbf{x}_0)}_{\text{should be zero}}~\underbrace{- E^m(\mathbf{x}_0) + E^{\ell}(\mathbf{x}_0)}_{\text{should be zero}}$$

To model the energy of a sequence $E^{\ell}(\mathbf{x}_2,\mathbf{x}_1,\mathbf{x}_0)$ correctly, then:

\begin{itemize}
    \item $- E^m(\mathbf{x}_1,\mathbf{x}_0) + E^{\ell}(\mathbf{x}_1,\mathbf{x}_0) = 0$ (spilled energy at timestep 2 if non-zero)
    \item $-E^m(\mathbf{x}_0) + E^{\ell}(\mathbf{x}_0) = 0$ (spilled energy at timestep 1 if non-zero)
\end{itemize}

so that $E(\mathbf{x}_2,\mathbf{x}_1,\mathbf{x}_0) = E^{\ell}(\mathbf{x}_2,\mathbf{x}_1,\mathbf{x}_0)$.

\newpage
\section{Reproducibility} \label{appendix:reproducibility}

For comparisons on real-world tasks, we adopt the same experimental setting as \citet{orgad2024llms}, whose implementation is publicly available at \url{https://github.com/technion-cs-nlp/LLMsKnow}. 
This ensures that our baselines and evaluation procedures follow an established and validated protocol. 

In addition, we release our codebase, which includes:
\begin{itemize}
    \item computation of the proposed energy-based measures;
    \item scripts for reproducing the synthetic arithmetic preliminary experiments;
    \item example of how our method can be integrated into a benchmarking or production pipeline.
\end{itemize}

\subsection{Exact Answer Token Detection Details}

To analyze the \textbf{spilled energy} specifically on the tokens carrying the semantic weight of the answer, we must first localize the "exact answer" span $[u, w]$ within the longer generated sequence $\hat{y}$. We adopt the methodology proposed by \citet{orgad2024llms}, utilizing a combination of heuristics and an auxiliary instruction-tuned LLM to perform this extraction.

\paragraph{Extraction Strategy}
Depending on the nature of the task, we employ two strategies to identify the exact answer substring $s$:
\begin{itemize}
    \item \textbf{Heuristic Matching:} For tasks with a closed set of possible labels (e.g., classification tasks or multiple-choice QA), we perform string matching to locate the label within the generation.
    \item \textbf{LLM-based Extraction:} For open-ended generation tasks (e.g., TriviaQA, Math), where the answer form varies, we employ an instruction-tuned model (Mistral-7B-Instruct) to extract the short answer from the long-form generation.
\end{itemize}

\paragraph{Prompting for Extraction}
Following \cite{orgad2024llms}, we prompt the auxiliary model with the original question $q$ and the generated long answer $\hat{y}$ using the following template:

\begin{tcolorbox}[colback=gray!10, colframe=gray!60, arc=2mm, boxrule=0.5pt, title=Prompt for Exact Answer Extraction]
\small
\texttt{Extract from the following long answer the short answer, only the relevant tokens. If the long answer does not answer the question, output NO ANSWER.}

\vspace{0.5em}

\texttt{Q: [Question 1]} \\
\texttt{A: [LLM long answer 1]} \\
\texttt{Exact answer: [Short exact answer 1]}

\vspace{0.5em}

\texttt{Q: [Question 2]} \\
\texttt{A: [LLM long answer that does not answer the question]} \\
\texttt{Exact answer: NO ANSWER}

\vspace{0.5em}

\texttt{Q: [Question]} \\
\texttt{A: [LLM long answer]} \\
\texttt{Exact answer:}
\end{tcolorbox}

\paragraph{Verification and Token Mapping}
To ensure robustness, we verify that the extracted string $s$ is a valid substring of the original generation $\hat{y}$. If the extraction is invalid or the model outputs "\texttt{NO ANSWER}," we retry the extraction up to five times. If a valid substring is still not found, the sample is excluded from the analysis to avoid identifying incorrect tokens.

Once the substring $s$ is validated, we map it to the corresponding token indices $[u, w]$ in the original sequence. The spilled energy analysis is then performed specifically over this interval, or pooled across it (e.g., via min-pooling) as described in \cref{sec:spilled-real-world}.

\begin{table}[ht]
    \centering
    \caption{Answer Extraction Success Rate across tasks for Mistral-Instruct.}
    \label{tab:extraction_success_rate}
    \begin{tabular}{l|c}
        \hline
        \textbf{Dataset} & \textbf{Success Rate (\%)} \\
        \hline
        TriviaQA & 90.29 \\
        HotpotQA & 87.37 \\
        Movies & 93.61 \\
        MNLI & 92.99 \\
        Math & 87.59 \\
        HotpotQA-WC & 92.38 \\
        \hline
    \end{tabular}
\end{table}

\paragraph{Answer Extraction Performance}
For answer localization, we achieve accuracy comparable to the results of \citet{orgad2024llms}. We report in \cref{tab:extraction_success_rate} the extraction success rate across the full datasets using Mistral-7B-Instruct. Note that some datasets have been excluded (e.g., IMDB, Winobias, Winogrande) since they have a finite set of possible answers that can be used to easily locate the exact answer within the model’s generation.

\section{LLM Usage}
Large language models were used exclusively for text polishing and minor exposition refinements. All substantive research content, methodology, and scientific conclusions were developed entirely by the authors.

\newpage
\section{Supplementary Material}\label{sec:suppmat}
This supplementary material is intended to complement the main paper by providing further motivation for our assumptions and design choices, as well as additional ablation studies or plots, such as ROCs and histograms, that could not fit in the main paper.

\subsection{Additional results for Synthetic Arithmetic}
\begin{figure}[bh]
\centering
\vspace{0.8em}
{Mistral-7B-Instruct v0.3}\\[0.8em]
\begin{subfigure}{0.24\textwidth}
  \tikz[remember picture,baseline] \node (tripletNW) {};
  \includegraphics[width=\linewidth]{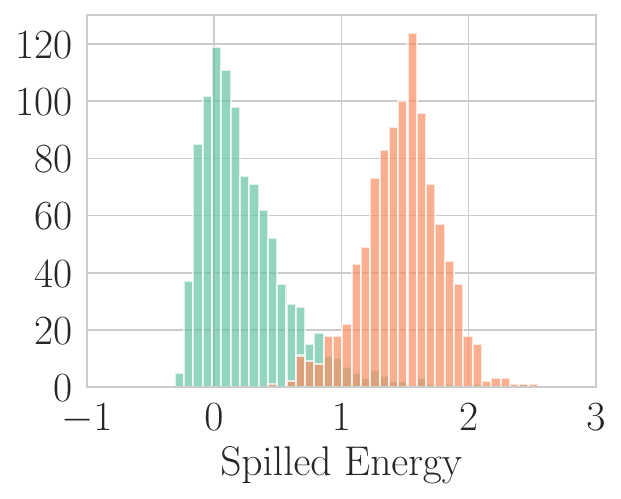}
  \caption{Easy}
  \label{fig:synth-math-mistral-large-range}
\end{subfigure}\hfill
\begin{subfigure}{0.24\textwidth}
  \tikz[remember picture,baseline] \node (tripletN) {};
  \includegraphics[width=\linewidth]{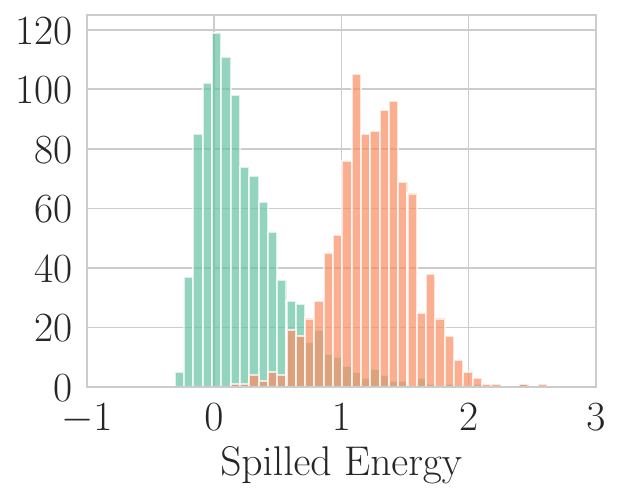}
  \caption{Medium}
  \label{fig:synth-math-mistral-mid-range}
\end{subfigure}\hfill
\begin{subfigure}{0.24\textwidth}
  \tikz[remember picture,baseline] \node (tripletNE) {};
  \includegraphics[width=\linewidth]{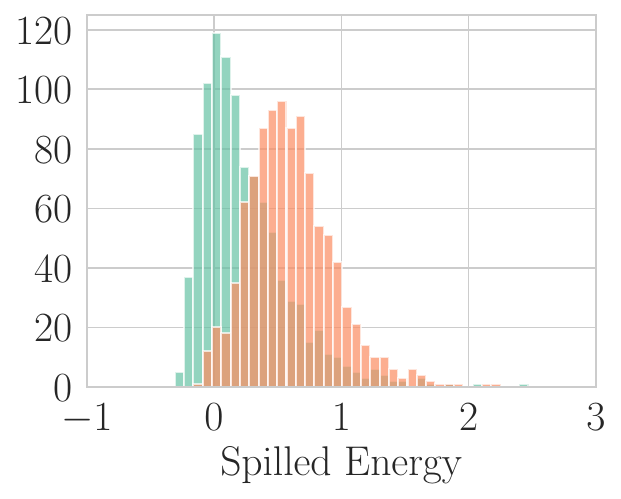}
  \caption{Hard}
  \label{fig:synth-math-mistral-small-range}
\end{subfigure}
\begin{subfigure}{0.24\textwidth}
  \tikz[remember picture,baseline] \node (rocAnchor) {};
  \includegraphics[width=\linewidth]{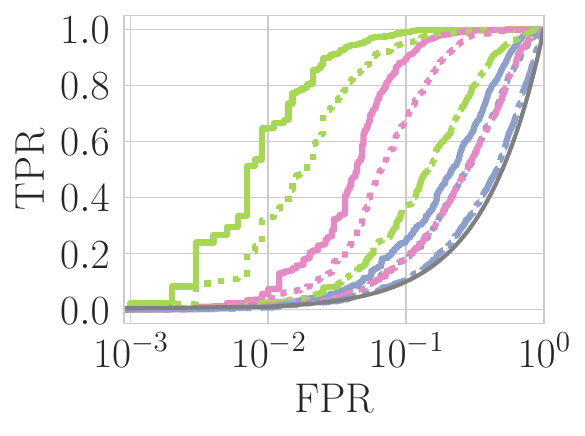}
  \caption{ROC}
  \label{fig:synth-math-mistral-roc}
\end{subfigure}

\begin{tikzpicture}[remember picture,overlay]
  \path let
    \p1 = (tripletNW),  
    \p2 = (tripletNE)   
  in node[anchor=center, yshift=16pt, xshift=-6pt,
          fill=white, draw=black, rounded corners=1pt, inner sep=3pt]
     at ($(\x1,\y1)!0.5!(\x2,\y2)$) 
     {\footnotesize
       \legendcolor{green}{Correct}\quad
       \legendcolor{orange}{Incorrect}\quad
     };
\end{tikzpicture}

\begin{tikzpicture}[remember picture,overlay]
  \node[anchor=north, xshift=16pt, yshift=37pt,
        fill=white, draw=black, rounded corners=1pt, inner sep=4pt]
       at (rocAnchor.north) 
       {\footnotesize
         \begin{tabular}{@{}l l@{}}
           \legendcolor{light-green}{Spilled Energy} &
           \legendline{black}{solid}{Easy} \\
           \legendcolor{pink}{Logit Energy} &
           \legendline{black}{dash pattern=on 3pt off 2pt}{Medium} \\
           \legendcolor{purple}{Marginal Energy} &
           \legendline{black}{densely dotted}{Hard} \\
         \end{tabular}
       };
\end{tikzpicture}
{Llama-3 8B}\\[0.3em]
\begin{subfigure}{0.24\textwidth}
  \includegraphics[width=\linewidth]{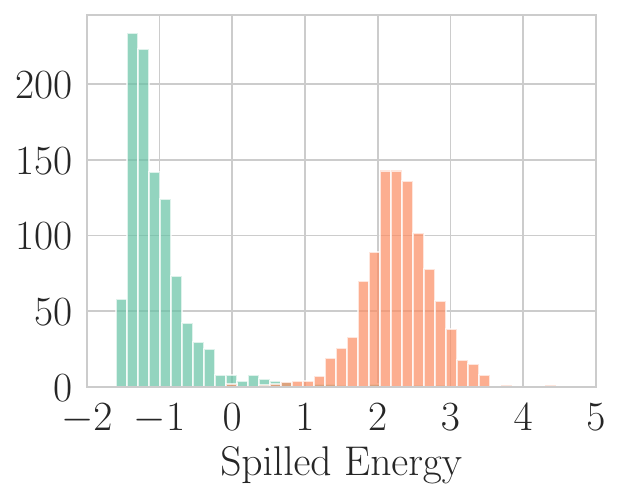}
  \caption{Easy}
  \label{fig:synth-math-llama-large-range}
\end{subfigure}
\begin{subfigure}{0.24\textwidth}
  \includegraphics[width=\linewidth]{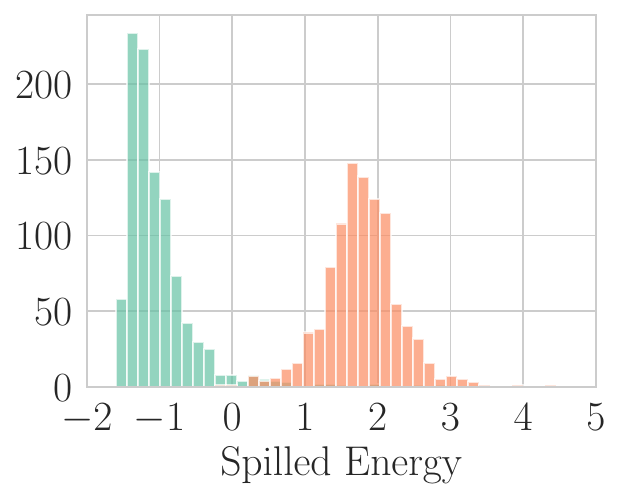}
  \caption{Medium}
  \label{fig:synth-math-llama-mid-range}
\end{subfigure}
\begin{subfigure}{0.24\textwidth}
  \includegraphics[width=\linewidth]{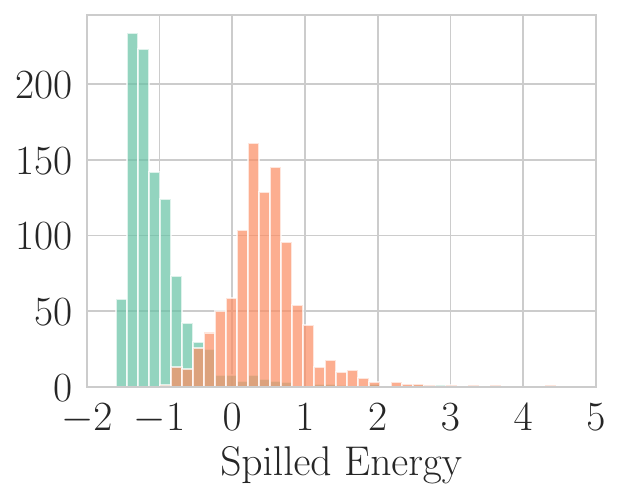}
  \caption{Hard}
  \label{fig:synth-math-llama-small-range}
\end{subfigure}
\begin{subfigure}{0.25\textwidth}
  \tikz[remember picture,baseline] \node (rocAnchor) {};
  \includegraphics[width=\linewidth]{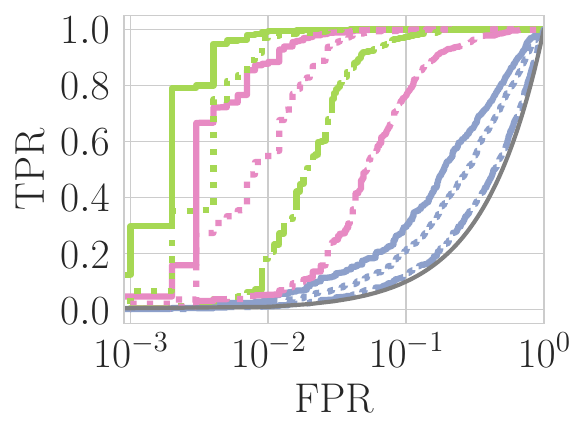}
  \caption{ROC}
  \label{fig:synth-math-llama-roc}
\end{subfigure}

\caption{Histograms of Spilled Energy values across models (rows) on Math Sums with different error ranges in the answer (columns, decreasing range left to right, making it harder to detect errors), as described in \cref{sec:synth-arithmetic}. In the fourth column, we show ROC curves for Hallucination Detection across the error ranges (colors) and methods (line styles).}
\label{fig:spilled-synth-arithmetic-add}
\end{figure}
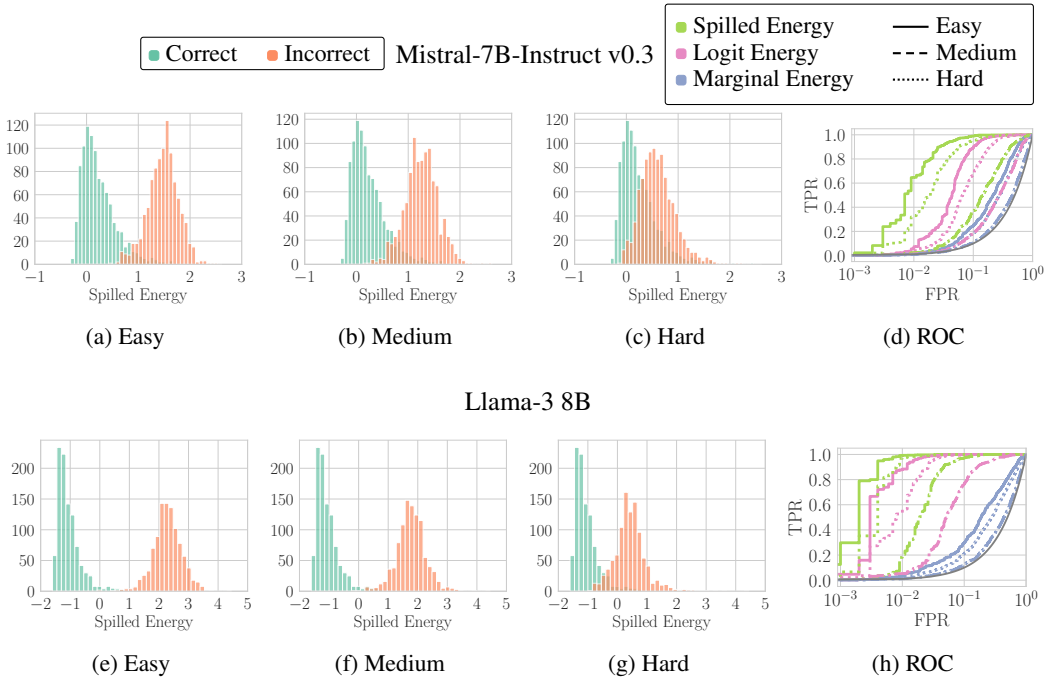

In \cref{fig:spilled-synth-arithmetic-add} we augmented \cref{fig:spilled-synth-arithmetic} in the main paper, also adding the results for \tbf{Mistral-7B-Instruct v0.3} and \tbf{LLaMa-3-8B}. The same findings of the figure in the paper also translate to this LLM, meaning that our method generalizes across LLMs.

\begin{figure}[]
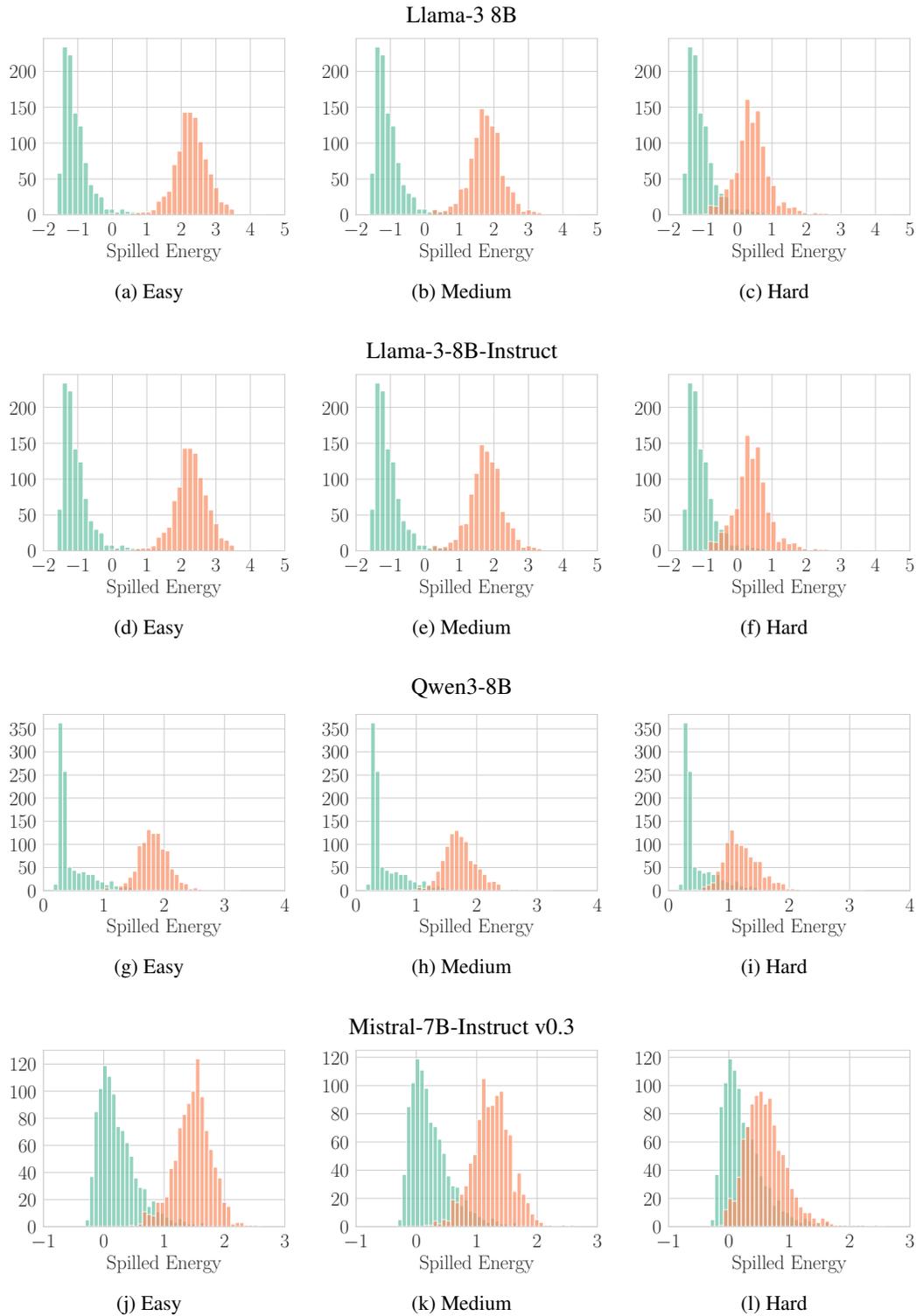

\centering

{Llama-3 8B}\\[0.3em]
\begin{subfigure}{0.32\textwidth}
  \includegraphics[width=\linewidth]{figs/synth_maths/Meta-Llama-3-8B/delta_energy_correct_incorrect_answers_histogram_1e13_1000_10000.pdf}
  \caption{Easy}
  \label{fig:llama-large-range-hist}
\end{subfigure}\hfill
\begin{subfigure}{0.32\textwidth}
  \includegraphics[width=\linewidth]{figs/synth_maths/Meta-Llama-3-8B/delta_energy_correct_incorrect_answers_histogram_1e13_100_1000.pdf}
  \caption{Medium}
  \label{fig:llama-mid-range-hist}
\end{subfigure}\hfill
\begin{subfigure}{0.32\textwidth}
  \includegraphics[width=\linewidth]{figs/synth_maths/Meta-Llama-3-8B/delta_energy_correct_incorrect_answers_histogram_1e13_1_10.pdf}
  \caption{Hard}
  \label{fig:llama-small-range-hist}
\end{subfigure}

\vspace{1.2em}

{Llama-3-8B-Instruct}\\[0.3em]
\begin{subfigure}{0.32\textwidth}
  \includegraphics[width=\linewidth]{figs/synth_maths/Meta-Llama-3-8B/delta_energy_correct_incorrect_answers_histogram_1e13_1000_10000.pdf}
  \caption{Easy}
  \label{fig:llama-inst-large-range-hist}
\end{subfigure}\hfill
\begin{subfigure}{0.32\textwidth}
  \includegraphics[width=\linewidth]{figs/synth_maths/Meta-Llama-3-8B/delta_energy_correct_incorrect_answers_histogram_1e13_100_1000.pdf}
  \caption{Medium}
  \label{fig:llama-inst-mid-range-hist}
\end{subfigure}\hfill
\begin{subfigure}{0.32\textwidth}
  \includegraphics[width=\linewidth]{figs/synth_maths/Meta-Llama-3-8B/delta_energy_correct_incorrect_answers_histogram_1e13_1_10.pdf}
  \caption{Hard}
  \label{fig:llama-inst-small-range-hist}
\end{subfigure}

\vspace{1.2em}

{Qwen3-8B}\\[0.3em]
\begin{subfigure}{0.32\textwidth}
  \includegraphics[width=\linewidth]{figs/synth_maths/Qwen3-8B/delta_energy_correct_incorrect_answers_histogram_1e13_1000_10000.pdf}
  \caption{Easy}
  \label{fig:qwen-large-range-hist}
\end{subfigure}\hfill
\begin{subfigure}{0.32\textwidth}
  \includegraphics[width=\linewidth]{figs/synth_maths/Qwen3-8B/delta_energy_correct_incorrect_answers_histogram_1e13_100_1000.pdf}
  \caption{Medium}
  \label{fig:qwen-mid-range-hist}
\end{subfigure}\hfill
\begin{subfigure}{0.32\textwidth}
  \includegraphics[width=\linewidth]{figs/synth_maths/Qwen3-8B/delta_energy_correct_incorrect_answers_histogram_1e13_1_10.pdf}
  \caption{Hard}
  \label{fig:qwen-small-range-hist}
\end{subfigure}

\vspace{1.2em}

{Mistral-7B-Instruct v0.3}\\[0.3em]
\begin{subfigure}{0.32\textwidth}
  \includegraphics[width=\linewidth]{figs/synth_maths/Mistral-7B-Instruct-v0.3/delta_energy_correct_incorrect_answers_histogram_1e13_1000_10000.pdf}
  \caption{Easy}
  \label{fig:mistral-large-range-hist}
\end{subfigure}\hfill
\begin{subfigure}{0.32\textwidth}
  \includegraphics[width=\linewidth]{figs/synth_maths/Mistral-7B-Instruct-v0.3/delta_energy_correct_incorrect_answers_histogram_1e13_100_1000.pdf}
  \caption{Medium}
  \label{fig:mistral-mid-range-hist}
\end{subfigure}\hfill
\begin{subfigure}{0.32\textwidth}
  \includegraphics[width=\linewidth]{figs/synth_maths/Mistral-7B-Instruct-v0.3/delta_energy_correct_incorrect_answers_histogram_1e13_1_10.pdf}
  \caption{Hard}
  \label{fig:mistral-small-range-hist}
\end{subfigure}

\caption{Histograms of Spilled Energy values for \textcolor{green-correct}{\rule{0.45cm}{0.15cm}} \texttt{Correct} and \textcolor{orange-incorrect}{\rule{0.45cm}{0.15cm}} \texttt{Incorrect} answers across models on Math Sums, increasing difficulty from left to right. We compute sums on 13-digit integers, for incorrect answers we add a random offset sampled uniformly from the error interval: Easy $\sim \mathcal{U}(1e3, 1e4)$ - Medium $\sim \mathcal{U}(1e2, 1e3)$ - Hard $\sim \mathcal{U}(1, 10)$; for more details see \cref{sec:synth-arithmetic}.}
\label{fig:models-vs-datasets-hist}
\end{figure}

\begin{figure}[]
\centering
{Llama-3-8B}\\[0.3em]
\begin{subfigure}{0.33\textwidth}
  \includegraphics[width=\linewidth]{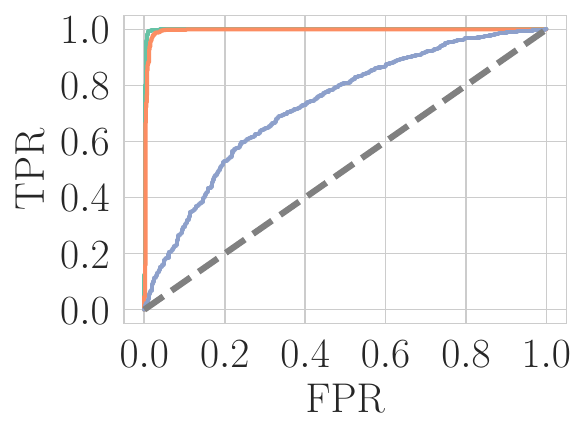}
  \caption{Easy}
  \label{fig:llama-large-range}
\end{subfigure}
\begin{subfigure}{0.33\textwidth}
  \includegraphics[width=\linewidth]{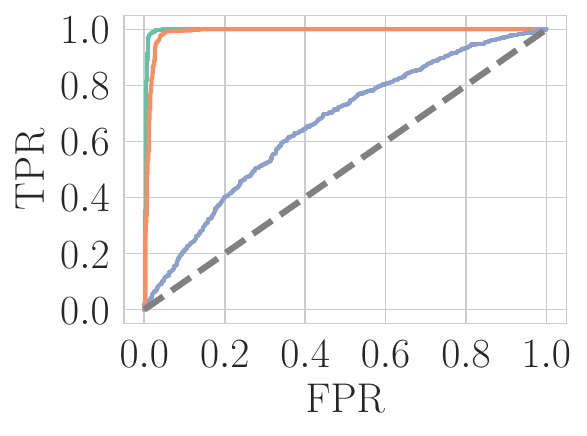}
  \caption{Medium}
  \label{fig:llama-mid-range}
\end{subfigure}
\begin{subfigure}{0.33\textwidth}
  \includegraphics[width=\linewidth]{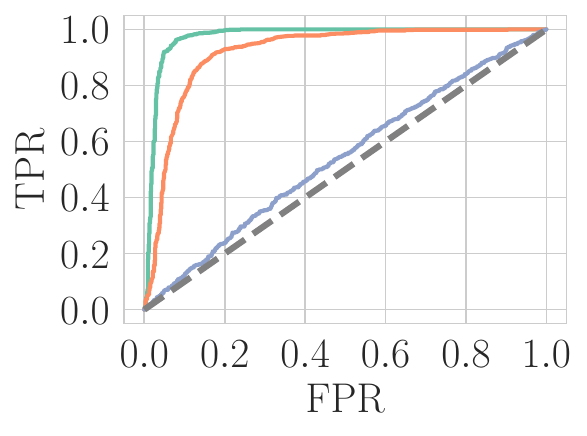}
  \caption{Hard}
  \label{fig:llama-small-range}
\end{subfigure}

\vspace{1.2em}

{Llama-3-8B-Instruct}\\[0.3em]
\begin{subfigure}{0.33\textwidth}
  \includegraphics[width=\linewidth]{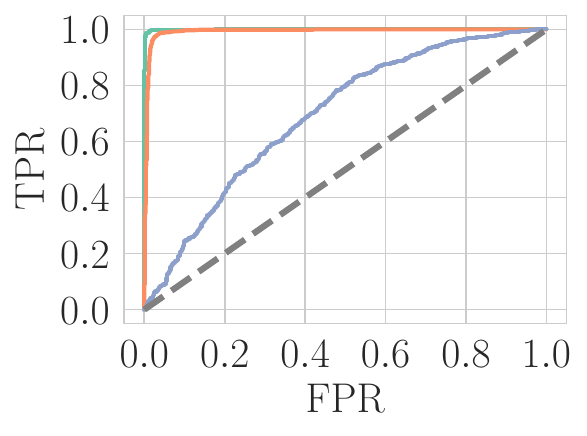}
  \caption{Easy}
  \label{fig:llama-inst-large-range}
\end{subfigure}
\begin{subfigure}{0.33\textwidth}
  \includegraphics[width=\linewidth]{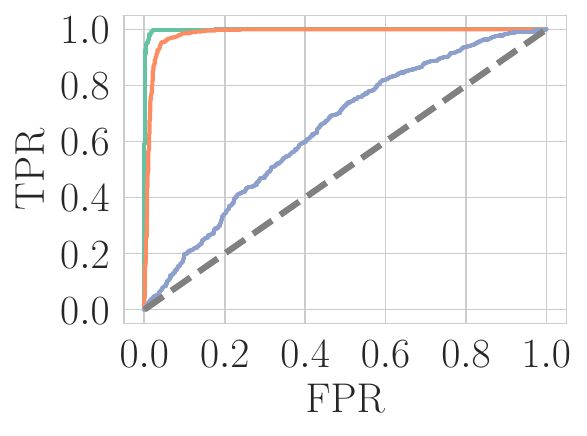}
  \caption{Medium}
  \label{fig:llama-inst-mid-range}
\end{subfigure}
\begin{subfigure}{0.33\textwidth}
  \includegraphics[width=\linewidth]{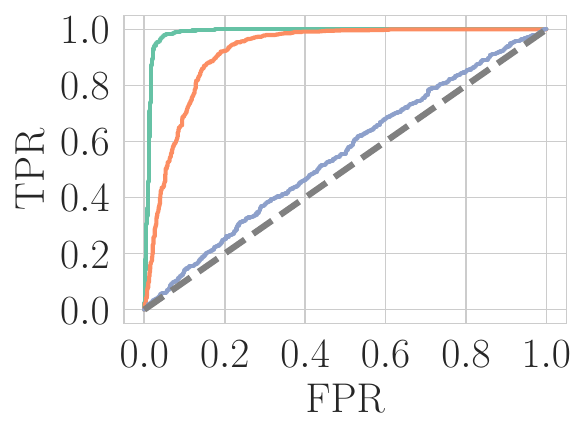}
  \caption{Hard}
  \label{fig:llama-inst-small-range}
\end{subfigure}

\vspace{1.2em}

{Qwen-3 8B}\\[0.3em]
\begin{subfigure}{0.33\textwidth}
  \includegraphics[width=\linewidth]{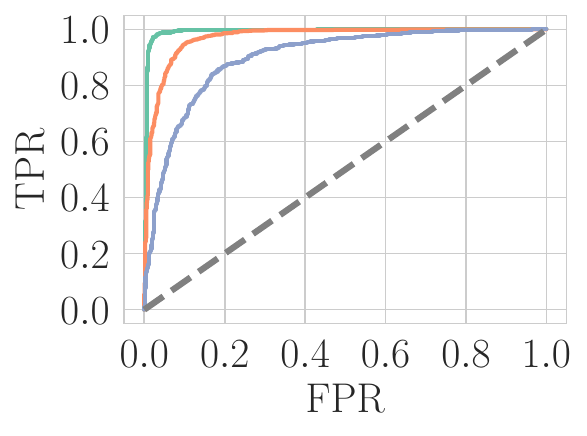}
  \caption{Easy}
  \label{fig:qwen-large-range}
\end{subfigure}
\begin{subfigure}{0.33\textwidth}
  \includegraphics[width=\linewidth]{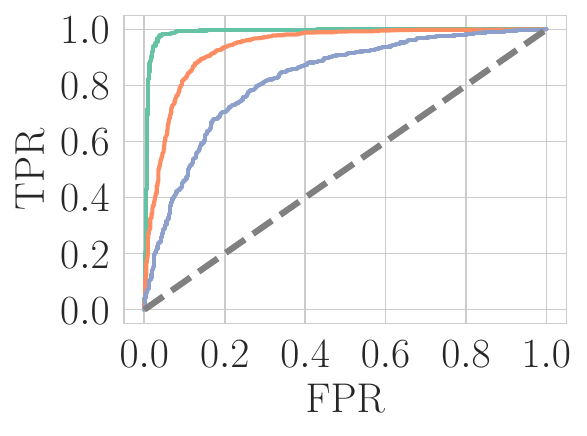}
  \caption{Medium}
  \label{fig:qwen-mid-range}
\end{subfigure}
\begin{subfigure}{0.33\textwidth}
  \includegraphics[width=\linewidth]{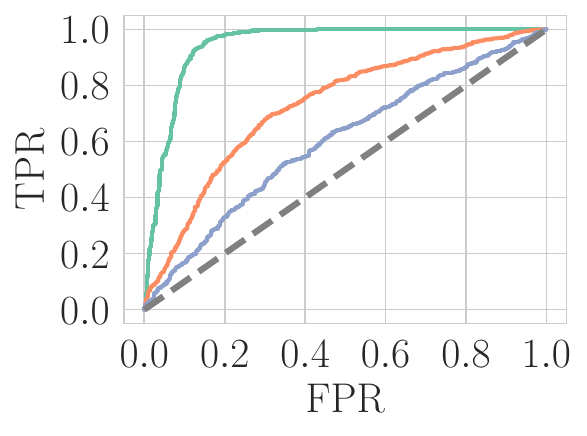}
  \caption{Hard}
  \label{fig:qwen-small-range}
\end{subfigure}

\vspace{1.2em}

{Mistral-7B-Instruct v0.3}\\[0.3em]
\begin{subfigure}{0.33\textwidth}
  \includegraphics[width=\linewidth]{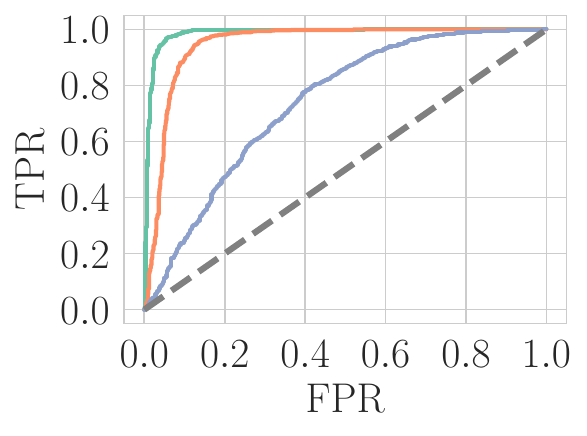}
  \caption{Easy}
  \label{fig:mistral-large-range}
\end{subfigure}
\begin{subfigure}{0.33\textwidth}
  \includegraphics[width=\linewidth]{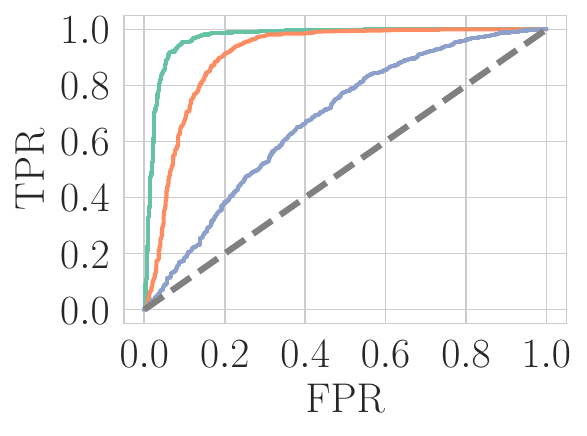}
  \caption{Medium}
  \label{fig:mistral-mid-range}
\end{subfigure}
\begin{subfigure}{0.33\textwidth}
  \includegraphics[width=\linewidth]{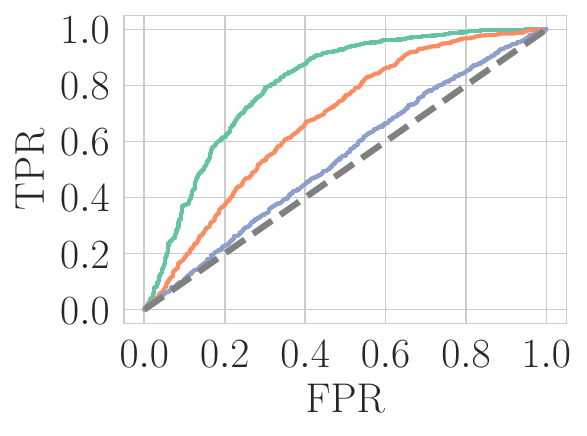}
  \caption{Hard}
  \label{fig:mistral-small-range}
\end{subfigure}

\caption{ROC curves for Hallucination Detection across models (rows) on Math Sums with different error ranges in the answer (columns, decreasing range left to right). All sums are performed on 13-digit integers. Legend: \textcolor{green-correct}{\rule{0.45cm}{0.15cm}} \tbf{Spilled (ours)} Spilled $\Delta E$  \textcolor{orange-incorrect}{\rule{0.45cm}{0.15cm}} Logit $E^\ell$ \textcolor{violet-marginal}{\rule{0.45cm}{0.15cm} } Marginal $E^m$ }
\label{fig:models-vs-datasets}
\end{figure}

\cref{fig:models-vs-datasets-hist} and \cref{fig:models-vs-datasets} also extend and provide more details of \cref{fig:spilled-synth-arithmetic} in the main paper by showing, respectively, the histograms and the ROC at a better resolution and displayed in different frames. Also, we have added results for Mistral-7B-Instruct v0.3 and LLaMa-3-8B.

\subsection{Additional Qualitative Results} \label{sec:supp-qual}
In this section, we offer additional results of the detection performance following what is shown in \cref{fig:teaser}. We report both success cases and failure cases. While it is difficult to draw conclusions and predict when, why, and on which topics spilled energy may work or not, we noticed that it appears to perform reliably on knowledge-based factual content but, at times, exhibits difficulties with reasoning tasks and numerical information, despite working well on math questions, as demonstrated in \cref{sec:synth-arithmetic}. Further investigation is required to better understand and validate these patterns.

\begin{table}[tb]
        \centering
        \setlength{\tabcolsep}{2pt}
        \resizebox{\textwidth}{!}{
        \begin{tabular}{lccccccccccb}
        \toprule
        & Pool & HotpotQA & HotpotQA-WC & IMDB & Math & MNLI & Movies & TriviaQA & Winobias & Winogrande & Average \\
        \midrule 
        & \multicolumn {11}{c}{LLaMA-Instruct} \\\midrule 

{\footnotesize \cite{orgad2024llms}} & Mean & 66.56\stdev{9.10} & 59.00\stdev{8.14} & \textit{69.78}\stdev{14.76} & \textit{66.56}\stdev{17.04} & 60.56\stdev{12.53} & 66.44\stdev{8.06} & 63.22\stdev{11.11} & \textbf{67.33}\stdev{11.97} & \textbf{58.00}\stdev{7.79} & 64.16\stdev{3.90} \\ \midrule 
Spilled $\Delta E$ & Min & \textbf{85.98}\stdev{1.09} & \textit{93.00}\stdev{1.61} & 47.66\stdev{4.06} & 65.58\stdev{3.02} & \textit{73.95}\stdev{1.97} & \textit{89.34}\stdev{1.04} & \textbf{87.07}\stdev{1.33} & \textit{60.72}\stdev{2.74} & \textit{55.11}\stdev{2.05} & \textbf{73.16}\stdev{15.64} \\
Marginal $E^m$ & Max & \textit{76.72}\stdev{1.38} & 30.74\stdev{3.45} & \textbf{85.63}\stdev{2.39} & 27.08\stdev{5.06} & \textbf{89.90}\stdev{1.25} & \textbf{96.17}\stdev{0.63} & \textit{80.13}\stdev{1.87} & 57.67\stdev{2.94} & 47.47\stdev{1.83} & \textit{65.72}\stdev{24.39} \\
Marginal $E^m$ & Min & 75.91\stdev{1.62} & \textbf{97.57}\stdev{0.75} & 14.37\stdev{2.39} & \textbf{70.55}\stdev{2.43} & 61.21\stdev{3.24} & 72.21\stdev{1.60} & 73.38\stdev{1.86} & 47.19\stdev{2.71} & 53.98\stdev{2.30} & 62.93\stdev{21.89} \\
Logit $E^\ell$ & Max & 72.85\stdev{2.12} & 91.11\stdev{1.52} & 42.08\stdev{5.07} & 57.81\stdev{3.82} & 25.52\stdev{3.00} & 43.97\stdev{1.38} & 68.89\stdev{1.96} & 39.95\stdev{2.41} & 49.40\stdev{2.16} & 54.62\stdev{18.97} \\
Spilled $\Delta E$ & Max & 54.34\stdev{1.58} & 47.68\stdev{2.81} & 52.34\stdev{4.06} & 40.33\stdev{3.05} & 56.44\stdev{2.81} & 68.56\stdev{1.87} & 47.54\stdev{2.40} & 38.40\stdev{2.61} & 44.97\stdev{1.51} & 50.07\stdev{8.66} \\

\midrule 
& \multicolumn {11}{c}{LLaMA} \\\midrule 

{\footnotesize \cite{orgad2024llms}} & Mean & 61.22\stdev{9.95} & 56.78\stdev{8.70} & \textbf{72.67}\stdev{13.91} & 69.67\stdev{15.07} & 60.33\stdev{13.77} & 64.00\stdev{8.40} & 66.44\stdev{8.20} & \textbf{60.89}\stdev{12.60} & \textbf{53.56}\stdev{4.36} & 62.84\stdev{5.71} \\ \midrule 
Logit $E^\ell$ & Min & \textbf{87.93}\stdev{1.01} & \textbf{91.24}\stdev{0.80} & 51.73\stdev{1.32} & 42.99\stdev{5.68} & 97.01\stdev{0.43} & \textbf{99.86}\stdev{0.16} & \textbf{84.53}\stdev{0.87} & 49.29\stdev{1.46} & 48.52\stdev{1.78} & \textbf{72.57}\stdev{22.36} \\
Spilled $\Delta E$ & Min & \textit{79.04}\stdev{1.78} & \textit{80.83}\stdev{1.87} & 43.22\stdev{1.67} & \textbf{74.36}\stdev{5.54} & \textbf{99.97}\stdev{0.08} & 61.97\stdev{2.81} & \textit{78.54}\stdev{1.57} & 52.11\stdev{2.58} & 48.21\stdev{1.62} & \textit{68.69}\stdev{17.48} \\
Spilled $\Delta E_{s}$ & Min & 77.75\stdev{1.52} & 79.44\stdev{2.05} & 43.39\stdev{1.82} & \textit{72.87}\stdev{6.10} & \textbf{99.97}\stdev{0.08} & 61.56\stdev{2.95} & 77.55\stdev{1.62} & \textit{52.34}\stdev{2.57} & 48.17\stdev{1.62} & 68.12\stdev{17.15} \\
Marginal $E^m$ & Max & 78.00\stdev{1.30} & 76.90\stdev{1.09} & 48.29\stdev{1.16} & 68.77\stdev{8.33} & 10.93\stdev{1.42} & \textit{80.70}\stdev{1.98} & 67.49\stdev{1.69} & 51.91\stdev{2.32} & 51.28\stdev{2.47} & 59.36\stdev{20.69} \\
Marginal $E^m$ & Min & 58.39\stdev{2.79} & 59.20\stdev{1.95} & 51.71\stdev{1.16} & 34.13\stdev{8.78} & 97.42\stdev{0.51} & 50.37\stdev{2.43} & 69.88\stdev{1.40} & 49.05\stdev{2.20} & 49.00\stdev{2.30} & 57.68\stdev{16.75} \\
Logit $E^\ell$ & Max & 53.47\stdev{2.13} & 49.02\stdev{1.79} & 48.27\stdev{1.32} & 57.38\stdev{6.09} & 91.76\stdev{0.91} & 57.42\stdev{1.43} & 52.77\stdev{2.58} & 50.74\stdev{1.51} & 51.17\stdev{1.83} & 56.89\stdev{12.70} \\
Logit $E^\ell$ & ALT & 43.56\stdev{1.95} & 39.74\stdev{1.73} & 48.27\stdev{1.32} & 57.41\stdev{6.06} & 91.71\stdev{0.94} & 43.11\stdev{1.57} & 43.62\stdev{2.57} & 50.74\stdev{1.51} & 51.17\stdev{1.83} & 52.15\stdev{14.88} \\
Logit $E^\ell$ & Last Token & 43.56\stdev{1.95} & 39.74\stdev{1.73} & 48.27\stdev{1.32} & 57.41\stdev{6.06} & 91.71\stdev{0.94} & 43.11\stdev{1.57} & 43.62\stdev{2.57} & 50.74\stdev{1.51} & 51.17\stdev{1.83} & 52.15\stdev{14.88} \\
Marginal $E^m$ & ALT & 61.59\stdev{1.88} & 58.64\stdev{1.60} & 48.29\stdev{1.16} & 67.93\stdev{9.32} & 10.75\stdev{1.44} & 61.39\stdev{1.80} & 49.73\stdev{1.45} & 51.19\stdev{2.59} & 51.44\stdev{2.50} & 51.22\stdev{15.61} \\
Marginal $E^m$ & Last Token & 61.59\stdev{1.88} & 58.64\stdev{1.60} & 48.29\stdev{1.16} & 67.93\stdev{9.32} & 10.75\stdev{1.44} & 61.39\stdev{1.80} & 49.73\stdev{1.45} & 51.19\stdev{2.59} & 51.44\stdev{2.50} & 51.22\stdev{15.61} \\
Marginal $E^m$ & Mean & 58.27\stdev{2.50} & 58.64\stdev{1.58} & 48.29\stdev{1.16} & 68.32\stdev{8.35} & 6.12\stdev{0.70} & 66.55\stdev{3.22} & 45.67\stdev{1.38} & 51.80\stdev{2.29} & 51.29\stdev{2.46} & 50.55\stdev{17.33} \\

\midrule 
& \multicolumn {11}{c}{Mistral-Instruct} \\\midrule 

{\footnotesize \cite{orgad2024llms}} & Mean & 64.78\stdev{10.56} & 56.78\stdev{7.95} & \textbf{82.67}\stdev{11.63} & \textit{68.78}\stdev{11.43} & 64.22\stdev{12.12} & 64.89\stdev{11.55} & 65.44\stdev{12.10} & \textbf{61.00}\stdev{12.23} & \textbf{61.44}\stdev{11.31} & 65.56\stdev{6.84} \\ \midrule 
Spilled $\Delta E$ & Min & \textbf{91.12}\stdev{1.10} & \textit{97.47}\stdev{0.78} & 59.77\stdev{2.57} & 66.63\stdev{3.46} & \textit{95.95}\stdev{0.83} & \textbf{94.99}\stdev{0.93} & \textbf{91.75}\stdev{1.01} & 50.74\stdev{3.15} & 49.00\stdev{1.92} & \textbf{77.49}\stdev{19.42} \\
Marginal $E^m$ & Min & \textit{87.58}\stdev{1.35} & \textbf{97.94}\stdev{0.62} & 18.67\stdev{2.27} & 67.58\stdev{3.37} & \textbf{97.96}\stdev{0.55} & 84.90\stdev{1.37} & \textit{87.75}\stdev{1.73} & 49.19\stdev{3.97} & 48.49\stdev{1.86} & \textit{71.12}\stdev{25.68} \\
Logit $E^\ell$ & Max & 77.24\stdev{1.66} & 83.84\stdev{1.66} & 22.28\stdev{2.54} & 57.67\stdev{3.29} & 78.98\stdev{1.58} & 76.89\stdev{1.49} & 80.35\stdev{1.88} & 45.53\stdev{2.60} & 48.17\stdev{1.97} & 63.44\stdev{19.99} \\
Marginal $E^m$ & Max & 64.63\stdev{1.97} & 33.42\stdev{1.90} & \textit{81.33}\stdev{2.32} & 26.52\stdev{2.28} & 17.62\stdev{1.20} & \textit{86.60}\stdev{1.20} & 65.46\stdev{2.25} & \textit{56.41}\stdev{4.44} & 51.14\stdev{1.71} & 53.68\stdev{22.53} \\
Logit $E^\ell$ & Last Token & 55.77\stdev{2.38} & 71.26\stdev{2.28} & 22.28\stdev{2.54} & \textbf{71.21}\stdev{2.42} & 47.78\stdev{2.26} & 42.93\stdev{1.96} & 58.36\stdev{3.52} & 45.65\stdev{2.94} & 48.30\stdev{2.04} & 51.50\stdev{14.26} \\
Logit $E^\ell$ & ALT & 55.77\stdev{2.38} & 71.26\stdev{2.28} & 22.28\stdev{2.54} & \textbf{71.21}\stdev{2.42} & 47.78\stdev{2.26} & 42.93\stdev{1.96} & 58.36\stdev{3.52} & 45.65\stdev{2.94} & 48.30\stdev{2.04} & 51.50\stdev{14.26} \\

\midrule 
& \multicolumn {11}{c}{Mistral} \\\midrule 

{\footnotesize \cite{orgad2024llms}} & Mean & 61.78\stdev{9.27} & 57.44\stdev{6.95} & \textbf{76.22}\stdev{12.82} & 65.78\stdev{15.27} & 56.67\stdev{11.83} & 64.22\stdev{8.91} & 64.33\stdev{10.40} & 58.00\stdev{12.29} & \textbf{54.56}\stdev{4.36} & 62.11\stdev{6.21} \\ \midrule 
Marginal $E^m$ & Min & \textbf{87.52}\stdev{1.31} & \textbf{90.91}\stdev{1.58} & 54.69\stdev{2.49} & \textbf{86.21}\stdev{1.96} & \textbf{98.80}\stdev{0.35} & \textit{94.41}\stdev{0.62} & \textit{83.66}\stdev{2.16} & 52.15\stdev{1.74} & 46.37\stdev{2.02} & \textbf{77.19}\stdev{19.05} \\
Marginal $E^m$ & Max & 83.57\stdev{1.13} & \textit{86.83}\stdev{1.70} & 45.31\stdev{2.49} & 62.26\stdev{4.29} & 96.03\stdev{0.83} & \textbf{99.27}\stdev{0.24} & \textbf{92.26}\stdev{1.31} & 51.31\stdev{3.35} & \textit{54.49}\stdev{2.48} & \textit{74.59}\stdev{19.91} \\
Spilled $\Delta E$ & Min & \textit{84.24}\stdev{1.18} & 83.74\stdev{1.41} & \textit{57.43}\stdev{2.99} & \textit{78.26}\stdev{2.93} & \textit{96.69}\stdev{0.62} & 84.47\stdev{1.17} & 81.27\stdev{1.83} & 50.62\stdev{1.72} & 48.72\stdev{1.75} & 73.94\stdev{16.18} \\
Spilled $\Delta E$ & Max & 61.50\stdev{1.88} & 63.60\stdev{1.68} & 42.57\stdev{2.99} & 76.27\stdev{3.42} & 47.01\stdev{2.48} & 81.84\stdev{1.60} & 68.07\stdev{1.30} & \textbf{58.71}\stdev{3.69} & 51.13\stdev{1.87} & 61.19\stdev{12.30} \\
Spilled $\Delta E_{s}$ & Max & 60.54\stdev{1.81} & 60.18\stdev{1.84} & 43.47\stdev{2.76} & 71.93\stdev{3.62} & 45.94\stdev{2.40} & 78.84\stdev{1.53} & 67.92\stdev{1.32} & 57.24\stdev{3.72} & 51.88\stdev{1.90} & 59.77\stdev{11.08} \\

\bottomrule
\end{tabular}
}
   \caption{Hallucination detection performance, in terms of AuROC, across nine benchmarks and different LLMs. We measure the generalization across all tasks by computing the average.}
   \label{tab:detection_full_no_flip}
\end{table}

\begin{figure}[]
    \centering
    
    \begin{subfigure}{0.48\textwidth}
        \centering
        \includegraphics[width=\linewidth]{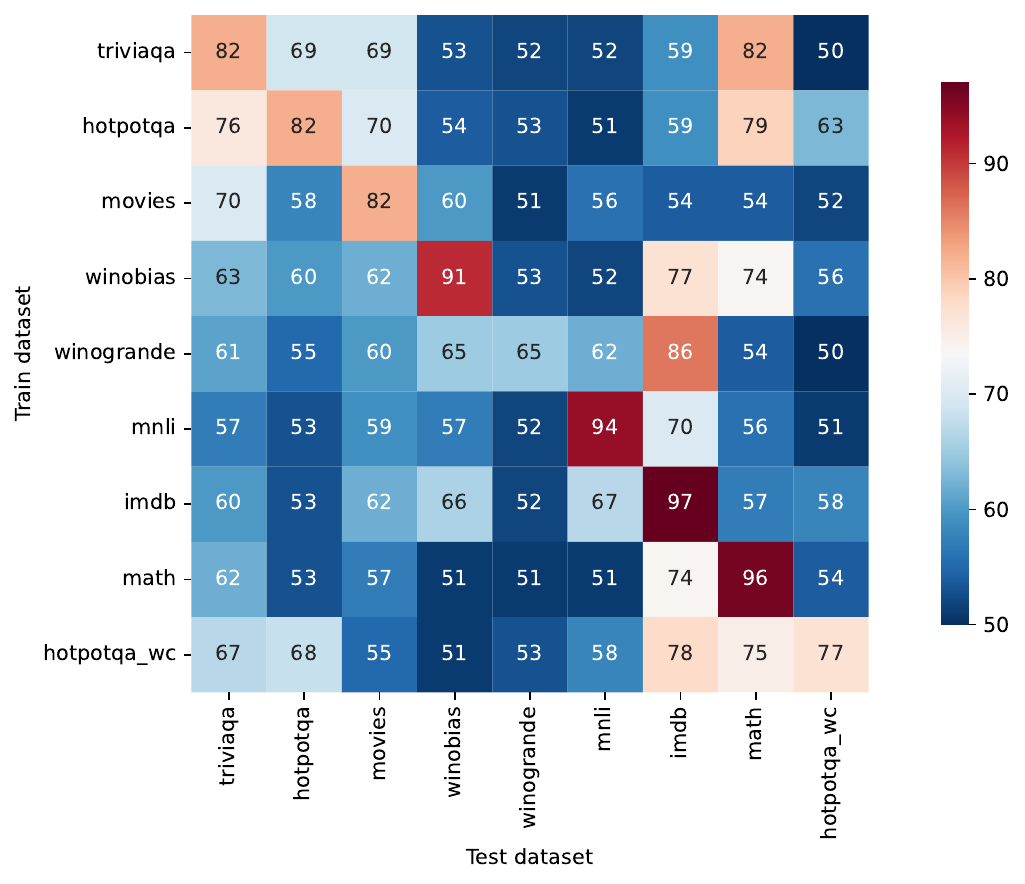}
        \caption{}
        \label{fig:orgadHeatMapsLlama}
    \end{subfigure}
    \hfill
    \begin{subfigure}{0.48\textwidth}
        \centering
        \includegraphics[width=\linewidth]{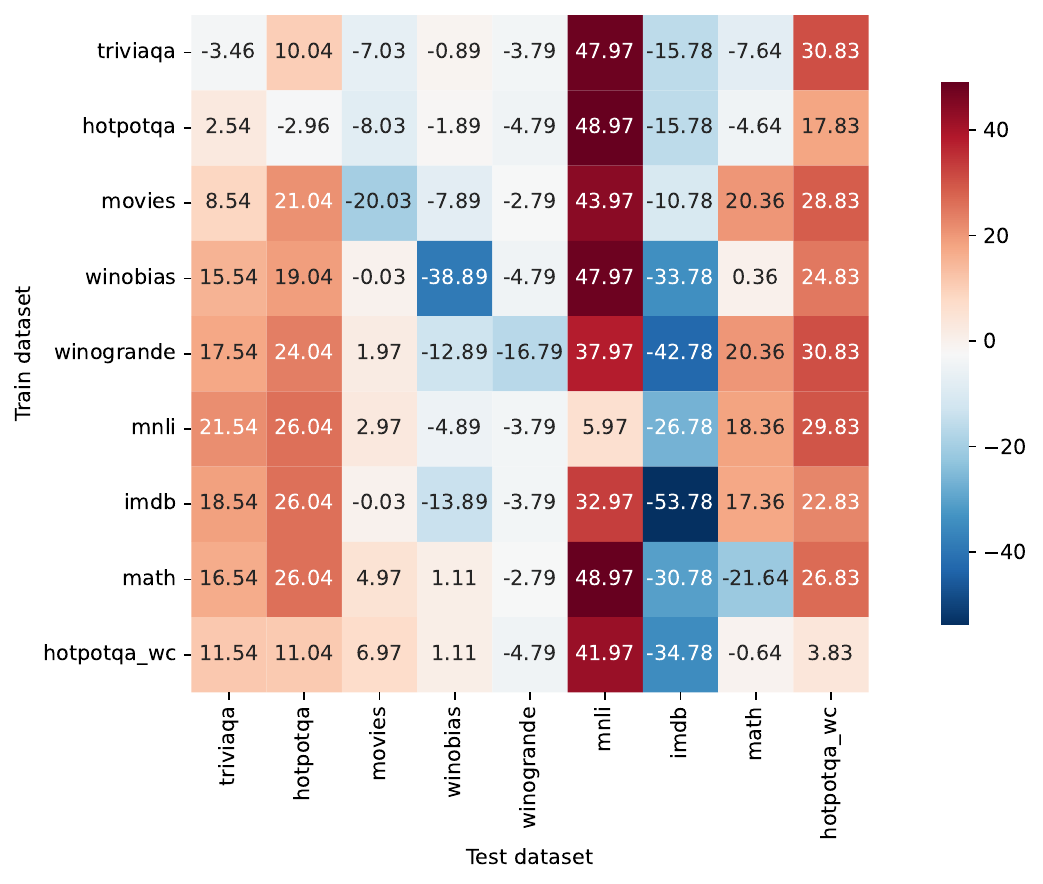}
        \caption{}
        \label{fig:diffOrgadLlama}
    \end{subfigure}

    \caption{\cref{fig:orgadHeatMapsLlama} presents the cross-dataset performance of the method proposed by \cite{orgad2024llms} using Llama-3. \cref{fig:diffOrgadLlama} depicts the performance difference between their method and our Spilled $\Delta E$ with Min pooling. Positive values indicate cases where Spilled $\Delta E$ outperforms the method of \cite{orgad2024llms}. All numbers are computed as percentages.}
    \label{fig:confusion_heatmap_LLAMA}
\end{figure}

\begin{figure}[]
    \centering
    
    \begin{subfigure}{0.48\textwidth}
        \centering
        \includegraphics[width=\linewidth]{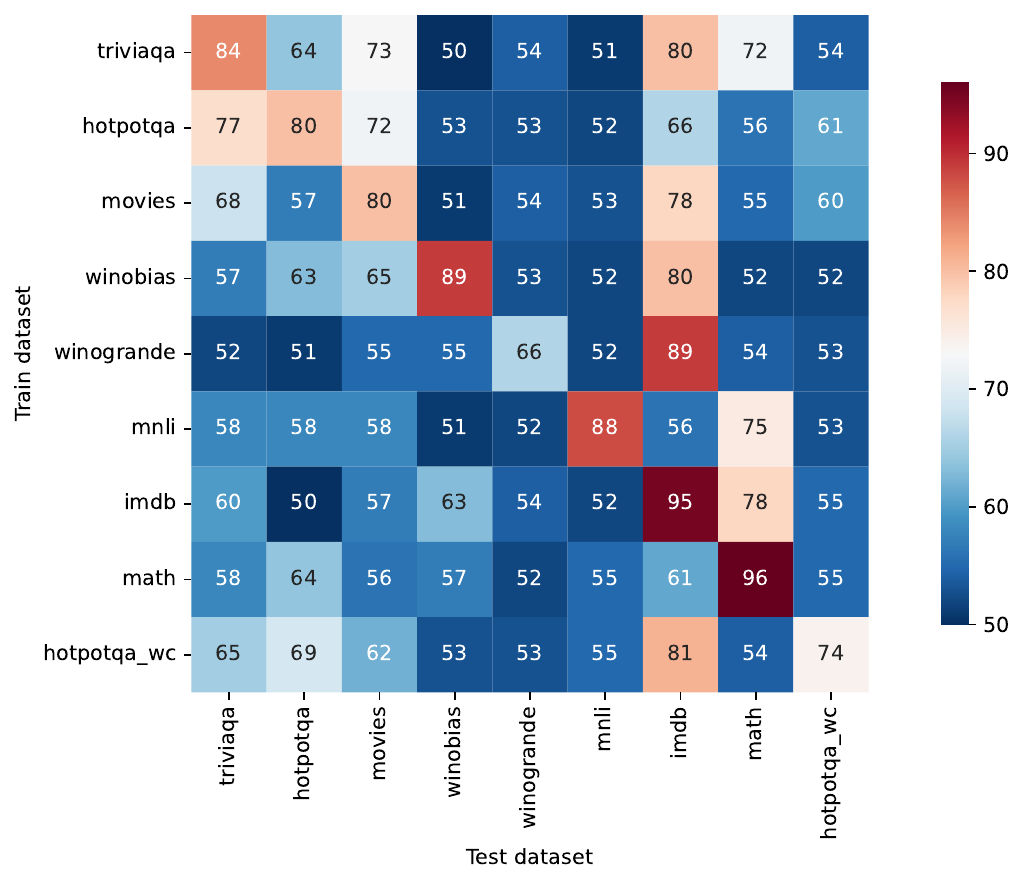}
        \caption{}
        \label{fig:orgadHeatMapsMistral}
    \end{subfigure}
    \hfill
    \begin{subfigure}{0.48\textwidth}
        \centering
        \includegraphics[width=\linewidth]{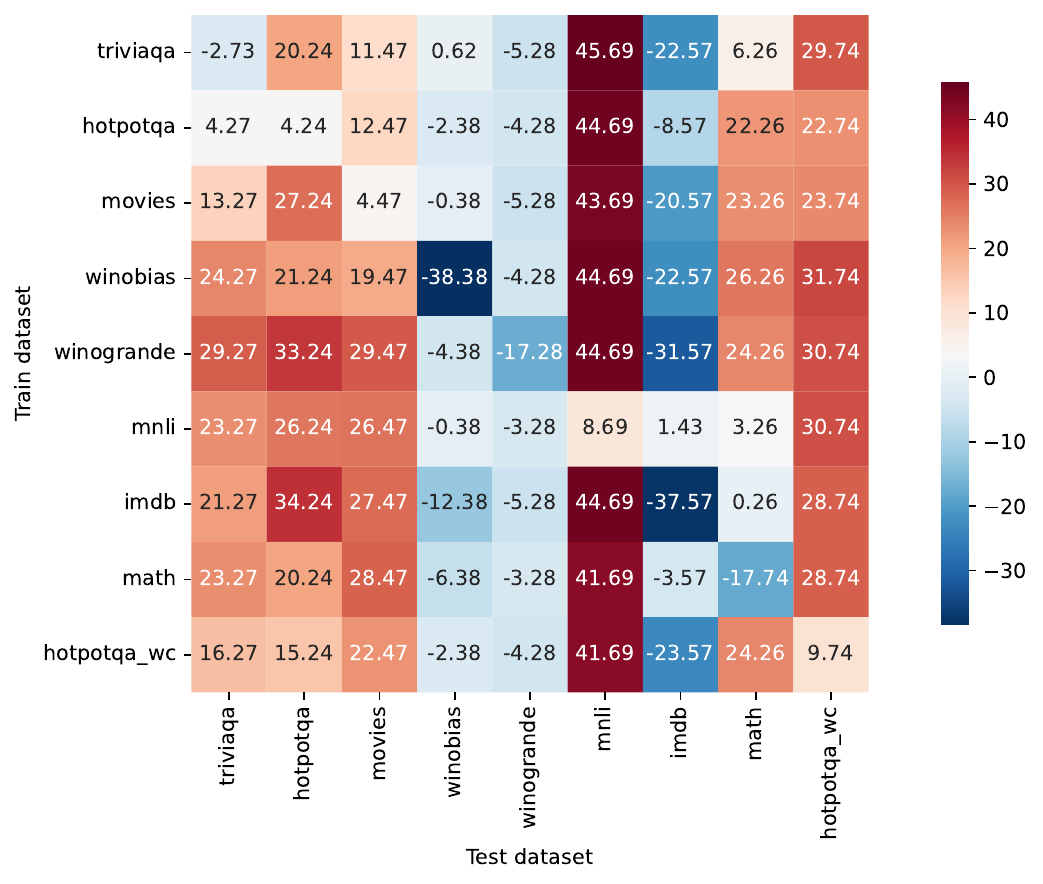}
        \caption{}
        \label{fig:diffOrgadMistral}
    \end{subfigure}

    \caption{\cref{fig:orgadHeatMapsMistral} presents the cross-dataset performance of the method proposed by \cite{orgad2024llms} using Mistral. \cref{fig:diffOrgadMistral} depicts the performance difference between their method and our Spilled $\Delta E$ with Min pooling. Positive values indicate cases where Spilled $\Delta E$ outperforms the method of \cite{orgad2024llms}. All numbers are computed as percentages.}
    \label{fig:confusion_heatmap_Mistral}
\end{figure}

\begin{figure}[]
    \centering
    
    \begin{subfigure}{0.48\textwidth}
        \centering
        \includegraphics[width=\linewidth]{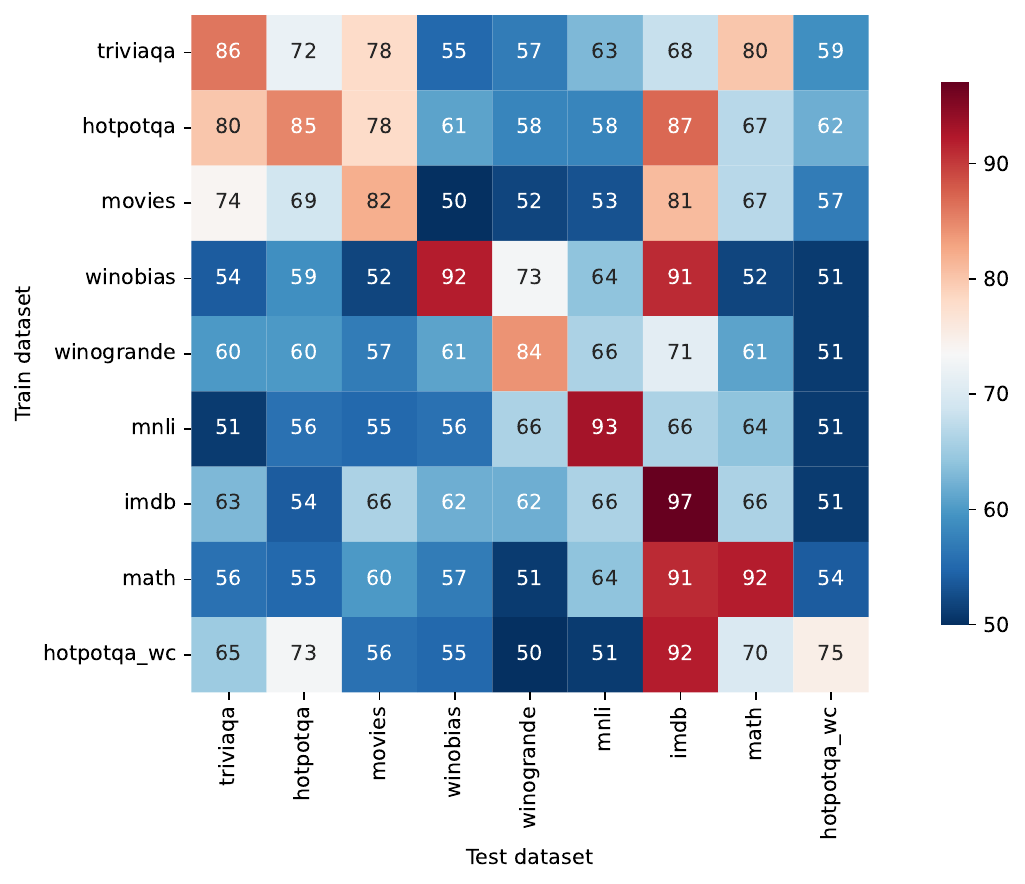}
        \caption{}
        \label{fig:orgadHeatMapsMistralInstr}
    \end{subfigure}
    \hfill
    \begin{subfigure}{0.48\textwidth}
        \centering
        \includegraphics[width=\linewidth]{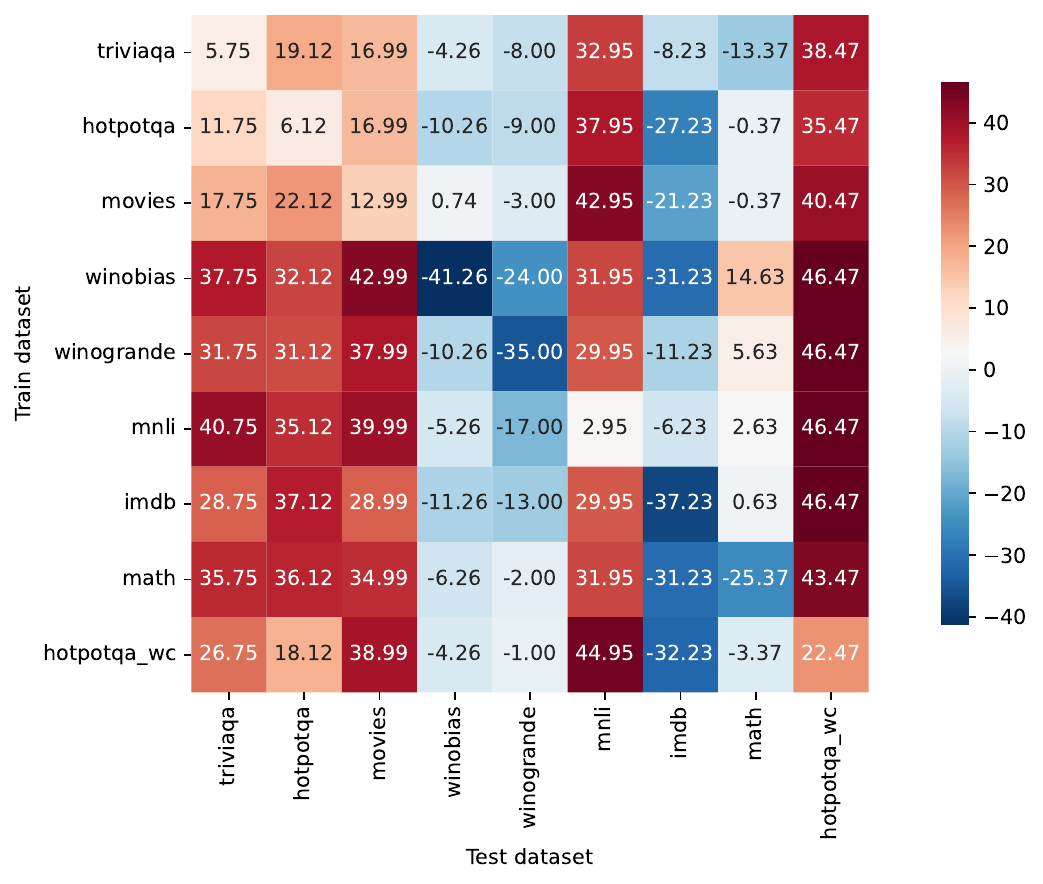}
        \caption{}
        \label{fig:diffOrgadMistralInstr}
    \end{subfigure}

    \caption{\cref{fig:orgadHeatMapsMistralInstr} presents the cross-dataset performance of the method proposed by \cite{orgad2024llms} using Mistral-Instruct. \cref{fig:diffOrgadMistralInstr} depicts the performance difference between their method and our Spilled $\Delta E$ with Min pooling. Positive values indicate cases where Spilled $\Delta E$ outperforms the method of \cite{orgad2024llms}. All numbers are computed as percentages.}
    \label{fig:confusion_heatmap_MistralInstruct}
\end{figure}

\clearpage
\subsection{Additional results for Cross-testing with Real World Benchmarks}
\label{sec:appendix-cross}
\begin{figure}
    \centering    \includegraphics[width=0.75\linewidth]{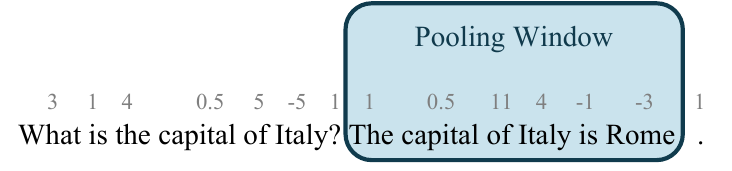}
    \caption{Example of the Pooling Window}
    \label{fig:PoolingWindow}
\end{figure}

\cref{tab:detection_full_no_flip} shows how our method compares with the baseline methods, \cite{orgad2024llms} and Logit $E^\ell$. This table was obtained by using various pooling methods in the pooling window from which we measure possible hallucinations. More details below based on the example in \cref{fig:PoolingWindow}:
\begin{itemize}
    \item \textbf{Min}: minimum energy value in the pooling window. Energy Measured: $-3$
    \item \textbf{Max}: maximum energy value in the pooling window. Energy Measured: $11$
    \item \textbf{Mean}: mean of all the energies in the pooling window. Energy Measured: $2.08$
    \item \textbf{Last Token}: energy of the last token in the pooling window. Energy Measured: $-3$
    \item \textbf{After Last Token}: energy of the first token after the pooling method. Energy Measured: $1$
\end{itemize}

\clearpage
\subsubsection{Success Cases}

{\footnotesize\texttt{Question: ``Which planet is known as the Red Planet?''}}
\vspace{2pt}

\begin{emphabox}
\textbf{Logits:}
\stkLOGITS{The}{0.71} \stkLOGITS{Red}{0.87} \stkLOGITS{Planet}{0.92} \stkLOGITS{is}{0.90} \stkLOGITS{Mars}{1.00} \stkLOGITS{.}{0.31} \correct

\vspace{4pt}
\textbf{Ours:}
\stk{The}{0.44} \stk{Red}{0.57} \stk{Planet}{0.12} \stk{is}{0.28} \stk{Mars}{0.22} \stk{.}{0.18} \correct
\end{emphabox}

\begin{emphabox}
\textbf{Logits:}
\stkLOGITS{The}{0.77} \stkLOGITS{Red}{0.95} \stkLOGITS{Planet}{1.00} \stkLOGITS{is}{0.98} \stkLOGITS{Jupiter}{1.00} \stkLOGITS{.}{0.36} \incorrect

\vspace{4pt}
\textbf{Ours:}
\stk{The}{0.44} \stk{Red}{0.57} \stk{Planet}{0.12} \stk{is}{0.28} \stk{Jupiter}{0.65} \stk{.}{0.24} \incorrect
\end{emphabox} 

\vspace{4pt}
{\footnotesize\texttt{Question: ``What is the largest mammal in the world?''}}
\vspace{2pt}

\begin{emphabox}
\textbf{Logits:}
 \stkLOGITS{The}{0.69} \stkLOGITS{largest}{0.90} \stkLOGITS{mamm}{0.72} \stkLOGITS{al}{0.39} \stkLOGITS{in}{0.98} \stkLOGITS{the}{1.00} \stkLOGITS{world}{0.76} \stkLOGITS{is}{0.81} \stkLOGITS{the}{0.84} \stkLOGITS{Blue}{0.91} \stkLOGITS{Whale}{0.94} \correct

\vspace{4pt}
\textbf{Ours:}
 \stk{The}{0.33} \stk{largest}{0.44} \stk{mamm}{0.64} \stk{al}{0.00} \stk{in}{0.39} \stk{the}{0.31} \stk{world}{0.23} \stk{is}{0.25} \stk{the}{0.25} \stk{Blue}{0.54} \stk{Whale}{0.20}  \correct
\end{emphabox}

\begin{emphabox}
\textbf{Logits:}
\stkLOGITS{The}{0.69} \stkLOGITS{largest}{0.90} \stkLOGITS{mamm}{0.72} \stkLOGITS{al}{0.39} \stkLOGITS{in}{0.98} \stkLOGITS{the}{1.00} \stkLOGITS{world}{0.76} \stkLOGITS{is}{0.81} \stkLOGITS{the}{0.84} \stkLOGITS{House}{0.76} \stkLOGITS{Cat}{0.78} \stkLOGITS{.}{0.20} \incorrect

\vspace{4pt}
\textbf{Ours:}
\stk{The}{0.33} \stk{largest}{0.44} \stk{mamm}{0.64} \stk{al}{0.00} \stk{in}{0.39} \stk{the}{0.31} \stk{world}{0.23} \stk{is}{0.25} \stk{the}{0.25} \stk{House}{0.70} \stk{Cat}{0.50} \stk{.}{0.31} \incorrect
\end{emphabox}

{\footnotesize\texttt{Question: ``Who painted the Mona Lisa?''}}
\vspace{2pt}

\begin{emphabox}
\textbf{Logits:}
\stkLOGITS{The}{0.40} \stkLOGITS{Mona}{0.97} \stkLOGITS{Lisa}{0.45} \stkLOGITS{was}{0.97} \stkLOGITS{painted}{0.88} \stkLOGITS{by}{0.89} \stkLOGITS{Leonardo}{1.00} \stkLOGITS{da}{0.87} \stkLOGITS{Vinci}{0.46} \stkLOGITS{.}{0.00} \correct

\vspace{4pt}
\textbf{Ours:}
\stk{The}{0.34} \stk{Mona}{0.69} \stk{Lisa}{0.01} \stk{was}{0.43} \stk{painted}{0.29} \stk{by}{0.10} \stk{Leonardo}{0.32} \stk{da}{0.33} \stk{Vinci}{0.00} \stk{.}{0.11} \correct
\end{emphabox}

\begin{emphabox}
\textbf{Logits:}
 \stkLOGITS{The}{0.58} \stkLOGITS{Mona}{0.99} \stkLOGITS{Lisa}{0.62} \stkLOGITS{was}{0.99} \stkLOGITS{painted}{0.93} \stkLOGITS{by}{0.93} \stkLOGITS{Pablo}{1.00} \stkLOGITS{Esc}{0.47} \stkLOGITS{obar}{0.00} \stkLOGITS{.}{0.25} \incorrect

\vspace{4pt}
\textbf{Ours:}
\stk{The}{0.41} \stk{Mona}{0.72} \stk{Lisa}{0.11} \stk{was}{0.49} \stk{painted}{0.37} \stk{by}{0.20} \stk{Pablo}{0.91} \stk{Esc}{0.95} \stk{obar}{0.00} \stk{.}{0.30} \incorrect
\end{emphabox}

\vspace{4pt}
{\footnotesize\texttt{Question: ``What gas do plants breathe in for photosynthesis?''}}
\vspace{2pt}

\begin{emphabox}
\textbf{Logits:}
\stkLOGITS{They}{0.71} \stkLOGITS{breathe}{0.68} \stkLOGITS{in}{1.00} \stkLOGITS{carbon}{0.84} \stkLOGITS{dioxide}{0.90} \correct

\textbf{Ours:}
\stk{They}{0.62} \stk{breathe}{0.46} \stk{in}{0.31} \stk{carbon}{0.46} \stk{dioxide}{0.18} \correct
\end{emphabox}

\begin{emphabox}
\textbf{Logits:}
\stkLOGITS{They}{0.71} \stkLOGITS{breathe}{0.68} \stkLOGITS{in}{1.00} \stkLOGITS{oxygen}{0.77} \incorrect

\textbf{Ours:}
\stk{They}{0.58} \stk{breathe}{0.42} \stk{in}{0.29} \stk{oxygen}{0.36} \incorrect
\end{emphabox}

\clearpage
\vspace{4pt}
{\footnotesize\texttt{Question: ``In which continent is Egypt Located?''}}
\vspace{2pt}

\begin{emphabox}
    
\textbf{Logits:}
\stkLOGITS{Egypt}{0.89} \stkLOGITS{is}{0.92} \stkLOGITS{located}{0.88} \stkLOGITS{in}{1.00} \stkLOGITS{Africa}{0.89} \correct

\textbf{Ours:}
\stk{Egypt}{0.30} \stk{is}{0.17} \stk{located}{0.20} \stk{in}{0.04} \stk{Africa}{0.00} \correct
\end{emphabox}

\begin{emphabox}
\textbf{Logits:}
\stkLOGITS{Egypt}{0.88} \stkLOGITS{is}{0.91} \stkLOGITS{located}{0.88} \stkLOGITS{in}{1.00} \stkLOGITS{Europe}{0.81} \incorrect

\textbf{Ours:}
\stk{Egypt}{0.27} \stk{is}{0.14} \stk{located}{0.17} \stk{in}{0.00} \stk{Europe}{0.48} \incorrect
\end{emphabox}

\vspace{4pt}
{\footnotesize\texttt{Question: ``What is the fastest land animal?''}}
\vspace{2pt}

\begin{emphabox}
\textbf{Logits:}
\stkLOGITS{The}{0.54} \stkLOGITS{fastest}{0.67} \stkLOGITS{land}{0.82} \stkLOGITS{animal}{0.52} \stkLOGITS{is}{0.69} \stkLOGITS{the}{0.72} \stkLOGITS{che}{0.58} \stkLOGITS{et}{1.00} \stkLOGITS{ah}{0.33} \correct

\textbf{Ours:}
\stk{The}{0.44} \stk{fastest}{0.61} \stk{land}{0.54} \stk{animal}{0.25} \stk{is}{0.38} \stk{the}{0.37} \stk{che}{0.77} \stk{et}{0.45} \stk{ah}{0.00} \correct
\end{emphabox}

\begin{emphabox}
\textbf{Logits:}
\stkLOGITS{The}{0.64} \stkLOGITS{fastest}{0.80} \stkLOGITS{land}{0.97} \stkLOGITS{animal}{0.62} \stkLOGITS{is}{0.82} \stkLOGITS{the}{0.86} \stkLOGITS{lion}{0.81} \incorrect

\textbf{Ours:}
\stk{The}{0.20} \stk{fastest}{0.38} \stk{land}{0.30} \stk{animal}{0.00} \stk{is}{0.14} \stk{the}{0.13} \stk{lion}{0.49} \incorrect
\end{emphabox}

\vspace{4pt}
{\footnotesize\texttt{Question: ``What is the hardest natural substance on Earth?''}}
\vspace{2pt}

\begin{emphabox}
\textbf{Logits:}
\stkLOGITS{The}{0.68} \stkLOGITS{hardest}{0.86} \stkLOGITS{natural}{0.82} \stkLOGITS{substance}{0.78} \stkLOGITS{is}{0.80} \stkLOGITS{diamond}{0.96} \correct

\textbf{Ours:}
\stk{The}{0.24} \stk{hardest}{0.43} \stk{natural}{0.21} \stk{substance}{0.08} \stk{is}{0.02} \stk{diamond}{0.26} \correct
\end{emphabox}

\begin{emphabox}
\textbf{Logits:}
\stkLOGITS{The}{0.68} \stkLOGITS{hardest}{0.86} \stkLOGITS{natural}{0.82} \stkLOGITS{substance}{0.78} \stkLOGITS{is}{0.80} \stkLOGITS{gold}{0.80} \incorrect

\textbf{Ours:}
\stk{The}{0.19} \stk{hardest}{0.34} \stk{natural}{0.17} \stk{substance}{0.06} \stk{is}{0.01} \stk{gold}{0.57} \incorrect
\end{emphabox}

\vspace{4pt}
{\footnotesize\texttt{Question: ``Which ocean is the largest?''}}
\vspace{2pt}

\begin{emphabox}
\textbf{Logits:}
\stkLOGITS{The}{0.85} \stkLOGITS{largest}{0.86} \stkLOGITS{ocean}{0.86} \stkLOGITS{is}{0.88} \stkLOGITS{the}{0.97} \stkLOGITS{Pacific}{0.94} \stkLOGITS{Ocean}{1.00} \correct

\textbf{Ours:}
\stk{The}{0.15} \stk{largest}{0.48} \stk{ocean}{0.06} \stk{is}{0.12} \stk{the}{0.00} \stk{Pacific}{0.18} \stk{Ocean}{0.03} \correct
\end{emphabox}

\begin{emphabox}
\textbf{Logits:}
\stkLOGITS{The}{0.85} \stkLOGITS{largest}{0.86} \stkLOGITS{ocean}{0.87} \stkLOGITS{is}{0.88} \stkLOGITS{the}{0.97} \stkLOGITS{Indian}{0.91} \stkLOGITS{Ocean}{1.00} \incorrect

\textbf{Ours:}
\stk{The}{0.20} \stk{largest}{0.51} \stk{ocean}{0.12} \stk{is}{0.18} \stk{the}{0.06} \stk{Indian}{0.76} \stk{Ocean}{0.00} \incorrect
\end{emphabox}

\clearpage
\subsubsection{Failure Cases}
\vspace{4pt}
{\footnotesize\texttt{Question: ``Who was the first person to walk on the moon?''}}
\vspace{2pt}

\begin{emphabox}
\textbf{Logits:}
\stkLOGITS{Neil}{0.73} \stkLOGITS{Armstrong}{1.00} \correct

\textbf{Ours:}
\stk{Neil}{0.46} \stk{Armstrong}{0.00} \correct
\end{emphabox}

\begin{emphabox}
\textbf{Logits:}
\stkLOGITS{Buzz}{0.67} \stkLOGITS{Ald}{0.31} \stkLOGITS{rin}{0.00} \incorrect

\textbf{Ours:}
\stk{Buzz}{0.63} \stk{Ald}{0.65} \stk{rin}{0.00} \incorrect
\end{emphabox}

{\footnotesize\texttt{Reasoning: ``If there are 3 cars and each car has 4 wheels , how many wheels are there in total? ''}}
\vspace{2pt}

\begin{emphabox}
\textbf{Logits:}
\stkLOGITS{Each}{0.50} \stkLOGITS{car}{0.85} \stkLOGITS{has}{0.83} \stkLOGITS{}{0.93} \stkLOGITS{4}{0.52} \stkLOGITS{wheels}{0.57} \stkLOGITS{.}{0.30} \stkLOGITS{So}{0.61} \stkLOGITS{,}{0.49} \stkLOGITS{for}{0.75} \stkLOGITS{}{0.94} \stkLOGITS{3}{0.36} \stkLOGITS{cars}{0.62} \stkLOGITS{,}{0.42} \stkLOGITS{the}{0.78} \stkLOGITS{total}{0.73} \stkLOGITS{number}{0.77} \stkLOGITS{of}{0.76} \stkLOGITS{wheels}{0.63} \stkLOGITS{is}{0.67} \stkLOGITS{}{0.98} \stkLOGITS{3}{0.25} \stkLOGITS{x}{0.79} \stkLOGITS{}{0.91} \stkLOGITS{4}{0.30} \stkLOGITS{=}{0.86} \stkLOGITS{}{0.95} \stkLOGITS{12}{0.32} \stkLOGITS{wheels}{0.40} \stkLOGITS{.}{0.13} \correct

\vspace{4pt}
\textbf{Ours:}
\stk{Each}{0.72} \stk{car}{0.49} \stk{has}{0.46} \stk{}{0.69} \stk{4}{0.24} \stk{wheels}{0.28} \stk{.}{0.46} \stk{So}{0.62} \stk{,}{0.42} \stk{for}{0.65} \stk{}{0.73} \stk{3}{0.18} \stk{cars}{0.42} \stk{,}{0.45} \stk{the}{0.55} \stk{total}{0.49} \stk{number}{0.42} \stk{of}{0.46} \stk{wheels}{0.36} \stk{is}{0.43} \stk{}{0.71} \stk{3}{0.23} \stk{x}{0.58} \stk{}{0.67} \stk{4}{0.14} \stk{=}{0.50} \stk{}{0.72} \stk{12}{0.00} \stk{wheels}{0.50} \stk{.}{0.43} \correct
\end{emphabox}

\begin{emphabox}
\textbf{Logits:}
\stkLOGITS{Each}{0.50} \stkLOGITS{car}{0.83} \stkLOGITS{has}{0.81} \stkLOGITS{}{0.91} \stkLOGITS{8}{0.45} \stkLOGITS{wheels}{0.59} \stkLOGITS{.}{0.29} \stkLOGITS{So}{0.60} \stkLOGITS{,}{0.45} \stkLOGITS{for}{0.73} \stkLOGITS{}{0.89} \stkLOGITS{3}{0.35} \stkLOGITS{cars}{0.60} \stkLOGITS{,}{0.42} \stkLOGITS{the}{0.75} \stkLOGITS{total}{0.72} \stkLOGITS{number}{0.74} \stkLOGITS{of}{0.73} \stkLOGITS{wheels}{0.62} \stkLOGITS{is}{0.66} \stkLOGITS{}{1.00} \stkLOGITS{3}{0.25} \stkLOGITS{x}{0.78} \stkLOGITS{}{0.91} \stkLOGITS{8}{0.27} \stkLOGITS{=}{0.85} \stkLOGITS{}{0.92} \stkLOGITS{14}{0.19} \stkLOGITS{wheels}{0.42} \stkLOGITS{.}{0.09} \incorrect

\vspace{4pt}
\textbf{Ours:}
\stk{Each}{0.68} \stk{car}{0.41} \stk{has}{0.38} \stk{}{0.65} \stk{8}{0.40} \stk{wheels}{0.25} \stk{.}{0.34} \stk{So}{0.63} \stk{,}{0.35} \stk{for}{0.57} \stk{}{0.71} \stk{3}{0.05} \stk{cars}{0.35} \stk{,}{0.38} \stk{the}{0.46} \stk{total}{0.40} \stk{number}{0.36} \stk{of}{0.38} \stk{wheels}{0.28} \stk{is}{0.33} \stk{}{0.69} \stk{3}{0.10} \stk{x}{0.49} \stk{}{0.65} \stk{8}{0.02} \stk{=}{0.40} \stk{}{0.64} \stk{14}{0.36} \stk{wheels}{0.46} \stk{.}{0.34} \incorrect
\end{emphabox}

\vspace{4pt}
{\footnotesize\texttt{Reasoning: ``What is the square root of 64?''}}
\vspace{2pt}
\begin{emphabox}
\textbf{Logits:}
\stkLOGITS{The}{0.61} \stkLOGITS{square}{0.94} \stkLOGITS{root}{0.81} \stkLOGITS{of}{0.85} \stkLOGITS{}{0.92} \stkLOGITS{64}{0.23} \stkLOGITS{is}{0.75} \stkLOGITS{}{0.84} \stkLOGITS{8}{0.17} \correct

\textbf{Ours:}
\stk{The}{0.53} \stk{square}{0.55} \stk{root}{0.33} \stk{of}{0.42} \stk{}{0.64} \stk{64}{0.12} \stk{is}{0.40} \stk{}{0.68} \stk{8}{0.00} \correct
\end{emphabox}

\begin{emphabox}
\textbf{Logits:}
\stkLOGITS{The}{0.61} \stkLOGITS{square}{0.94} \stkLOGITS{root}{0.81} \stkLOGITS{of}{0.85} \stkLOGITS{}{0.92} \stkLOGITS{64}{0.23} \stkLOGITS{is}{0.75} \stkLOGITS{}{0.84} \stkLOGITS{10}{0.20} \incorrect

\textbf{Ours:}
\stk{The}{0.51} \stk{square}{0.53} \stk{root}{0.30} \stk{of}{0.40} \stk{}{0.62} \stk{64}{0.09} \stk{is}{0.38} \stk{}{0.67} \stk{10}{0.31} \incorrect
\end{emphabox}

\vspace{4pt}
{\footnotesize\texttt{Question: ``What blood type is known as the universal donor?''}}
\vspace{2pt}

\begin{emphabox}
\textbf{Logits:}
\stkLOGITS{O}{1.00} \stkLOGITS{negative}{0.62} \correct

\textbf{Ours:}
\stk{O}{0.17} \stk{negative}{0.28} \correct
\end{emphabox}

\begin{emphabox}
\textbf{Logits:}
 \stkLOGITS{AB}{0.88} \stkLOGITS{positive}{0.75} \incorrect

\textbf{Ours:}
\stk{AB}{0.41} \stk{positive}{0.27} \incorrect
\end{emphabox}

\end{document}